\documentclass[review]{elsarticle}

\usepackage{lineno,hyperref}
\modulolinenumbers[5]

\journal{Journal of \LaTeX\ Templates}

\usepackage{graphics} % for pdf, bitmapped graphics files
\usepackage{graphicx}
\usepackage[caption=false]{subfig}
\usepackage{arydshln}
\usepackage{hyperref}
\usepackage{multirow}
\usepackage[dvipsnames,table,xcdraw]{xcolor}

\usepackage{amsmath} % assumes amsmath package installed
\usepackage{amssymb}  % assumes amsmath package installed
\usepackage{threeparttable}
\usepackage[]{algorithm}
\usepackage{adjustbox}

\usepackage[noend]{algpseudocode}
\DeclareMathOperator*{\argmin}{arg\,min}

\usepackage{eurosym}
\usepackage{comment}

\usepackage{scalefnt}
\usepackage{tikz}
\usetikzlibrary{mindmap, shadows, arrows.meta,patterns,decorations.pathmorphing,decorations.markings,
                chains, shapes,backgrounds, shapes.multipart,
                positioning, shapes.geometric, arrows, positioning, calc, matrix}
                
\tikzset{
  basic box/.style={
    shape=rectangle, rounded corners, align=center,
    draw=#1, fill=#1!25},
  header node/.style={
    Minimum Width=header nodes,
    font=\strut\Large\ttfamily,
    text depth=+0pt,
    fill=white, draw},
  header/.style={%
    inner ysep=+1.5em,
    append after command={
      \pgfextra{\let\TikZlastnode\tikzlastnode}
      node [header node] (header-\TikZlastnode) at (\TikZlastnode.north) {#1}
      node [span=(\TikZlastnode)(header-\TikZlastnode)] at (fit bounding box) (h-\TikZlastnode) {}
    }
  },
  hv/.style={to path={-|(\tikztotarget)\tikztonodes}},
  vh/.style={to path={|-(\tikztotarget)\tikztonodes}},
  fat blue line/.style={ultra thick, blue}
}

\newcommand*{\info}[4][12]{%
\node [ annotation, #3, scale=0.9, text width = #1em,
      inner sep =5mm ] at (#2) {%
\list{$\bullet$}{\topsep=0pt\itemsep=0pt\parsep=4pt
\parskip=0pt\labelwidth=8pt\leftmargin=4pt
\itemindent=0pt\labelsep=2pt}%
#4
\endlist
 };
}

\newcommand*{\infolong}[4][16]{%
\node [ annotation, #3, scale=0.9, text width = #1em,
      inner sep =5mm ] at (#2) {%
\list{$\bullet$}{\topsep=0pt\itemsep=0pt\parsep=4pt
\parskip=0pt\labelwidth=6pt\leftmargin=4pt
\itemindent=0pt\labelsep=2pt}%
#4
\endlist
 };
}

\def\checkmark{\tikz\fill[scale=0.4](0,.35) -- (.25,0) -- (1,.7) -- (.25,.15) -- cycle;} 

\usepackage{longtable}
\usepackage{amssymb}% http://ctan.org/pkg/amssymb
\usepackage{pifont}% http://ctan.org/pkg/pifont
\newcommand{\xmark}{\ding{55}}%
\usepackage[nonumberlist, acronym, toc, nogroupskip]{glossaries}
\setglossarystyle{alttree}
\setglossarypreamble[acronym]{\vspace*{-\baselineskip}}

\newacronym{OCP}{OCP}{Optimal Control Problem}
\newacronym{OCPs}{OCPs}{Optimal Control Problems}
\newacronym{MHE}{MHE}{Model Horizon Estimation}
\newacronym{NMPC}{NMPC}{Nonlinear Model Predictive Control}
\newacronym{LQR}{LQR}{Linear Quadratic Regulator}
\newacronym{ILQR}{iLQR}{Iterative Linear Quadratic Regulator}
\newacronym{MPC}{MPC}{Model Predictive Control}
\newacronym{DDP}{DDP}{Differential Dynamic Programming}
\newacronym{NLP}{NLP}{Nonlinear Programming}
\newacronym{QP}{QP}{Quadratic Programming}
\newacronym{MIQP}{MIQP}{ Mixed Integer Quadratic Programming}
\newacronym{CBF}{CBFs}{Control Barrier Functions}
\newacronym{SDDM}{SDDM}{State-dependant Distance Metric}
\newacronym{CMPCC}{CMPCC}{Corridor-based Model Predictive Contouring Control}
\newacronym{RRG}{RRG}{Rapidly-exploring Random Graph}
\newacronym{IRIS}{IRIS}{Iterative Regional Inflation by Semi-definite Programming}
\newacronym{SFC}{SFC}{Safe Flight Corridor}
\newacronym{JPS}{JPS}{Jump Point Search}
\newacronym{GTO}{GTO}{Gradient-based Trajectory Optimization}
\newacronym{SQP}{SQP}{Sequential Quadratic Programming}
\newacronym{MPCC}{MPCC}{Mathematical Program with Complementarity Constraints}
\newacronym{ESDF}{ESDF}{Euclidean Signed Distance Field}
\newacronym{PGO}{PGO}{Path-guided Optimization}
\newacronym{LTI}{LTI}{Linear Time Invariant}
\newacronym{TOPP}{TOPP}{Time-Optimal Parameterization of a given Path}
\newacronym{CHOMP}{CHOMP}{Covariant Hamiltonian Optimization for Motion Planning}
\newacronym{MAVs}{MAVs}{Multirotor Aerial Vehicles}
\newacronym{MAV}{MAV}{Multirotor Aerial Vehicle}
\newacronym{UAVs}{UAVs}{Unmanned Aerial Vehicles}
\newacronym{UAV}{UAV}{Unmanned Aerial Vehicle}
\newacronym{LQG}{LQG}{Linear Quadratic Gaussian}
\newacronym{KF}{KF}{Kalman Filter}
\newacronym{EO}{EO}{Elastic Optimization}
\newacronym{QCQP}{QCQP}{Quadratically Constrained Quadratic Programming}
\newacronym{RHC}{RHC}{Receding Horizon Control}
\newacronym{BFGS}{BFGS}{Broyden—Fletcher—Goldfarb—Shanno}
\newacronym{TSDF}{TSDF}{Truncated Signed Distance Field}
\newacronym{PRM}{PRM}{Probabilistic Road Map}
\newacronym{GTC}{GTC}{Geometric Tracking Control}

\newglossary{symbols}{sym}{sbl}{List of Symbols}

\makeglossaries

\newglossaryentry{state}{type=symbols,name={\ensuremath{\mathbf{x}}}, sort=state, description={state vector and its derivative is denoted as $\mathbf{\dot{x}}$. Term $\mathbf{x^+}$ depicts the next state given the current state $\mathbf{x}$, and term $\mathbf{x}_k$ denotes discrete state at time t equals k}}
\newglossaryentry{input}{type=symbols,name={\ensuremath{\mathbf{u}}}, sort=input, description={control input. The term $\mathbf{u}*$ denoted as the optimal control inputs}}
\newglossaryentry{p}{type=symbols,name={\ensuremath{\mathbf{p}}}, sort=p, description={position (m) in $\mathbb{R}^3$ and its derivative is denoted as $\dot{\mathbf{p}}$. $\mathbf{p}_*, * \in {x, y, z}$, stands for position alone * component}}
\newglossaryentry{poly}{type=symbols,name={\ensuremath{p}}, sort=poly, description={dth order polynomial, which is a function of time. Term $\dot{p}(t), \ddot{p}(t)$ (or  $p(t)^{(1)}, p(t)^{(2)}$) denote the higher order derivatives of $p(t)$}}
\newglossaryentry{polyc}{type=symbols,name={\ensuremath{\lambda}}, sort=c, description={polynomial coefficients, e.g., $p(t) = \lambda_d t^d + \hdots + \lambda_1 t + \lambda_0, \; t \in [0, dt]$, where d is the order of the polynomial}}
\newglossaryentry{v}{type=symbols,name={\ensuremath{\mathbf{v}}}, sort=v, description={velocity (m/s) in $\mathbb{R}^3$  and its derivative is denoted as $\dot{\mathbf{v}}$. $\mathbf{v}_*, * \in {x, y, z}$, stands for velocity alone * component}}
\newglossaryentry{omega}{type=symbols,name={\ensuremath{\mathbf{\omega}}}, sort=omega, description={angular velocity (rad/s) in $\mathbb{R}^3$  and its derivative is denoted as $\dot{\mathbf{\omega}}$}}
\newglossaryentry{psi}{type=symbols,name={\ensuremath{\mathbf{\psi}}}, sort=psi, description={orientation is represented as quaternion in $\mathbb{R}^3$  and its derivative is denoted as $\dot{\mathbf{\psi}}$. $\mathbf{\psi}_*, * \in {x, y, z}$, stands for orientation alone * component}}
\newglossaryentry{force}{type=symbols,name={\ensuremath{\mathbf{f}=[f_1, f_2, f_3, f_4]^T}}, sort=force, description={system input or total trust that is applied for each of the motors in N (Newton}}
\newglossaryentry{fdes}{type=symbols,name={\ensuremath{\mathbf{f}_d}}, sort=fdes, description={discrete system dynamics}}
\newglossaryentry{fcons}{type=symbols,name={\ensuremath{\mathbf{f}_c}}, sort=fcons, description={continuous system dynamics}}
\newglossaryentry{delta}{type=symbols,name={\ensuremath{\delta}}, sort=delta, description={Euler or Runge Kutta discretization time step }}
\newglossaryentry{z}{type=symbols,name={\ensuremath{\mathbf{z}}}, sort=z, description={system output}}
\newglossaryentry{qz}{type=symbols,name={\ensuremath{(q)}}, sort=qz, description={apices $^{(q)}$ stipulates the qth derivative, for example $\mathbf{z}^{(q)}$}}
\newglossaryentry{space}{type=symbols,name={\ensuremath{C}}, sort=space, description={configuration space that can be one of these: $C_{free}$, $C_{obs}$, $C_{unknown}$, and  $C_{unknown}$}}
\newglossaryentry{d}{type=symbols,name={\ensuremath{d}}, sort=d, description={order of polynomial}}
\newglossaryentry{traj}{type=symbols,name={\ensuremath{\Gamma}}, sort=traj, description={initial trajectory; the optimal trajectory is defined as $\Gamma^*$, trajectory derivatives are defined as $\dot{\Gamma}$ and $\ddot{\Gamma}$, and trajectory is a function of time, i.e., $\Gamma_T(t)$}}
\newglossaryentry{params}{type=symbols, name={\ensuremath{\xi}}, sort=params, description={regularization parameter}}
\newglossaryentry{cc}{type=symbols, name={\ensuremath{c}}, sort=params, description={formulation of cost function, where $c(\cdot), \cdot$ is denotes the inputs}}
\newglossaryentry{hrep}{type=symbols, name={\ensuremath{A}}, sort=hrep, description={H representation of polytope, i.e, $A^T\mathbf{p} = b$}}
\newglossaryentry{w}{type=symbols, name={\ensuremath{\mathbf{w}}}, sort=hrep, description={the optimal estimation for states or/and controls after minimizing given cost function}}
\newglossaryentry{g}{type=symbols, name={\ensuremath{g}}, sort=g, description={Equality constraints are denoted by $g_1(\mathbf{w}$), whereas inequality constraints are denoted by $g_2(\mathbf{w}$)}}

\glsaddall
\glsfindwidesttoplevelname

\begin{document}

\begin{frontmatter}

\title{Motion Planning for Multirotor Aerial Vehicles in Plan-based Control Paradigm: a Review}
% \tnotetext[mytitlenote]{Fully documented templates are available in the elsarticle package on \href{http://www.ctan.org/tex-archive/macros/latex/contrib/elsarticle}{CTAN}.}

% \author{Geesara Kulathunga}
% \address{Radarweg 29, Amsterdam}
% \fntext[myfootnote]{Since 1880.}

\author[mymainaddress]{Geesara Kulathunga}
% \author[mymainaddress]{Dmitry Devitt}
\author[mymainaddress]{Alexandr Klimchik}


% \author[mysecondaryaddress]{Global Customer Service\corref{mycorrespondingauthor}}
% \cortext[mycorrespondingauthor]{Corresponding author}

\address[mymainaddress]{Center for Technologies in Robotics and Mechatronics Components, Innopolis University, Russia}

\begin{abstract}
In general, optimal motion planning can be performed both locally and globally. In such a planning, the choice in favour of either local or global planning technique mainly depends on whether the environmental conditions are dynamic or static. Hence, the most adequate choice is to use local planning or local planning alongside global planning. When designing optimal motion planning both local and global, the key metrics to bear in mind are execution time,  asymptotic optimality, and quick reaction to dynamic obstacles. Such planning approaches can address the aforesaid target metrics more efficiently compared to other approaches such as path planning followed by smoothing. Thus, the foremost objective of this study is to analyse related literature in order to understand how the motion planning, especially trajectory planning, problem is formulated, when being applied for generating optimal trajectories in real-time for Multirotor Aerial Vehicles (MAVs), impacts the listed metrics. As a result of the research,  the trajectory planning problem was broken down into a set of subproblems, and the lists of methods for addressing each of the problems were identified and described in detail. Subsequently,  the most prominent results from 2010 to 2022 were summarized and presented in the form of a timeline.
\end{abstract}


\end{frontmatter}

\linenumbers

% \printglossary[type=\acronymtype, title=List of Abbreviations, nonumberlist]
% \printglossary[type=symbolslist, title=Notations]

\printglossaries

\section{Introduction}
%motivation
Adroit motion planning of flying little creatures, such as birds and butterflies, is an extraordinarily demanding task for several reasons, including aggressive maneuver. An example of such high-speed maneuver need is one in particularly tight spots where the environment is obstacle-rich. Researchers have been trying to replicate similar maneuvers using two different types of aerial vehicles: conventional and unconventional. In this research we deal with conventional areal vehicles, for instance, \gls*{UAVs}, \gls*{MAVs}, etc. Recent progression in computation capabilities and embedded sensing has been boosting the procedure of mimicking natural flying animals; this advancement has enabled plenty of new opportunities in diverse fields: inspection, autonomous transportation, logistics, delivery, areal photography, post-disaster and medical services. Yet optimal motion planning remains a crucial task in all the fields listed above. In optimal motion planning, the environmental reasoning can not be predictable since environmental conditions change rapidly. Hence, there are various challenges to be addressed to obtain highly efficient and optimal motion planning. In this paper, we mainly focus on how researchers have been addressing these challenges in optimal motion planning to obtain robust navigation in various domains for ~\gls*{MAVs}.  

\begin{figure}[h!]
    \centering
    \includegraphics[width=1\linewidth]{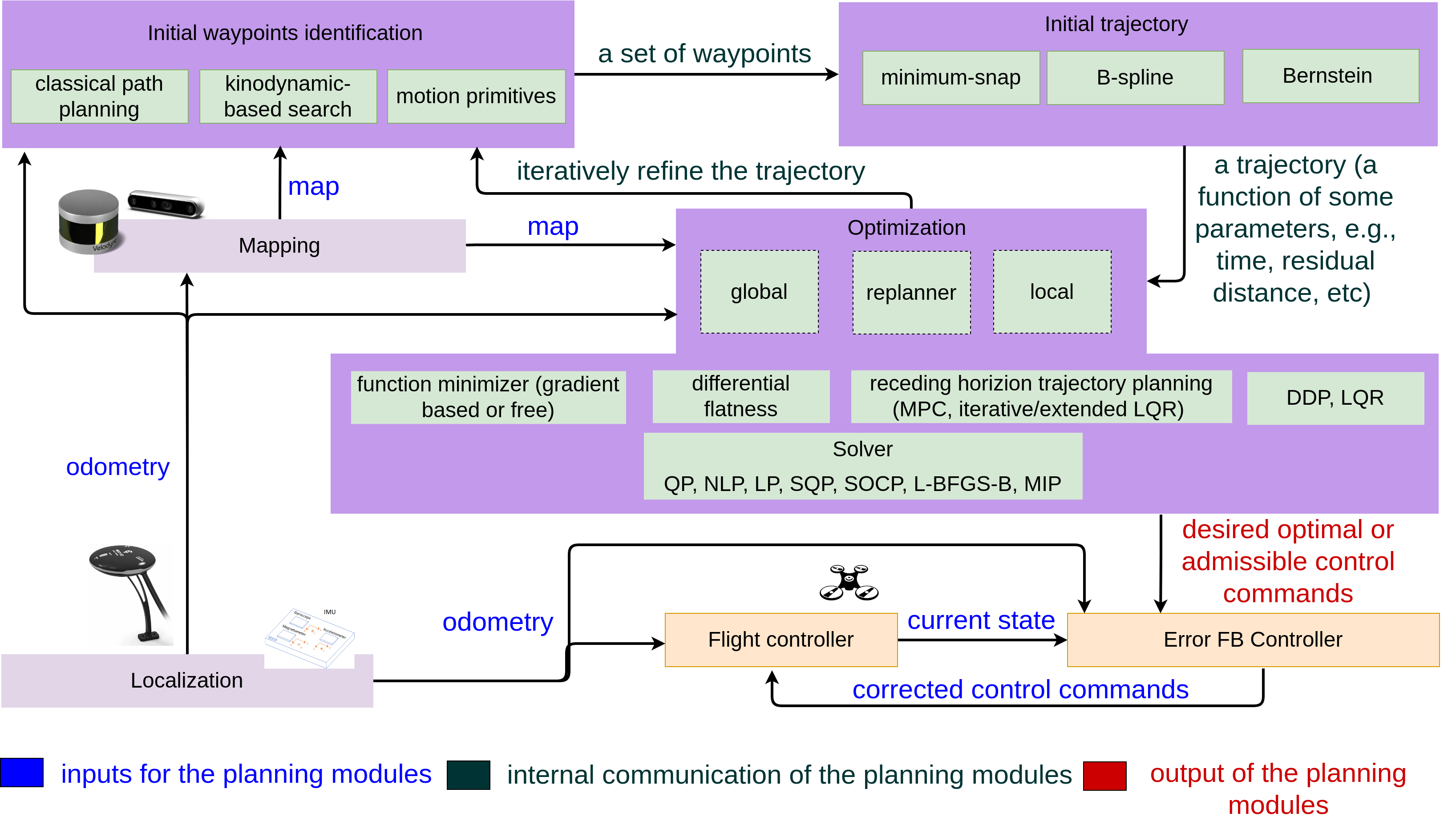}
    \caption{The overview of plan-based control paradigm in the context of trajectory planning problem formulation. There are various ways to formulate the trajectory planning problem, each of which consists of a set of sub-modules (green color boxes) depending on the problem behaviour}
    \label{fig:system_overview}
\end{figure}

%why we select this topic
In most of the foregoing applications, the environment is entirely or partially unexplored. Furthermore, unpredictable events can occur at any time due for numerous reasons. Thus, to tackle those unexpected problems in real time, a fast and accurate optimal motion planning technique is required. In general, the optimal motion planning problem is divided into a few subcategories: path planning followed by smoothing, kinodynamic search-based trajectory generation, and motion primitive-based approaches. Among them, plan-based control approaches are the most widely used and efficient way to address the considered problem compared to the other two approaches. Plenty of plan-based control strategies have been proposed throughout the recent decade, showing promising results; this is one of the key motivating factors for reviewing plan-based control, especially for industrial~\gls*{MAVs}. Most of the industrial~\gls*{MAVs} such as quadrotors have their low-level controllers, for example, PX4~\cite{px4}, DJI~\cite{dji}, that can be operated independently irrespective of high-level execution commands. Moreover, such controllers reduce the overhead and complexity for developing high-level planning algorithms due to their independence. In other words, the same planner can be deployed on different firmware by implementing an interface between a high-level planner and a low-level controller. Thus, we narrowed down our study to considering only plan-based control approaches (Fig.~\ref{fig:system_overview}), particularly in application to industrial~\gls*{MAVs}.  

\begin{figure}[H]
% \centering
\begin{tikzpicture}[thick,scale=0.7, every annotation/.style = {fill=none, font=\tiny	}, trim left=-8cm]
  \path[mindmap, concept color=gray!20,text=black]
    node[concept, xshift=-2cm, scale=0.5] (CORE) {Motion planning in plan-based control paradigm}
    [clockwise from=180]
    child[concept] { node[concept, scale=0.9, fill=CornflowerBlue, ] (cg) {Receding horizon trajectory planning} }
    child[concept] { node[concept, scale=0.9, fill=Emerald, ] (ms) {Motion model selection} }
    child[concept] { node[concept, scale=0.9, fill=OliveGreen, ] (fx) {Free space segmentation} }
    child[concept] { node[concept, scale=0.9, fill=Periwinkle, ] (iw) {Intermediate waypoints identification} }
    child[concept] { node[concept, scale=0.9, fill=RoyalBlue, ] (tg) {Initial trajectory generation} }
    child[concept] { node[concept, scale=0.9, fill=PineGreen, ] (ct) {Continuous trajectory refinement} };
    \info{ms.north east}{above, yshift=0em, xshift=-5em}{
      \item \textcolor{Emerald}{Differential Flatness~\cite{mellinger2011minimum} and Partial Differential Flatness~\cite{ramasamy2014dynamically}}
      \item \textcolor{Emerald}{Empirical model~\cite{wanasinghe2015relative}}
      \item \textcolor{Emerald}{Exact model~\cite{van2016extended}}
    }
    \infolong{fx.north east}{above, yshift=1em, xshift=0em}{
      \item \textcolor{OliveGreen}{Convex segmentation: \gls*{IRIS}~\cite{deits2015computing}, \gls*{SFC}~\cite{ling2017building}, Stereographic Projection~\cite{savin2017algorithm}, ~Extracting convex polytopes\cite{zhong2020generating}}
      \item \textcolor{OliveGreen}{Octomap and \gls*{ESDF} mapping\cite{kulathunga2022optimization},  map building and construct KD-tree~\cite{bentley1975multidimensional}}
      \item \textcolor{OliveGreen}{A set of geometrical shapes such as cubes~\cite{chen2015real, gao2018online}, spheres~\cite{gao2016online, gao2019flying} and  polyhedrons~\cite{liu2017planning}}}, 
    \info{iw.north east}{above, yshift=0em, xshift=2.0em}{
      \item \textcolor{Periwinkle}{Path planning e.g., graph search techniques such as A* and D*~\cite{stentz1997optimal}, sampling-based techniques, i.e., RRT, RRT*~\cite{noreen2016comparison}, \gls*{RRG}~\cite{gao2017gradient}}
      \item \textcolor{Periwinkle}{Kinodynamic and kinematic enable , A*~\cite{zhou2019robust}, RRT*~\cite{webb2013kinodynamic}, FMT*~\cite{allen2016real}}
      \item \textcolor{Periwinkle}{Incorporate motion primitive~\cite{zhou2019robust}}
    }
    \info{tg.north east}{above right, yshift=-6.3em, xshift=-0.2em}{
      \item \textcolor{RoyalBlue}{Minimum-snap~\cite{mellinger2011minimum}}  
      \item \textcolor{RoyalBlue}{B-spline (uniform or non-uniform)~\cite{zhou2019robust, ding2019efficient}, Minimum-time B-spline~\cite{rousseau2019minimum}}
      \item \textcolor{RoyalBlue}{Bernstein basis polynomial~\cite{gao2018online}}
    }
    \info{ct.north east}{left,yshift=-4em, xshift=-4em}{
      \item \textcolor{PineGreen}{Refinement trajectory cost in most of the cases, defined by $J(\Gamma) =  \xi_{smooth}J_{smooth}(\Gamma) + \xi_{obs}J_{obs}(\Gamma) + \xi_{soft}J_{soft}(\Gamma) + \xi_{end}J_{end}(\Gamma) $; different types of techniques are employed considering a few or all of the preceding individual costs: jerk or snap~\cite{mellinger2011minimum}, end point~\cite{usenko2017real}, obstacle~\cite{ratliff2009chomp}, elastic band~\cite{ding2019efficient} for control points refinement} 
    }
    \info{cg.north east}{above,yshift=-2em, xshift=-7em}{
      \item \textcolor{CornflowerBlue}{\gls*{ILQR}~\cite{li2004iterative},  Extended \gls*{LQR}~\cite{van2014iterated}, \gls*{LQG}~\cite{todorov2008general}, \gls*{MPC}~\cite{nageli2017real}, \gls*{CMPCC}~\cite{ji2020cmpcc}}
      \item \textcolor{CornflowerBlue}{A set of \gls*{CBF} for improving the robustness~\cite{ames2016control}}
    };
\end{tikzpicture}
\caption{The basic building blocks that encounter in trajectory planning problem. In general, a considered trajectory planning problem can be comprised of one or more blocks sequentially or in parallel to fulfil the desired needs} \label{fig:main_classification}
\end{figure}

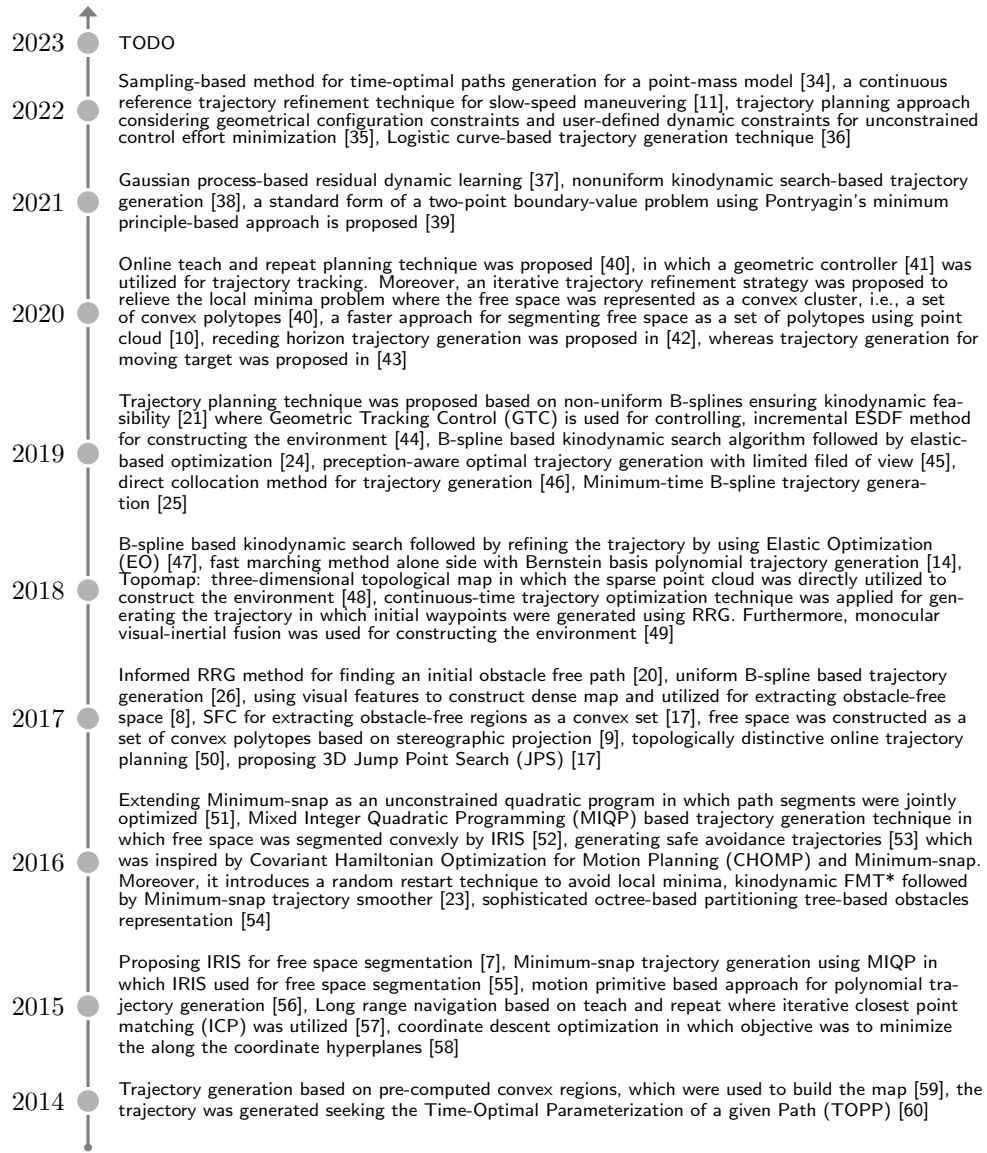
\begin{figure} [H]
\centering
\begin{tikzpicture}[
node distance = 1mm and 3mm,
  start chain = A going below,
   dot/.style = {circle, draw=white, very thick, fill=gray!60, minimum size=3mm},
   box/.style = {rectangle, text width=115mm, inner xsep=4mm, inner ysep=1mm, font=\sffamily\scriptsize		\linespread{0.84}\selectfont, on chain},]
    \begin{scope}[every node/.append style={box}]
    \node {TODO};
    \node {Sampling-based method for time-optimal paths generation for a point-mass model~\cite{romero2022time}, a continuous reference trajectory refinement technique for slow-speed maneuvering~\cite{kulathunga2022optimization}, trajectory planning approach considering geometrical configuration constraints and user-defined dynamic constraints for unconstrained control effort minimization~\cite{wang2022geometrically}, Logistic curve-based trajectory generation technique~\cite{upadhyay2022generation}};
    \node {Gaussian process-based residual dynamic learning~\cite{torrente2021data}, nonuniform kinodynamic search-based trajectory generation~\cite{tang2021real}, a standard form of a two-point boundary-value problem using Pontryagin’s minimum principle-based approach is proposed~\cite{heidari2021trajectory}};
    \node {Online teach and repeat planning technique was proposed~\cite{gao2020teach}, in which a geometric controller~\cite{lee2010geometric} was utilized for trajectory tracking. Moreover, an iterative trajectory refinement strategy was proposed to relieve the local minima problem where the free space was represented as a convex cluster, i.e., a set of convex polytopes~\cite{gao2020teach}, a faster approach for segmenting free space as a set of polytopes using point cloud~\cite{zhong2020generating}, receding horizon trajectory generation was proposed in~\cite{zinage20203d}, whereas trajectory generation for moving target was proposed in~\cite{xi2020trajectory}};
    \node {Trajectory planning technique was proposed based on non-uniform B-splines ensuring kinodynamic feasibility~\cite{zhou2019robust} where~\gls*{GTC} is used for controlling, incremental~\gls*{ESDF} method for constructing the environment~\cite{han2019fiesta}, B-spline based kinodynamic search algorithm followed by elastic-based optimization~\cite{ding2019efficient}, preception-aware optimal trajectory generation with limited filed of view~\cite{murali2019perception}, direct collocation method for trajectory generation~\cite{abadi2019optimal}, Minimum-time B-spline trajectory generation~\cite{rousseau2019minimum}};
    \node {B-spline based kinodynamic search followed by refining the trajectory by using~\gls*{EO}~\cite{ding2018trajectory}, fast marching method alone side with Bernstein basis polynomial trajectory generation~\cite{gao2018online}, Topomap: three-dimensional topological map in which the sparse point cloud was directly utilized to construct the environment~\cite{blochliger2018topomap}, continuous-time trajectory optimization technique was applied for generating the trajectory in which initial waypoints were generated using \gls*{RRG}. Furthermore, monocular visual-inertial fusion was used for constructing the environment~\cite{lin2018autonomous}};
    \node {Informed \gls*{RRG} method for finding an initial obstacle free path~\cite{gao2017gradient}, uniform B-spline based trajectory generation~\cite{usenko2017real}, using visual features to construct dense map and utilized for extracting obstacle-free space~\cite{ling2017building}, \gls*{SFC} for extracting obstacle-free regions as a convex set~\cite{liu2017planning}, free space was constructed as a set of convex polytopes based on stereographic projection~\cite{savin2017algorithm}, topologically distinctive online trajectory planning~\cite{rosmann2017integrated}, proposing 3D \gls*{JPS}~\cite{liu2017planning}};
    \node {Extending Minimum-snap as an unconstrained quadratic program in which path segments were jointly optimized~\cite{richter2016polynomial}, \gls*{MIQP} based trajectory generation technique in which free space was segmented convexly by \gls*{IRIS}~\cite{landry2016aggressive}, generating safe avoidance trajectories~\cite{oleynikova2016continuous} which was inspired by~\gls*{CHOMP} and Minimum-snap. Moreover, it introduces a random restart technique to avoid local minima, kinodynamic FMT* followed by Minimum-snap trajectory smoother~\cite{allen2016real}, sophisticated octree-based partitioning tree-based obstacles representation~\cite{chen2016online}};
    \node {Proposing \gls*{IRIS} for free space segmentation~\cite{deits2015computing}, Minimum-snap trajectory generation using~\gls*{MIQP} in which~\gls*{IRIS} used for free space segmentation~\cite{deits2015efficient}, motion primitive based approach for polynomial trajectory generation~\cite{mueller2015computationally}, Long range navigation based on teach and repeat where iterative closest point matching (ICP) was utilized~\cite{krusi2015lighting}, coordinate descent optimization in which objective was to minimize the along the coordinate hyperplanes~\cite{wright2015coordinate}};
    \node {Trajectory generation based on pre-computed convex regions, which were used to build the map~\cite{schulman2014motion}, the trajectory was generated seeking the \gls*{TOPP}~\cite{pham2014general}};
    \end{scope}
\draw[very thick, gray, {Triangle[length=4pt)]}-{Circle[length=3pt]},
      shorten <=-3mm, shorten >=-3mm]           % <--- here is adjusted additional arrow's 
    (A-1.north west) -- (A-10.south west);
\foreach \i [ count=\j] in {2023, 2022, 2021, 2020, 2019, 2018, 2017, 2016, 2015, 2014}
    \node[dot,label=left:\i] at (A-\j.west) {};
    \end{tikzpicture}
\caption{The most prominent related research outcomes which led the success of the trajectory planning for ~\gls*{MAVs} in the last decade} \label{fig:time_line}
\end{figure}  

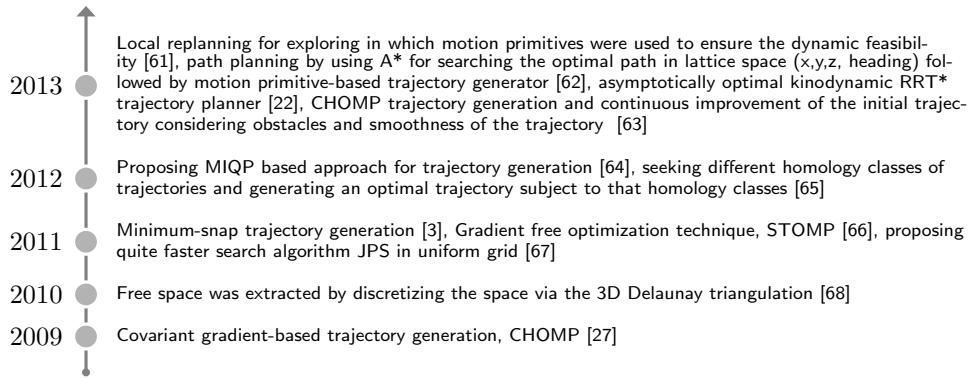
\begin{figure}[H]
\centering
\begin{tikzpicture}[
node distance = 1mm and 3mm,
  start chain = A going below,
   dot/.style = {circle, draw=white, very thick, fill=gray!60, minimum size=3mm},
   box/.style = {rectangle, text width=115mm, inner xsep=4mm, inner ysep=1mm, font=\sffamily\scriptsize		\linespread{0.84}\selectfont, on chain},]
    \begin{scope}[every node/.append style={box}]
    \node {Local replanning for exploring in which motion primitives were used to ensure the dynamic feasibility~\cite{pivtoraiko2013incremental}, path planning by using A* for searching the optimal path in lattice space (x,y,z, heading) followed by motion primitive-based trajectory generator~\cite{macallister2013path},  asymptotically optimal kinodynamic RRT* trajectory planner~\cite{webb2013kinodynamic},  \gls*{CHOMP} trajectory generation and continuous improvement of the initial trajectory considering obstacles and smoothness of the trajectory ~\cite{zucker2013chomp}};
    \node {Proposing \gls*{MIQP} based approach for trajectory generation~\cite{mellinger2012mixed}, seeking different homology classes of trajectories and generating an optimal trajectory subject to that homology classes~\cite {bhattacharya2012topological} };
    \node {Minimum-snap trajectory generation~\cite{mellinger2011minimum}, Gradient free optimization technique, STOMP~\cite{kalakrishnan2011stomp}, proposing quite faster search algorithm \gls*{JPS} in uniform grid~\cite{harabor2011online}};
    \node { Free space was extracted by discretizing the space via the 3D Delaunay triangulation~\cite{lovi2010incremental}} ;
    \node {Covariant gradient-based trajectory generation, \gls*{CHOMP}~\cite{ratliff2009chomp}} ;
    \end{scope}
\draw[very thick, gray, {Triangle[length=4pt)]}-{Circle[length=3pt]},
      shorten <=-3mm, shorten >=-3mm]           % <--- here is adjusted additional arrow's 
    (A-1.north west) -- (A-5.south west);
\foreach \i [ count=\j] in {2013, 2012, 2011, 2010, 2009}
    \node[dot,label=left:\i] at (A-\j.west) {};
    \end{tikzpicture}
\caption{The most prominent related research outcomes which led the success of the trajectory planning for ~\gls*{MAVs} in the last decade} \label{fig:time_line1}
\end{figure}  

%what are we focusing on
The main limitation of~\gls*{MAVs} is low flight time. Hence, a MAV should be capable of executing robust, agile, aggressive maneuver while ensuring dynamic feasibility and guaranteeing smoothness of the trajectory in low flight time. Furthermore, trajectory plotting should be performed within an obstacle-free zone at high-speed to handle a given mission effectively. Such behaviour is imposed by adhering to a set of constraints. If and only if the constraints are incorporated appropriately, desired needs can be fulfilled. Obtaining the right constraints at the right moment and applying appropriate control sequences to improve motion quality is the key objective of any plan-based control approach. Yet the procedure of obtaining such right constraints is an open research problem due to its complexity and numerous other challenges that should be handled simultaneously. For example, \gls*{MAV} has been widely employed in video-making related fields in recent years, cinematographic aerial shooting being one of the popular areas of interest during the last five years. In such shooting, generating smooth, obstacle-free trajectories is the main challenge. Besides, various other challenges exist, and most of them are application-specific. In this work, we examine the most common problems related to trajectory planning applications in the paradigm of plan-based control, and how researchers have been alleviating those problems by proposing compelling solutions. 

%what will be there in the paper
In optimal trajectory planning, trajectory generation and controlling the \gls*{MAV} are strongly interconnected. For~\gls*{MAVs}, the trajectory generation process is relatively easy due to the dynamic properties of the~\gls*{MAVs}. When dynamic obstacles are incorporated, the trajectory has to be refined at a high rate in order to keep a smooth maneuver despite increased computational demands. Moreover, understanding close-in obstacles' positions relative to the \gls*{MAV} is crucial for making decisions in real-time; this arises a new challenge: the one of the rapidity and accuracy of relative environment reconstruction, which essentially is how obstacles constraints are added to the problem formulation. Yet another challenge is of the impact of the obstacles and constraints on the smoothness and dynamic feasibility of the generated trajectory. After conducting an extensive literature review on the topic of trajectory planning for~\gls*{MAVs}, we were able to isolate basic building blocks that are essential for optimal motion planning as shown in Fig.~\ref{fig:main_classification}. Each of the primary components plays a key role in the process of trajectory generation. The rest of the paper focuses on understanding how those building blocks are interconnected in solving trajectory planning problems. 

The rest of the paper is organized as follows: section~\ref{sec:model_selection} explains what type of motion model is likely to be suitable for defining the dynamics of ~\gls*{MAV} based on the chosen trajectory generation technique. Then, state-of-the-art techniques on how to find initial tentative waypoints for trajectory generation is explained in section~\ref{sec:initial_waypoints_generation}. Section~\ref{sec:init_trajecotry_generation} presents an extensive review on initial trajectory generation techniques.  Section~\ref{sec:free_space_extraction} explains how free space is extracted and incorporated into trajectory planning.  The trajectory refinement process is explained in section.~\ref{sec:continuous_trajectory}. Horizon-based trajectory planning techniques are described in section~\ref{sec:residing_horizon}. Various solvers which can be used to solve the optimization problem are detailed under section~\ref{sec:solvers_opt}. 

\section{Motion Model Selection}\label{sec:model_selection}
Exact model, empirical model and differential flatness are the main techniques that can be employed for selecting the most appropriate motion model for a specified application. The appropriate motion model selection procedure varies depending on the problem formulation. For example, planning followed by controlling approaches does not necessarily have an exact motion model mainly due to high computational demands. In such scenarios, an empirical motion model is sufficient for planning, since a dedicated controller is utilized for controlling the quadrotor. 

\subsection{Exact Model}
In general, ~\gls*{MAV} dynamics is described by 12-DOF. However, in planning followed by high-level controlling approaches, it is not required to define an actual motion model for planning, since a high-level controller consists of a fully-fledged quadrotor motion model. In most circumstances, the planner is comprised of approximated quadrotor dynamics; this is due to computational complexity, which is not adequate for real-time onboard processing. Hence, the motion model selection process depends on the approach that formulates needs. In~\cite{van2016extended}, the researchers  proposed a 12-DOF motion model whose state vector is defined by \gls*{state} =  $[\gls*{p}^{\top}, \gls*{v}^{\top}, \gls*{psi}^{\top}, \gls*{omega}^{\top}]$, where $\gls*{psi}$, \gls*{p}, \gls*{v} and \gls*{omega} stand for orientation (rad), position (m), velocity (m/s) and angular velocity (rad/s) in $\mathbb{R}^3$, respectively. The system input or total trust that is applied for each of the motors is given by \gls*{force} (N). System dynamics is determined as $\dot{\mathbf{x}} = [\dot{\mathbf{p}}^{\top} , \dot{\gls*{v}}^{\top}, \dot{\gls*{psi}}^{\top}, \dot{\gls*{omega}}^{\top}]$, where  $\dot{\mathbf{p}} = \gls*{v}$,
    $\dot{\gls*{v}} = -g \cdot \mathbf{e}_z+ \frac{(\mathbf{f} \cdot \exp{[\gls*{psi}]}\cdot \mathbf{e}_z-k_v\cdot \gls*{v})}{m}$, 
    $\dot{\gls*{psi}} = \gls*{omega} + \frac{1}{2}[\gls*{psi}]\cdot \gls*{omega}+(1-\frac{1}{2} \frac{\left \|\gls*{psi} \right \|}{tan(\frac{1}{2}\left \| \gls*{psi} \right \|)})[\gls*{psi}]^2 \cdot \gls*{omega}/\left \| \gls*{psi}\right \|^2$,
     $\dot{\gls*{omega}} = J^{-1}(\rho(f_2-f_4)\mathbf{e}_x) + \rho(f_3-f_1)\mathbf{e}_y + k_m(f_1-f_2+f_3-f_4)\mathbf{e}_z-[\gls*{omega}]\cdot J \cdot w)$, $g=9.8ms^{-2}$ and $\mathbf{e}_i, \; i=x,y,z$ stand for standard basis vectors in $\mathbb{R}^3$, $k_v, m, J, \rho$ and $k_m$ are robot specific constants.

\subsection{Empirical Model}
Other than the exact model, a 6-DOF motion model was proposed for governing quadrotor in a distributed setup~\cite{trawny2010interrobot}. Later, it was reduced to 4-DOF motion model~\cite{wanasinghe2015relative}. Furthermore, in~\cite{mehrez2017optimization}, a 4-DOF motion was used for controlling several quadrotors in a distributed setup in which~\gls*{NMPC} and \gls*{MHE} are incorporated for relative tracking where the relative motion model was defined as:

\begin{equation}\label{eq:discrete_model}
    \begin{aligned}
     \dot{\gls*{state}} = \gls*{fcons}(\gls*{state}, \gls*{input}, \gls*{psi}_z)
       = \begin{bmatrix}
\dot{\gls*{p}}_x \\ 
\dot{\gls*{p}}_y \\ 
\dot{\gls*{p}}_z \\ 
\dot{\gls*{psi}_z}
\end{bmatrix} = 
\begin{bmatrix}
\gls*{v}_x cos(\gls*{psi}_z) - \gls*{v}_y sin(\gls*{psi}_z) - \bar{\gls*{v}}_{x} + \gls*{p}_y\bar{\dot{\gls*{psi}_z}}\\ 
\gls*{v}_x sin(\gls*{psi}_z) + \gls*{v}_y cos(\gls*{psi}_z) - \bar{\gls*{v}}_{y} - \gls*{p}_x\bar{\dot{\gls*{psi}_z}} \\ 
\gls*{v}_z  - \bar{\gls*{v}}_{z}\\ 
\dot{\gls*{psi}_z} - \bar{\dot{\gls*{psi}_z}}
\end{bmatrix},
    \end{aligned}
\end{equation} where the function $\gls*{fcons}(\cdot) : \mathbb{R}^{n_u} \times \mathbb{R}^{n_x} \times \mathbb{R}^{n_{ru}} \rightarrow \mathbb{R}^{n_x}$ and $n_x = n_u = n_{ru} = 4$. The current control input is given by $\gls*{input} = [\gls*{v}_x, \gls*{v}_y, \gls*{v}_z, \dot{\gls*{psi}_z}]$, whereas relative control input $\gls*{input}_{ru}$ is denoted by $[\bar{\gls*{v}}_x, \bar{\gls*{v}}_y, \bar{\gls*{v}}_z, \bar{\dot{\gls*{psi}_z}}]$. $\gls*{state} = [\gls*{p}_x, \gls*{p}_y, \gls*{p}_z, \gls*{psi}_z]$ is the state of the motion model, where $\gls*{p}_i, i \in \{x, y, z\}$ is the position of the ~\gls*{MAV} in the world frame. $\gls*{psi}_z$ and $\bar{\gls*{psi}_z}$ denote the yaw angle or heading angle around the z axis and relative yaw angle, respectively. Derivative of $\gls*{psi}_z$ and $\bar{\gls*{psi}_z}$ are denoted by $\dot{\gls*{psi}_z}$ and $\bar{\dot{\gls*{psi}_z}}$, respectively. $\gls*{v}_i, i \in \{x, y, z\}$ denote the velocities on each direction, whereas $\dot{\gls*{p}}_i, i \in \{x, y, z\}$ gives the derivatives of $\gls*{p}_i$. Since discrete space was chosen for controlling the system, Euler discrete model~(\ref{eq:discrete_model}) was formulated as follows: 
\begin{equation}\label{eq:euler_discretization_dis}
\begin{aligned}
\mathbf{\gls*{state}^+} = \gls*{fdes}(\gls*{state}, \gls*{input}, \gls*{psi}_z) 
 = \begin{bmatrix}
p_x\\ 
p_y\\ 
p_z\\ 
\gls*{psi}_z
\end{bmatrix} + \gls*{delta}\begin{bmatrix}
v_x cos(\gls*{psi}_z) - v_y sin(\gls*{psi}_z) - \bar{v}_{x} + y\bar{\dot{\gls*{psi}_z}}\\ 
v_x sin(\gls*{psi}_z) + v_y cos(\gls*{psi}_z) - \bar{v}_{y} - x\bar{\dot{\gls*{psi}_z}} \\ 
v_z  - \bar{v}_{z}\\ 
\dot{\gls*{psi}_z} - \bar{\dot{\gls*{psi}_z}}
\end{bmatrix},
\end{aligned}
\end{equation} where $\gls*{delta}$ is the sampling period and $\gls*{fdes}(\cdot) : \mathbb{R}^{n_x} \times \mathbb{R}^{n_u} \times \mathbb{R}^{n_{ru}} \rightarrow \mathbb{R}^{n_x}$. $\gls*{fcons}$ and $\gls*{fdes}$ denote continuous and discrete dynamics, respectively. $\mathbf{x^+}$ depicts the next state given the current state $\mathbf{x}$. Subsequently, the motion model was simplified to 4-DOF for trajectory tracking for a quadrotor ~\cite[eq.(1)]{mpc_mc}. In this trajectory-tracking approach, planning followed by the high-level controlling paradigm was applied. Such an approach was introduced because a simplified motion model is a reasonable choice for achieving real-time performance. Quadrotor state was defined as  $\gls*{state} = [\gls*{p}_x, \gls*{p}_y, \gls*{p}_z,\gls*{psi}_z]^T \in  \mathbb{R}^{n_x}$, whereas input to the system was given by $\gls*{input} = [\gls*{v}_x, \gls*{v}_y, \gls*{v}_z, \dot{\gls*{psi}_z}]^T \in  \mathbb{R}^{n_u}$. The simplified motion model was given by
\begin{equation}\label{eq:treacking_motion}
    \begin{aligned}
     \dot{\gls*{state}} = \gls*{fcons}(\gls*{state}, \mathbf{u})
       = \begin{bmatrix}
\dot{\gls*{p}}_x \\ 
\dot{\gls*{p}}_y \\ 
\dot{\gls*{p}}_z \\ 
\dot{\gls*{psi}}_z
\end{bmatrix} = 
\begin{bmatrix}
\gls*{v}_x cos(\gls*{psi}_z) - \gls*{v}_y sin(\gls*{psi}_z)\\ 
\gls*{v}_x sin(\gls*{psi}_z) + \gls*{v}_y cos(\gls*{psi}_z)\\ 
\gls*{v}_z \\ 
\dot{\gls*{psi}_z} 
\end{bmatrix},
    \end{aligned}
\end{equation}  where $\gls*{fcons}(\cdot) : \mathbb{R}^{n_x}\times  \mathbb{R}^{n_u} \rightarrow \mathbb{R}^{n_x}$ and $n_x = n_u = 4$. The discretization of~(\ref{eq:treacking_motion}) was given by:    
\begin{equation}\label{eq:euler_discretization}
\begin{aligned}
\mathbf{x^+} = \gls*{fdes}(\gls*{state}, \gls*{input}) 
 = \begin{bmatrix}
\gls*{p}_x\\ 
\gls*{p}_y\\ 
\gls*{p}_z\\ 
\gls*{psi}_z
\end{bmatrix} + \gls*{delta}\begin{bmatrix}
\gls*{v}_x cos(\gls*{psi}_z) - \gls*{v}_y sin(\gls*{psi}_z)\\ 
\gls*{v}_x sin(\gls*{psi}_z) + \gls*{v}_y cos(\gls*{psi}_z)\\ 
\gls*{v}_z\\ 
\dot{\gls*{psi}_z}
\end{bmatrix},
\end{aligned}
\end{equation} where $\mathbf{f_d(\cdot)}: \mathbb{R}^{n_x} \times \mathbb{R}^{n_u} \rightarrow \mathbb{R}^{n_x}$.

\subsection{Differential Fatness} \label{diff_flatness}
Here differential flatness~\cite{van1998real} provides algebraic functions (e.g., polynomials) which analytically map the trajectory and whose higher-order derivatives map to system states and inputs. Since the Nth order polynomial can be differentiated up to N-1 times, the differential fatness property ensures the feasibility of the trajectory and generates appropriate control commands. More precisely, let 
\begin{equation}
    \begin{aligned}
    \dot{\gls*{state}} = \gls*{fcons}(\gls*{state}, \gls*{input}) \quad \gls*{state} \in \mathbb{R}^{n_x}, \gls*{input} \in \mathbb{R}^{n_u}.\\
    \end{aligned}
\end{equation} be a nonlinear system. According to to~\cite{sferrazza2016numerical}, if the system is differentially flat, there always exists a flat output, namely $\gls*{z} \in \mathbb{R}^{n_z}$, where the dimension of the output is given by $n_z$. In such a system, states and control inputs can also be formulated from the system flat outputs whose derivatives are mapped through functions, namely $\varrho$ and $\tau$. Let $\gls*{z} = \Im (\gls*{state}, \gls*{input}, \dot{\gls*{input}},...,\gls*{input}^{\gls*{qz}})$ be the flat output, holding $\gls*{state} = \varrho( \gls*{z}, \dot{\gls*{z}},..., \gls*{z}^{(r)})$ and $\gls*{input} = \tau ( \gls*{z}, \dot{\gls*{z}},..., \gls*{z}^{(r)})$, where apices $^{(i)}$ stipulates the ith derivative. Along with that, the explicit trajectory generation process can benefit when it uses differentially flat systems, for example, $\varrho$ and $\tau$ can be a dth order polynomial $\gls*{poly}(t)$. Then, $\gls*{state}^{\top}(t) = [\gls*{poly}^{\top}(t) \; \dot{\gls*{poly}}^{\top}(t) \; \ddot{\gls*{poly}}^{\top}(t)]$ be the state of the system at time $t$ in which $\dot{\gls*{poly}}^T$ and $\ddot{\gls*{poly}}^T$ indicate the velocity and acceleration of the system, respectively. Control inputs can be determined by jerk~\cite{krishnan2019towards}, namely $\dddot{\gls*{poly}}^T(t)$ where $\gls*{poly}(t) = \gls*{polyc}_d t^d + \hdots + \gls*{polyc}_1 t + \gls*{polyc}_0, \; t \in [0, dt]$, where $\gls*{polyc}_i, i=0,...,d$ are the polynomial coefficients. There are various ways to construct these kinds of polynomials, including Minimum-snap, B-spline, etc. 

\section{Initial Waypoints Identification}\label{sec:initial_waypoints_generation}
Generally speaking, robots have a limited sensing range. So, planning a trajectory out of such a sensing range would be counterproductive. Hence, local trajectory planning and refinement when a robot moves is the optimal choice. With the help of sensing capabilities within the robots' sensing range, the robot's surrounded environment can be constructed as the intersection of three separate disjoint sets: free-known ($\gls*{space}_{free}$), occupied ($\gls*{space}_{obs}$) and unknown ($\gls*{space}_{unknown}$). Once $\gls*{space}_{free} \cup \gls*{space}_{unknown}$ is identified, a set of intermediate waypoints is needed to navigate the robot along the trajectory from the start position to the desired position. There are various techniques for finding a set of intermediate waypoints: sampling-based techniques (e.g., RRT*, \gls*{PRM}), path-searching techniques (e.g., A*, D*, \gls*{JPS}) and so forth. Moreover, kinodynamic properties are incorporated into preceding intermediate waypoints finding techniques to ensure the dynamic feasibility of the robot. One of the first kinodynamic-based path planning approaches was proposed in~\cite{dolgov2010path} in which a variant of the A* method alongside with kinodynamic properties was applied to ensure the dynamic feasibility. Subsequently, several different methods were proposed for enhancing path planning, ensuring the dynamic feasibility by kinodynamic properties, including motion primitive-based approaches.

Motion primitive-based approaches(~\cite{mueller2015computationally,florence2020integrated, lopez2017aggressive}) can be utilized for finding intermediate waypoints and for trajectory generation. Gordon et al.~\cite{gordon1974b} proposed a set of motion primitives for connecting edges of the graph that was constructed from A*. In this method, motion primitives were used to defining state vector \gls*{state}(t) and control input $\gls*{input}(t)$ as a \gls*{LTI} system as follows:
\begin{equation}
\begin{aligned}
    \gls*{state}_i(t) = [\gls*{poly}_i(t)^{\top}, \dot{\gls*{poly}_i}(t)^{\top}, ..., \gls*{poly}_i^{({k_r-1})}(t)^{\top}]^{\top} \in \gls*{state}_i(t) \subset \mathbb{R}^{3 \times k_r},  \\ \quad \gls*{poly}_i(t) = [\gls*{poly}_x(t), \gls*{poly}_y(t), \gls*{poly}_z(t)]^T, 
    \gls*{input}_i(t) = \gls*{poly}^{({k_r})}(t),
    \end{aligned} 
\end{equation} where $\gls*{p}_\mu(t) = \Sigma_{j=0}^{d} \gls*{polyc}_jt^j, \; \mu \in \{x,y,z\}$, which is formulated similar to~(\ref{eq_min_poly_dim_1}), while $k_r$ and $\gls*{d}$ are the order of the derivative and the order of the polynomial, respectively.  
\begin{equation}
\begin{aligned}
\dot{\mathbf{x}}_i(t) = A\mathbf{x}_i(t) + B\mathbf{u}_i(t), \\
    A = \begin{bmatrix}
0 & I_3 & 0 & \cdots  & 0\\ 
0 &  0& I_3 & \cdots & 0\\ 
\vdots & \vdots & \vdots & \ddots & \vdots \\ 
0 & \hdots & \hdots  & 0 & I_3 \\ 
0 & \hdots & \hdots & 0 & 0 
\end{bmatrix}, \quad B = \begin{bmatrix}
0\\ 
0\\ 
\vdots \\ 
0 \\ 
I_3
\end{bmatrix}. 
\end{aligned}
\end{equation} Hence, given control policy $\gls*{input}_i(t)$ and initial state $\gls*{state}(0)$, a sequence of succeeding states for a given time duration is determined by
\begin{equation}\label{eq:motion_pre_19}
    \gls*{state}_i(t) = e^{At}\gls*{state}(0) + \int_0^t e^{A(t-\gamma )}B\gls*{input}(\gamma )d\gamma, 
\end{equation} where $\gamma $ is the time duration that control policy is applied. In~\cite{gordon1974b}, to define the actual and heuristic cost of A*, the researchers used motion primitives, which are defined (as shown) in~(\ref{eq:motion_pre_19}), and calculated initial waypoints set.  

Another interesting approach to finding a set of initial intermediate waypoints is by using fast marching methods. In general, fast marching methods~\cite{sethian1999level} are applied to track the propagation of a convoluted interface such as wavefront, especially in image processing. Let $\varphi$ be a close curve in a plane $\in \mathbb{R}^3$ that propagates orthogonally to the plane with a speed $v(\gls*{p})$, assume $v>0$. Given $\bigtriangledown T$ time period, propagation of the plane can be described by $|\bigtriangledown T(x)| = \frac{1}{v(\gls*{p})}$ based on Eikonal partial differential equation~\cite{sava20013} where $\gls*{p}$ is the position in $\mathbb{R}^3$ and the arrival time is formulated by $T(x)$. Fast marching concept was applied for path searching in~\cite{gao2018online} by proposing a method for calculating velocity map. In this method, the arrival time was determined by assessing the desired velocity at the considered position. Hence, arrival time was calculated by backtracking from the goal pose to the start pose along the minimum cost path, which can be estimated from the gradient descendant. Though gradient descendant may trap in a local minimum, when smart marching is applied, gradient descendant does not trap in local minimum due to fast marching nature; this property was proved in~\cite{lavalle2006planning}. To define the velocity map,~\gls*{ESDF} was utilized to get the closest obstacle poses from the given pose. A quadrotor should move faster when there are no close-in obstacles and should be slower when it is moving through a cluttered environment. Such a behaviour was mimicked by incorporating a hyperbolic tangential function, i.e., tanh. With such an assumption, the corresponding velocity was calculated based on~(\ref{velocity_prof})
\begin{equation}\label{velocity_prof}
    v(l) = \left\{\begin{matrix}
v_{max} (tanh(l-e)+1)/2, & 0 \leq l\\ 
0, & l< 0
\end{matrix}\right., 
\end{equation} where $v_{max}$ is the maximum velocity a quadrotor can fly, l is the distance to the closest obstacle from the considered pose $\gls*{p}$ and e is Euler's constant. 

\section{Initial Trajectory Generation}\label{sec:init_trajecotry_generation}
Let us consider a non-linear system in the form of $\dot{\gls*{state}}(t) = \gls*{fcons}(\gls*{state}(t), \gls*{input}(t))$ with initial state $\gls*{state}(t_0) = \gls*{state}_0$,   
 where state vector and control inputs are denoted by $\mathbf{x} \in R^{n_x}$  and $ \mathbf{u} \in R^{n_u}$, respectively. When generating an initial trajectory ($\gls*{traj}$), ensuring dynamic feasibility is a must. In other words, $\gls*{state}$ and $\gls*{input}$ satisfy the following constraints:
\begin{equation}
     \mathbf{x} \in X \subseteq \mathbf{R}^{n_x}, \quad \mathbf{u} \in U \subseteq \mathbf{R}^{n_u}
\end{equation} In addition to these constraints, safety constraints should also be imposed after reasoning the environment, to guarantee safety. The environment or configuration space C can be decomposed into $C_{obs}$ and $C_{free}$. Hence, a set of constraints should be introduced for the quadrotor to always be within free space $\gls*{state} \in C_{free} = C/\ C_{obs}$. Hence, the initial trajectory generation process has to consider both said types of constraints simultaneously so that the quadrotor would have a smooth flying experience.      

\subsection{Define Trajectory}
Let $\gls*{traj}  \gets C \subset \mathbb{R}^{\gls*{d}}$ be an initial trajectory, which is parameterized as a function of time where d denotes the C's dimension. Since $\gls*{traj}$ is a function, the objective of the trajectory generator is to determine the precise objective, which will eventually provide the optimal trajectory in a timely manner satisfying constraints and hypotheses that are imposed. Hence, optimal trajectory, namely $\gls*{traj}^*$, can be posed as a discrete or continuous~\gls*{OCP}~\cite{bergman2019optimization}:

\begin{equation}\label{eq:trajectory_info}
\begin{aligned}
\gls*{traj}^* = \min_{\gls*{input}(\cdot)} \quad & J(\gls*{state}(0), \gls*{input}(\cdot))\\
\textrm{s.t.} \quad & \gls*{state}(0) = \mathbf{x}_0,\; \gls*{state}(t_n) = \gls*{state}_n\\
  & \dot{\gls*{state}}(t) = \gls*{fcons}(\gls*{state}(t), \gls*{input}(t)) \\ 
  & \gls*{state}(t) \in C_{free}, \; \gls*{input}(t) \in U, \; t \in [t_0, t_n],
\end{aligned}
\end{equation} where $t_0$ and $t_n$ denote the start and terminal time, respectively. Yet another challenging problem is to formulate the objective function, namely $J$. In the following subsections, we discuss several approaches to address this problem.  
   
\subsection{Minimum-snap based Trajectory Generation}\label{min_snap}
Minimum-snap trajectory generation~\cite{mellinger2011minimum} uses the differential flatness property (section~\ref{diff_flatness}) to automate the trajectory generation process. Let quadrotor trajectory be $\gls*{traj}_T(t) = [r_T(t), \gls*{psi}_T(t)]^T$ for flat output $[x,y,z, \gls*{psi}_z]^T$ where $r=[x,y,z]$ is the center position of the \gls*{MAV} with respect to world coordinate system and $\gls*{psi}_z$ is the yaw angle of the \gls*{MAV}. The continuous trajectory can be expressed as follows:
\begin{equation}
    \gls*{traj} (t) : [t_0, t_n] \gets \mathbb{R}^d,
\end{equation} where \gls*{d} is the dimension of the space, e.g., 3. As we defined in section~\ref{diff_flatness}, system states and inputs can be determined in terms of $\gls*{traj}$ and its derivatives. $\gls*{traj}, \dot{\gls*{traj}}$ and $\ddot{\gls*{traj}}$ will correspond to position, velocity and acceleration, respectively. Flat output and its derivatives estimation in Minimum-snap refer to the original work~\cite[eqs. (1-35)]{mellinger2011minimum}.  

In Minimum-snap trajectory parameterization, the total time duration of the trajectory is divided into a set of sub-intervals, i.e., keyframes. Each keyframe consists of a desired position and a yaw angle. A safe corridor is constructed between consecutive keyframes as a set of piecewise polynomial functions to estimate smooth transitions through the keyframes. Let $m_d$ and $\gls*{d}$ be the number of keyframes and the order of the piecewise polynomial functions, respectively. Hence, $\gls*{traj}_T(t)$ can be formulated as
\begin{equation}\label{eq:minimum_snap_total_time}
   \gls*{traj}_T(t) = \left\{\begin{matrix}
 \Sigma_{i=0}^{\gls*{d}} \gls*{traj}_{{i,1}}(t-t_0)^i& t_0 \leq t < t_1 \\ 
\Sigma_{i=0}^{\gls*{d}} \gls*{traj}_{{i,2}}(t-t_1)^i& t_1 \leq t < t_2 \\ 
 \vdots & \\ 
\Sigma_{i=0}^{\gls*{d}} \gls*{traj}_{{i,m_d}}(t-t_{m_d-1})^i& t_{m_d-1} \leq t < t_{m_d}
\end{matrix}\right..
\end{equation}
To generate an optimal trajectory, the following objective is utilized: 
\begin{equation}\label{eq:min_snap}
\begin{aligned}
J(r_T, \psi_T) & = \int_{t_0}^{t_{m_d}} \xi_r\left \| \frac{d^{k_r}r_T}{dt^{k_r}} \right \|^2dt + \xi_\psi  \frac{d^{k_\psi}\psi_T}{dt^{k_\psi}}^2dt \\
 \min_{w} \quad & J(r_T, \psi_T) \\
\textrm{s.t.} \quad & \gls*{traj}_T(t_i) = \gls*{traj}_i \quad i=1,...,{m_d} \\
  \quad & \frac{d^{p}x_T}{dt^{p}}|_{t=t_j} \leq 0 \quad j=0,m_d; \: p=1,...,k_r\\
  \quad & \frac{d^{p}y_T}{dt^{p}}|_{t=t_j} \leq 0 \quad j=0,m_d; \: p=1,...,k_r\\
  \quad & \frac{d^{p}z_T}{dt^{p}}|_{t=t_j} \leq 0 \quad j=0,m_d; \: p=1,...,k_r\\
  \quad & \frac{d^{p}\psi_T}{dt^{p}}|_{t=t_j} \leq 0 \quad j=0,m_d; \: p=1,...,k_\psi,
\end{aligned}
\end{equation} where $\gls*{params}_r$ and $\gls*{params}_\psi$ are regulation parameters, $k_r$ and $k_\psi$ are the order of derivation at each keyframe and $\gls*{traj}_T(t_i) = [x_i, y_i, z_i, {\gls*{psi}_z}_i]^T, i= 0,..., T$. Time intervals, $t_1, t_2,..., t_{m_d}$ can be kept constant or varying when deriving the Minimum-snap trajectory generation. In most cases, having varying time intervals between keyframes is necessary. Mellinger et al.~\cite{mellinger2011minimum} proposed a gradient descent-based approach for finding optimal time intervals between keyframes. Further,  Chen et al.~\cite{chen2015real} utilized A* to find the intermediate waypoints. Based on these estimations, time segments or keyframes are calculated incorporating both velocity and acceleration limits. In the latter approach, the steps listed below were used to obtain intermediate waypoints.  Initially, the environment was constructed as a map using OctoMap. Afterwards, the formed map was split into two subsets: allocated and non-allocated (a set of free spaces). Then, the discrete graph was constructed connecting consecutive free spaces, which were represented as cubes. Afterwards, A* was applied for finding the optimal path segment within each cube. Similar to~(\ref{eq:min_snap}), the researchers set $k_r = 3$ and minimized only total jerk~(\ref{eq:only_jerk}) to minimize the angular velocity. As an aside, minimizing the angular velocity helps to avoid fast rotation.  
\begin{equation}\label{eq:only_jerk}
J = \int_{t_0}^{t_{m_d}} \gls*{params}_r\left \| \frac{d^{k_r}\gls*{traj}_T(t)}{dt^{k_r}} \right \|^2 dt.
\end{equation}

\subsection{Polynomial Trajectory Generation as QP}
 In Minimum-snap trajectory generation, total trust force, i.e., attitude acceleration, is proportional to the fourth derivative (snap) of the trajectory~\cite{mellinger2011minimum}. The gracefulness of such behaviour helps to avoid generating excessive control commands. Subsequently, a slight variation of Minimum-snap trajectory generation was proposed in~\cite{richter2016polynomial}, where segment times or keyframes were fixed initially. Once start and goal positions were provided,  RRT*~\cite{webb2013kinodynamic} was utilized for finding an obstacle-free path between the start and the goal poses as a sequence of optimal waypoints. Initial segment times ($m_d$), which were estimated using optimal waypoints,  were calculated according to the maximum velocities that the quadrotor is allowed to fly due to set technical limits.       
Let $\gls*{poly}_i(t)$ be the $\gls*{d}$th order polynomial in the ith segment that describes as follows:
\begin{equation}\label{eq_min_poly_dim_1}
    \gls*{poly}_i(t) = \gls*{polyc}_0t^0 + \gls*{polyc}_1t^1 + \gls*{polyc}_2t^2+ \gls*{polyc}_3t^3 + ...+ \gls*{polyc}_{\gls*{d}}t^{\gls*{d}}.
\end{equation} Each $\gls*{poly}_i(t)$ provides a flat output for a given time index t. $\gls*{polyc}_j, j=0,...,\gls*{d}$ denotes the polynomial coefficients. The objective or cost function $J(\gls*{traj}_i)$ can be fully determined by penalizing the derivatives of squares~\cite{richter2016polynomial}:
\begin{equation}
    J(\gls*{traj}_i) = \int_{t_i}^{t_{i+1}} \gls*{params}_0\gls*{poly}_i(t)^2 +  \gls*{params}_1\dot{\gls*{poly}}_i(t)^2 + \gls*{params}_2\ddot{\gls*{poly}}_i(t)^2 + ... + \gls*{params}_{k_r}\gls*{poly}^{({k^i_r})}(t)^2 = P_i^TQ(T_i)P_i,
\end{equation} where $P_i$ is a vector whose elements contain polynomial coefficients: $\gls*{params}_0, \gls*{params}_1,..., \gls*{params}_{k_r^i}$, $k^i_r$ is the highest order of derivative and $Q(T_i)$ is Hassin matrix, which contains the ith segment squares of derivatives. Since there are $m_d$ number of segments, total cost $J(\gls*{traj})$ can be expressed by
\begin{equation} \label{min_snap_1}
    J(\gls*{traj}) = \begin{bmatrix}
P_1\\ 
\vdots \\
P_{m_d}
\end{bmatrix}^T\begin{bmatrix}
 Q(T_1)&  & \\ 
 & \ddots & \\ 
 &  & Q(T_{m_d})
\end{bmatrix}\begin{bmatrix}
P_1\\ 
\vdots \\
P_{m_d}
\end{bmatrix}.
\end{equation}
For a smooth flight experience, ensuring the continuity of derivatives between segments is necessary. Hence, imposing constraints between segments, e.g., velocity, acceleration, jerk and snap is needed, which can be formulated as follows:
\begin{equation} \label{min_snap_2}
    C_i\gls*{poly}_i = \mathbf{d}_i, \quad C_i = \begin{bmatrix}
\gls*{params}_0\\ 
\gls*{params}_{k_r}
\end{bmatrix}_i, \quad \mathbf{d}_i = \begin{bmatrix}
d_0\\ 
d_{k_r}
\end{bmatrix}_i,
\end{equation} where $C_i$ contains a mapping matrix whose entries contain the start and end coefficients of ith segment, whereas $d_i$ contains derivative values, i.e.,  start and end of ith segment. Taking all constraints of $m_n$ segments, 
\begin{equation} \label{min_snap_3}
    C\begin{bmatrix}
\gls*{poly}_1\\ 
\vdots \\
\gls*{poly}_{m_d}
\end{bmatrix} = \begin{bmatrix}
\mathbf{d}_1\\ 
\vdots \\
\mathbf{d}_{m_d}
\end{bmatrix}.
\end{equation} Now this can be solved as a constrained \gls*{QP} problem.

\subsection{Unconstrained Polynomial Trajectory Generation}

The techniques that are used for uconstrained trajectory optimization are faster than constraints optimization. In~\cite{richter2016polynomial}, the researchers extended Minimum-snap trajectory generation as an unconstrained \gls*{QP}. According to their findings, Minimum-snap works well for small segments size. For higher-order polynomials with varying segment sizes, Minimum-snap becomes ill-conditioned. Thus, an unconstrained \gls*{QP} was proposed. After substituting~(\ref{min_snap_2}) and~(\ref{min_snap_3}) into~(\ref{min_snap_1}), $J(\gls*{traj})$ can be reformulated as
\begin{equation}\label{min_snap_4}
\begin{aligned}
    J(\gls*{traj}) =  \underbrace{ \begin{bmatrix}
\mathbf{d}_1\\ 
\vdots \\
\mathbf{d}_{m_d}
\end{bmatrix}^T}_{\mathbf{d}} \underbrace{ \begin{bmatrix}
 C(T_1)&  & \\ 
 & \ddots & \\ 
 &  & C(T_{m_d})
\end{bmatrix}^{-T}}_{C^{-T}}
\underbrace{
\begin{bmatrix}
 Q(T_1)&  & \\ 
 & \ddots & \\ 
 &  & Q(T_{m_d})
\end{bmatrix}
}_Q \\
\begin{bmatrix}
 C(T_1)&  & \\ 
 & \ddots & \\ 
 &  & C(T_{m_d})
\end{bmatrix}^{-1} \begin{bmatrix}
\mathbf{d}_1\\ 
\vdots \\
\mathbf{d}_{m_d}
\end{bmatrix} \\
= \begin{bmatrix}
\mathbf{d_f}\\ 
\mathbf{d}_{p}
\end{bmatrix}^T \underbrace{SC^{-T}QC^{-1}S^T}_R \begin{bmatrix}
\mathbf{d}_f\\ 
\mathbf{d}_{p}
\end{bmatrix} = \begin{bmatrix}
\mathbf{d}_f\\ 
\mathbf{d}_{p}
\end{bmatrix}^T \begin{bmatrix}
R_{ff} & R_{fp}\\ 
R_{pf} & R_{pp}
\end{bmatrix} \begin{bmatrix}
\mathbf{d}_f\\ 
\mathbf{d}_{p}
\end{bmatrix},
\end{aligned}
\end{equation} where $\mathbf{d}$ contains fixed derivatives ($\mathbf{d}_f$) and free derivatives ($\mathbf{d}_p$), S is a permutation matrix (ones and zeros), which is used to correct the order. Then, $\frac{dJ(\gls*{traj})}{d\mathbf{d}_p} = 0 $ yields the optimal value for $\mathbf{d}_p$:
\begin{equation}
    \mathbf{d}_p^* = -R_{pp}^{-1}R_{fp}^T\mathbf{d}_f.
\end{equation} Once $\mathbf{d}_p$ is determined, a polynomial that corresponds to each segment can be recovered.  

\subsection{Unconstrained Polynomial Trajectory Generation with Collision Avoidance}

Oleynikova et al.~\cite{oleynikova2016continuous} extended what Richter~\cite{richter2016polynomial} proposed for adding support for collision avoidance capabilities. They added additional term for calculating the collision cost,
\begin{equation}
\begin{aligned}
    J(\gls*{traj}) = \gls*{params}_{obs}J_{obs}(\gls*{traj}) + \gls*{params}_{smooth}J_{smooth}(\gls*{traj}), \\
    J_{smooth} = \mathbf{d}_f^TR_{ff} + \mathbf{d}_f^TR_{fp}\mathbf{d}_p + \mathbf{d}_pR_{pf}\mathbf{d}_f + \mathbf{d}_p^TR_{pp}\mathbf{d}_p,
\end{aligned}
\end{equation} where $J_{smooth}$ exactly equals ~(\ref{min_snap_4}). To estimate $J_{obs}(\gls*{traj})$, it is required to initially calculate position $\gls*{p}_i(t)$~(\ref{eq_min_poly_dim_1}) and velocity $\gls*{v}_i(t)$ for each axis at time t after selecting the corresponding segment ($i, i=1,...,m_d$)
\begin{equation}
    \begin{aligned}
    \gls*{p}_i(t) = T\gls*{poly}_i, \quad p_i = [\gls*{polyc}_0, \gls*{polyc}_1, ..., \gls*{polyc}_{\gls*{d}}]_i^T, \quad T = [t^0, t^1, t^2 ,..., t^{\gls*{d}}], \\ \quad  \gls*{v}_i(t) = \dot{\gls*{p}}_i(t) = TVp_i, \\ \gls*{p}_i(t)= [\gls*{p}_x(t) \; \gls*{p}_y(t) \; \gls*{p}_z(t)]_i,
    \quad \gls*{v}_i(t) = [\gls*{v}_x(t) \; \gls*{v}_y(t) \; \gls*{v}_z(t)]_i.
    \end{aligned}
\end{equation} Knowing (the values of) $\gls*{p}_i(t)$ and $\gls*{v}_i(t)$, $J_{obs}(\gls*{traj}_i)$ can be fully determined by 
\begin{equation}\begin{aligned}
    J_{obs}(\gls*{traj}_i) = \int_{S}^{} \gls*{cc}(\gls*{p}_i(t))ds = \int_{t=0}^{t^{\gls*{d}}} \gls*{cc}(\gls*{p}_i(t))\left \| \gls*{v}_i(t) \right \|dt = \sum_{t=0}^{t^{\gls*{d}}} \gls*{cc}(\gls*{p}_i(t))\left \| \gls*{v}_i(t) \right \|\Delta t \\
    \frac{\partial J_{obs}(\gls*{traj}_i)}{\partial d\gls*{p}_i(t)} =  \sum_{t=0}^{t^{\gls*{d}}}\left \| \gls*{v}_i(t) \right \|\bigtriangledown_i\gls*{cc}(T(C^{-1}S)_{pp})\Delta t + \gls*{cc}(\gls*{p}_i(t))\frac{\gls*{v}_i(t)}{\left \| \gls*{v}_i(t) \right \|} TV(C^{-1}S)_{pp}\Delta t,
    \end{aligned}
\end{equation} where $(C^{-1}S)_{pp}$ is the right-side matrix which corresponds to $\mathbf{d}_p$. For representing the collision cost $\gls*{cc}(\gls*{p}_i(t))$, a line integral of a potential function, i.e.,~(\ref{obs_map}), was used. As total cost is given~(\ref{min_snap_4}), $ J_{obs}(\gls*{traj})$ can be calculated for all the segments provided that $\mathbf{d}_p^*$ can be estimated. In a cluttered environment, optimization problem is most likely to be non-linear as well as non-convex. Thus, \gls*{BFGS}~\cite{head1985broyden} was used to solve the optimization problem. Yet the solver failed to obtain the global minimum most of the time. Hence, several random restarts were needed to find the optimal solution. A thorough discussion of how random restarts were invoked into the optimization problem was detailed in~\cite{schulman2014motion}. 

\subsection{Covariant Gradients for Trajectory Generation}
The significance of covariant gradients technique is that both $J_{obs}(\gls*{traj})$ and $J_{smooth}(\gls*{traj})$ depend solely on physical characteristic of the desired trajectory. In other words, the trajectory generation is invariant to its parameterization. If gradient descent is applied, it depends on the way trajectory is parameterized. The covariant gradients technique removes this dependency. Hence, covariant gradient technique depends solely on physical representation or dynamic quantities of the trajectory with respect to an operator, $\Theta$.
\begin{equation}\label{com_operator}
    \left \| \gls*{traj}  \right \|^2_\Theta = \int \sum_{n=1}^k \gls*{params} (\gls*{traj}(t)^{(n)})^2 dt,  
\end{equation} where $\gls*{params}$ is a constant and apices $^{(n)}$ determine the nth order derivative. The correlation of derivatives between two trajectories: $\gls*{traj}_1$ and $\gls*{traj}_2$, is defined by assuming inner product as given~(\ref{eta_detivation}).  

\begin{equation}\label{eta_detivation}
     <\gls*{traj}_1, \gls*{traj}_2> = \int \sum_{n=1}^k \gls*{params} \gls*{traj}_1(t)^{(n)}\gls*{traj}_2(t)^{(n)} dt.  
\end{equation}The primary objective of $\Theta$ is to distinguish the norm~(\ref{com_operator}) and the inner product~(\ref{eta_detivation}) from the L2 norm~\cite{zucker2013chomp}.

\subsection{B-spline based Trajectory Generation}

$\gls*{d}^{th}$ order B-spline can be defined for a given knot sequence $p_{k} = \{ t_0, t_1, ..., t_{n_k} \}$ and control points $p_{c} = \{\gls*{p}_0, \gls*{p}_1,..., \gls*{p}_{n_p}\}$, where $t_* \in \mathbb{R}$, $\gls*{p}_* \in \mathbb{R}^d$ and $n_k=n_p+d+1$. If d is set to 3, each $\gls*{p}_i$ represents position in $\mathbb{R}^3$, where $i=0,...,n_p$. For a given time index t, the corresponding position $\gls*{p}(t)$ can be fully determined by using  De-Boor-Cox formula~\cite{de1971subroutine}.
\begin{equation}\label{deboor}
    \gls*{p}(t) = De\-Boor\-Cox(t, p_{c}).
\end{equation}Estimation is not limited to the position; velocity, acceleration or any high order derivative of $p_{c}$ can be estimated using $De\-Boor\-Cox(t, p_c^{(*)})$ as given in Algorithm.~\ref{alg:deboor}, where $^{(*)}$ depicts the order of the derivative of $p_{c}$ such that $(*) < d$.

\begin{algorithm}
\caption{The B-spline trajectory (p) and its derivative estimation for a given time index t, where p equals $p_c^{(*)}$ }\label{alg:deboor}
\begin{algorithmic}[1]
\Procedure{DeBoorCox}{$t, p$}
\State $t = \left\{\begin{matrix}
   p_{k}[d], \quad if \: t < p_{k}[d] \\ 
   p_{k}[n_k], \quad if \: t > p_{k}[n_k] \\
   t, \quad otherwise
\end{matrix}\right.$
\State k = d
\While{$true$}
  \If{$p_{k}[k+1] \geq t$}
    \State break
  \EndIf
  \State k++
\EndWhile
\State $\gls*{p}_e$[d]
\For{$i \gets 0 \quad to \quad d$}
   \State $\gls*{p}_e[i] \gets p[k-d+i]$
\EndFor
 
\For{$r \gets 1 \quad to \quad d$} 
    \For{$i \gets d \quad to \quad r$}
        \State $\beta \gets \frac{t -p_{k}[i+k-d]}{p_{k}[i+1+k-r]-p_{k}[i+k-d]}$
        \State $\gls*{p}_e[i] \gets (1-\beta) \times \gls*{p}_e[i-1] + \beta \times \gls*{p}_e[i]$
    \EndFor
\EndFor
\State return $\gls*{p}_e[d]$
\EndProcedure
\end{algorithmic} 
\end{algorithm}

Later, the B-spline matrix representation was proposed by Qin~\cite{qin2000general}. B-spline can be formulated as uniform or non-uniform. J. Hu et al.~\cite{hu2019real} detailed the uniform B-spline matrix representation. In uniform B-spline, knot span is the same for any considered consecutive time interval, i.e., $\Delta t = t_{i+1}-t_i, \; i \in [0, n_k)$. Any position of the trajectory can be parameterized by considering only d+1 consecutive control points: $[\gls*{p}_i, \gls*{p}_{i+1}, ..., \gls*{p}_{i+d}]$. Hence, corresponding normalized time $q(t)$ can be calculated as follows:
\begin{equation}
q(t) = \frac{t-t_i}{t_{i+1}-t_i} = \frac{t-t_i}{\Delta t}, \quad t \in [t_i, t_{i+1}].
\end{equation} In the matrix representation, $c(q(t))$, which is given in~(\ref{deboor}), can be determined by:
\begin{equation}
\begin{aligned}
    c(q(t)) =  \mathbf{q}(t)M_{d}p_i, \quad \mathbf{q}(t) = [1, q(t), q^2(t), ..., q^{d}(t)]^T, \; p_i = [\gls*{p}_{i}, \gls*{p}_{i+1}, ..., \gls*{p}_{i+d}]^T,\\
    M_d \in \mathbb{R}^{d+1\times d+1},  \quad M_{r,c} = \frac{1}{d!}\binom{d}{d-r} \Sigma _{s=c}^{d} (-1)^{s-c} \times \binom{d}{s-c} (d-s)^{d+1-r-s}.
    \end{aligned}
\end{equation}Since each control point $\gls*{p}_i$ belongs to $d+1$ of successive spans, B-spline can be controlled locally. Due to such controllability, b-spline is suitable for local trajectory planning~\cite{usenko2017real}. Moreover, the derivatives of a given B-spline are also B-spline~\cite{kulathunga2022optimization}. Hence, B-spline's derivatives (e.g., velocity, acceleration, jerk) can be calculated considering corresponding span $[t_i, t_i+1)$ for a given d+1 consecutive control points $\gls*{poly}_i = [\gls*{p}_{i}, \gls*{p}_{i+1}, ..., \gls*{p}_{i+d}]^T \in \mathbb{R}^{d\times3}$ and corresponding knot vector.
\begin{equation}
\begin{aligned}
    \begin{aligned}
    \frac{dc(q(t))}{du} = \frac{1}{(\Delta t)} b_1M_d\mathbf{v_i}^T, \quad b_1 = [0, 1, u, ..., u^{d-1}] \in \mathbb{R}^{d+1}, \\
    \frac{d^2c(q(t))}{d^2u} = \frac{1}{(\Delta t^2)} b_2M_d\mathbf{v_i}^T, \quad b_2 = [0,0, 1, u, ..., u^{d-2}] \in \mathbb{R}^{d+1}.
    \end{aligned}
\end{aligned}
\end{equation}The explicit form of estimation of velocity and acceleration of a given time index is calculated as follows: \begin{equation}\label{eq:vel_acce}
\begin{aligned}
    \frac{dc(q(t))}{du} = d \cdot \frac{p_{c}(i+1) - p_{c}(i)}{p_{k}(i+d+1) - p_{k}(i+1)}, \\
   \frac{d^2c(q(t))}{d^2u} = \\(d^2 -d)\cdot(\frac{p_{c}(i+2) - p_{c}(i+1)}{p_{k}(i+d+2) - p_{k}(i+2)}  -  \frac{p_{c}(i+1) - p_{c}(i)}{p_{k}(i+d+1) - p_{k}(i+1)} ).
\end{aligned}
\end{equation} In most of the situations, initial control points are generated as explained in section~\ref{sec:initial_waypoints_generation}. Such methods may or may be not smooth enough for initial trajectory generation. There are various ways to generate intermediate waypoints to improve the quality of the trajectory using B-splines. For example, the initial trajectory was constructed using cubic B-Spline in~\cite{mpc_mc}. Such a capability is mainly due to B-spline's properties. 

It is particularly continuity and convex-hall properties that make B-spline trajectory generation such a robust technique.

% Let $p^{init}$ and $p_{c}$ be control points before and after interpolation respectively. The relationship between $p^{init}$ and $p_{c}$ is given by
% \begin{equation}\label{eq:control_points}. 
%     \frac{1}{6}\begin{bmatrix}
% 1 & 4 &  1  & \hdots & 0\\ 
%  0&  1& 4    & \hdots & 0\\ 
% %  0& 0 &1  &4  &1  & \hdots  &0 \\ 
%  \vdots &  \vdots  & \vdots   &  \ddots & 0 \\ 
%  0& 0 & 1   &  4& 1\\ 
% \end{bmatrix}p_{c} = p^{init}
% \end{equation} in which $p_{c}$ can be determined with the help of Householder QR decomposition~\cite{dietrich1976new}.

\subsubsection{Convex Hull Property}\label{sec:convex_hull}
\begin{figure}[!ht]
\begin{center}
\includegraphics[width=\textwidth]{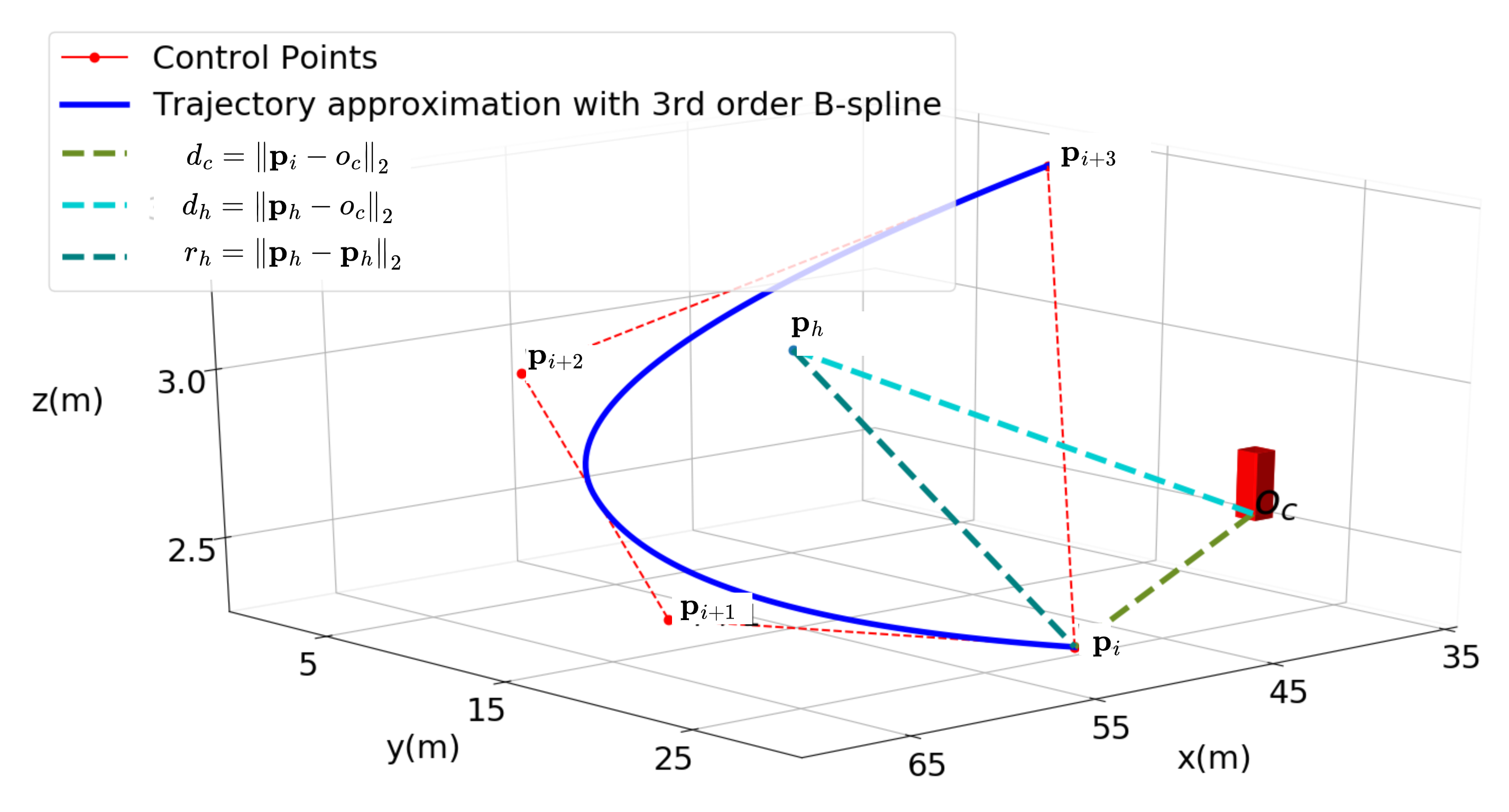}
\caption{Showing the B-spline convex-hull property. Convex hull, which comprises consecutive control points, e.g., $\gls*{p}_i, \gls*{p}_{i+1}, \gls*{p}_{i+2}$ and $\gls*{p}_{i+3}$, always belongs to obstacle-free space if the preceding control points satisfy the inequality~(\ref{eq:convex_hull})}
\label{fig:covex_hull}
\end{center}
\end{figure} Among the properties of the B-spline, the convex hull property is the most significant property due to its capabilities for checking the dynamical feasibility and the collision. How convex hull property is incorporated for calculating dynamical feasibility is given in~(\ref{eq:vel_acce}). As shown in Fig.~\ref{fig:covex_hull}, $d_h >0$ and $d_h > d_c - r_h$ should be held for a considered point in the trajectory to ensure a collision-free trajectory, where $d_c$ is the distance between a given control point and its closest obstacle position.  In $d$th order B-spline, a convex hull is formed by connecting any successive $d+1$ control points, e.g., $\gls*{p}_i, \gls*{p}_{i+1}, \gls*{p}_{i+2}, ..., \gls*{p}_{i+{d}}$ or union of all consecutive control points that lie on the corresponding b-spline curve~\cite{gordon1974b}. Moreover, $r_h$ can be substituted with $d_{i,i+1}+d_{i+1, i+2}+d_{i+2, i+3}$ since $r_h \leq d_{i,i+1}+d_{i+1, i+2}+d_{i+2, i+3}$, $d_h > d_c - (d_{i,i+1}+d_{i+1, i+2}+d_{i+2, i+3})$, where $d_{i,i+1} = \left \| \gls*{p}_{i+1} - \gls*{p}_{i} \right \|$, $d_{i+1,i+2} = \left \| \gls*{p}_{i+2} - \gls*{p}_{i+1} \right \|$ and $d_{i+2,i+3} = \left \| \gls*{p}_{i+4} - \gls*{p}_{i+3} \right \|$. As mentioned in~\cite{zhou2019robust}, the following condition should hold for collision-free trajectory planning:
\begin{equation}\label{eq:convex_hull}
    d_{i, i+1} < \frac{d_c}{3}, \quad d_c > 0, \quad i \in \{1,2,3\}.
\end{equation}

\subsubsection{Continuity}
B-spline-based trajectory generation has several advantages over the piece-wise-based trajectory generation~\cite{richter2016polynomial, oleynikova2016continuous}. The boundary constraints are to be satisfied explicitly to guarantee the continuity of a piece-wise trajectory. In such a trajectory, the smoothness of the trajectory solely depends on the way control points are formed. On the other hand, boundary constraints can be neglected since the whole trajectory can be treated as one segment in B-spline-based trajectory generation. Moreover, the B-spline-based trajectory can be controlled locally, as explained in section~\ref{sec:convex_hull}, without affecting the rest of the trajectory.

\begin{figure}[!ht]
\begin{center}
\includegraphics[width=\textwidth]{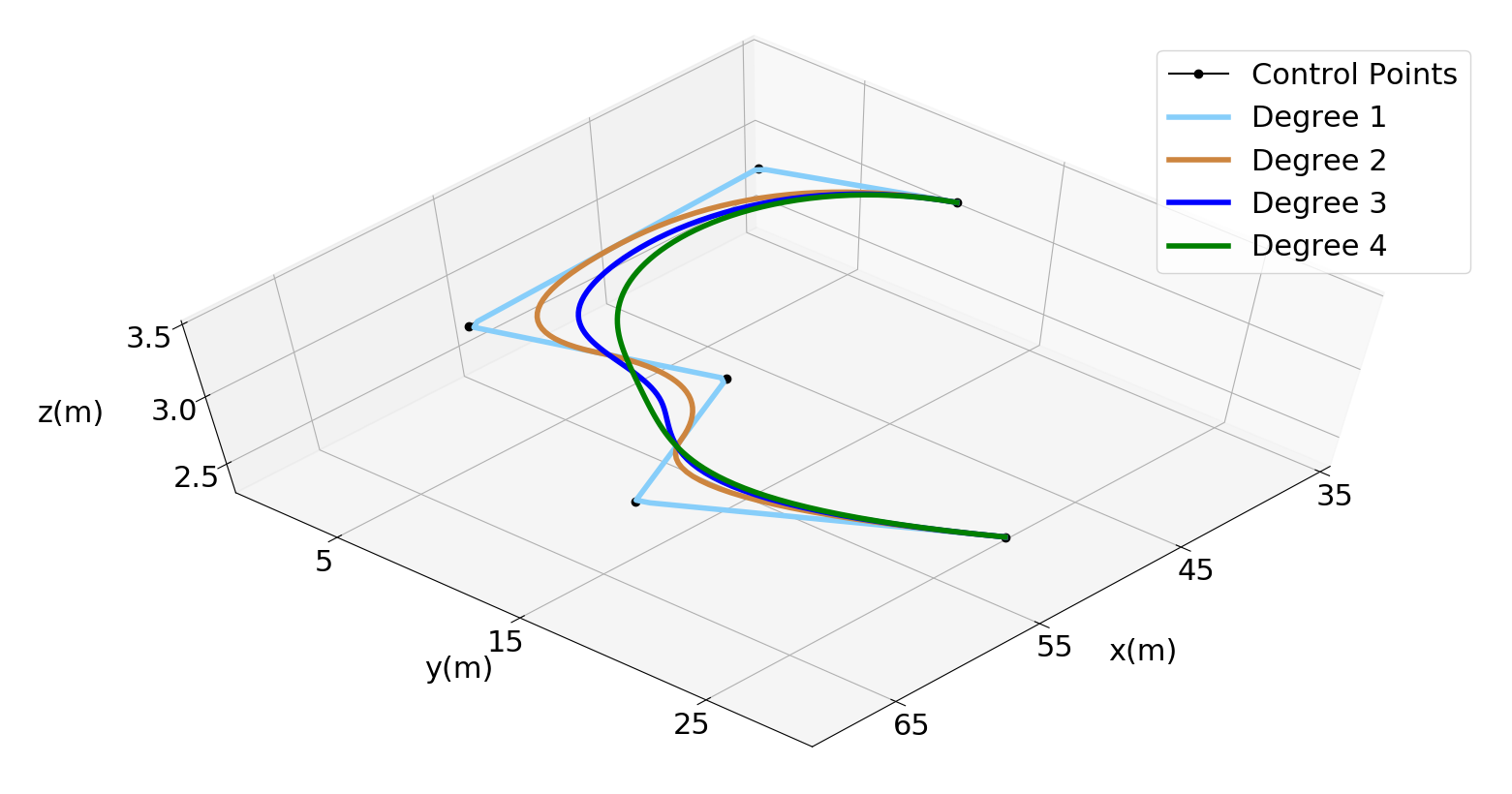}
\caption{Trajectory generation using uniform B-spline. The smoothness of the curve is dependent on the degree of the B-spline. The trajectory passes precisely through the given control points at the degree equal to 1, as depicted in light blue color. The smoothness of the trajectory increases with the order of the B-spline}
\label{f:trajectory_generation_bspline}
\end{center}
\end{figure}

\subsection{Bernstein Piecewise Trajectory Generation}

Bernstein polynomial is a specific form of B-spline, which is similar to the Bezier curve~\cite{flores2008real, preiss2017downwash}. Bernstein polynomial can be described as follows:
\begin{equation}
    \begin{aligned}
       \gls*{traj}_j(t) = \gls*{polyc}_j^0p_{d}^0(t) + \gls*{polyc}_j^1p_{d}^1(t) + ... + \gls*{polyc}_j^{d} p_{d}^{d}(t) = \Sigma_{i=0}^{d} \gls*{polyc}^{i}_jp_{d}^{i}(t), \\ \quad p_{d}^i(t) = \binom{d}{i}\cdot
t^i\cdot(1-t)^{d-i},
    \end{aligned}
\end{equation} where $d$ is the degree of the polynomial, $\gls*{polyc}_j^0, \gls*{polyc}_j^1,..., \gls*{polyc}_j^{d}$ are the control points of jth polynomial segment and $t \in [0,1]$. Since Bezier is a particular form of B-spline curve, such curves hold convex hull property. Hence, given a sequence of control points, a constrained convex hull can be defined using the control points that are considered. Both the beginning and end of the curve are determined by the first and the last control points, respectively. Further, the derivative of Bezier is also a Bezier curve. 
 \begin{equation}
     \Gamma_\mu (t) = \left\{\begin{matrix}
 s_1 \cdot \Sigma_{i=0}^{d} \gls*{polyc}^i_{1,\mu }p_{d}^i(\frac{t-t_0}{s_1})& t_0 \leq t < t_1 \\ 
 s_2 \cdot \Sigma_{i=0}^{d} \gls*{polyc}^i_{2,\mu }p_{d}^i(\frac{t-t_1}{s_2})& t_1 \leq t < t_2 \\  
 \vdots & \\ 
 s_m \cdot \Sigma_{i=0}^{d} \gls*{polyc}^i_{m_d,\mu }p_{d}^i(\frac{t-t_{{m_d}-1}}{s_{m_d}})& t_{m_d-1} \leq t < t_{m_d} \\  
\end{matrix}\right.,
 \end{equation}   
 where i, j refer to $i^{th}$ control point in $j^{th}$ segment, i.e., $\gls*{polyc}_j^i$, $s_j$ is a scaling factor of $j^{th}$ segment for mapping time duration from $[0\;, 1]$ to $[t_{j-1}, \; t_j]$ and $\mu \in \{x,y,z\}$. Once $\Gamma_\mu(t)$ is obtained, the objective is to minimize the total cost, which can be determined by taking the integral of square error up to $k_r$ order as given in~(\ref{eq:only_jerk}). Such a problem can be formulated as a \gls*{QP} constraint problem. For instance, Gao and Wu~\cite{gao2018online} proposed a Bernstein-based trajectory optimization approach in which three types of constraints piecewise trajectory continuity, safety constraints which are based on convex hull property, and dynamical feasibility constraints enforced~\cite{gao2018online}.  
 
\subsection{Comparison of several trajectory techniques}
\begin{figure}[!ht]
\begin{center}
\includegraphics[width=\textwidth]{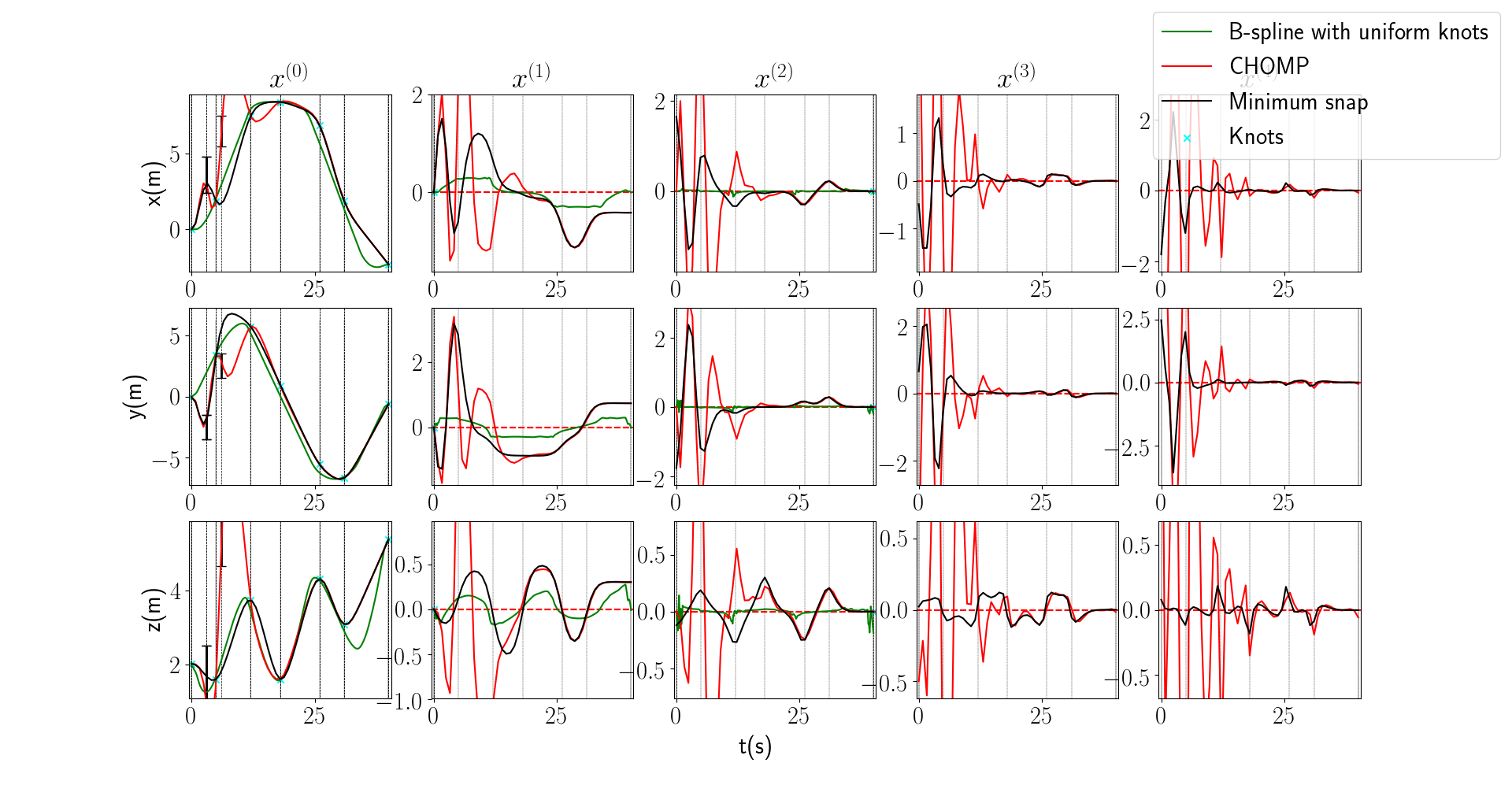}
\caption{Changes of position, velocity, acceleration, jerk, and snap profiles over time for the provided control points sequence and knot vector (\ref{dataset_trajectory})}
\label{f:trajectory_comparison}
\end{center}
\end{figure}

In the preceding subsections, several types of trajectory parameterization techniques were considered. We have selected three different types of trajectory parameterization techniques for this comparison: piecewise-polynomials technique, fitting based on a sequence of points, and the third is uniform B-spline-based technique. The objective of piecewise-polynomials is to find optimal polynomial coefficients~\cite{mellinger2011minimum} or end-derivatives~\cite{richter2016polynomial} of consecutive segments, whereas the objective of the third technique is to find a set of points satisfying the provided constraints~\cite{ratliff2009chomp}. A comparison of how velocity, acceleration, jerk, and snap are varied for selected techniques in terms of mean, standard deviation (std), min and max for the same a set of control points and knot vector is present below. Considered knot vector  and control points are 
\begin{equation}\label{dataset_trajectory}
\begin{aligned}
   p_{ctrl} = [[0.011, -0.0329,  2.017  ], [ 1.867,   3.408, 1.6], [ 7.514, 5.715, 3.735], \\ [ 8.410,  0.911,  1.600], [ 6.902, -5.531, 4.306],[ 1.899,-6.680,3.082 ],\\ [-2.302, -0.611, 5.375]] \\
   p_{knot} = [0.0, 5.0, 12.0, 18.0, 26.0, 31.0, 40]
\end{aligned}
\end{equation} Each approach has its own set of parameters to fine-tune
for obtaining an optimal trajectory. The generated trajectories are shown in Fig.~\ref{f:trajectory_generation_bspline} with different configuration setup (with different parameter set). Fig.~\ref{f:trajectory_comparison} shows how the derivatives up to the 4th change over time on each direction, i.e.,x,y,z, separately for each technique. When looking at the derivatives of each method, it is clear that smoothness, which is the main point to be considered for motion planning, is higher in both B-spline and Minimum-snap compared to CHOMP. Since uniform B-spline is used in this comparison, smoothness changes of each derivative between B-spline and Minimum-snap can not be compared directly due to time allocation when generating the trajectories. Hence, Minimum-snap trajectory smoothness can be changed, optimizing the time allocation process~\cite{richter2016polynomial}. On the contrary, such a time allocation process is not necessary for uniform B-spline. Yet control points are interpolated appropriately to generate a continuous and smooth trajectory. 

We varied the parameter set of each approach appropriately and estimated mean, std, max, min of velocity, acceleration, jerk, and snap profile; the result is given in Table.~\ref{table:parameter_comparison}. The results clearly indicate that the consistency of the trajectory depends on the parameters that are used to parameterize the trajectory. Hence, appropriate parameter set selection for a given task is of utmost importance, which can be seen by looking at the statistical properties (mean, std, min, and max) of higher-order derivatives, e.g., velocity, acceleration, jerk, and snap. As described in the previous paragraph, the time allocation process directly affects the parameter selection of Minimum-snap. Further, the optimal polynomial coefficients process depends on time allocation as given in~(\ref{eq:minimum_snap_total_time}). On the other hand, Poly-traj~\cite{richter2016polynomial} generation process has fewer parameters to be optimized since it uses free end-derivatives of each segment. Hence, the latter technique is faster than Minimum-snap.  

\begin{figure}[!ht]
\begin{center}
\includegraphics[width=\textwidth]{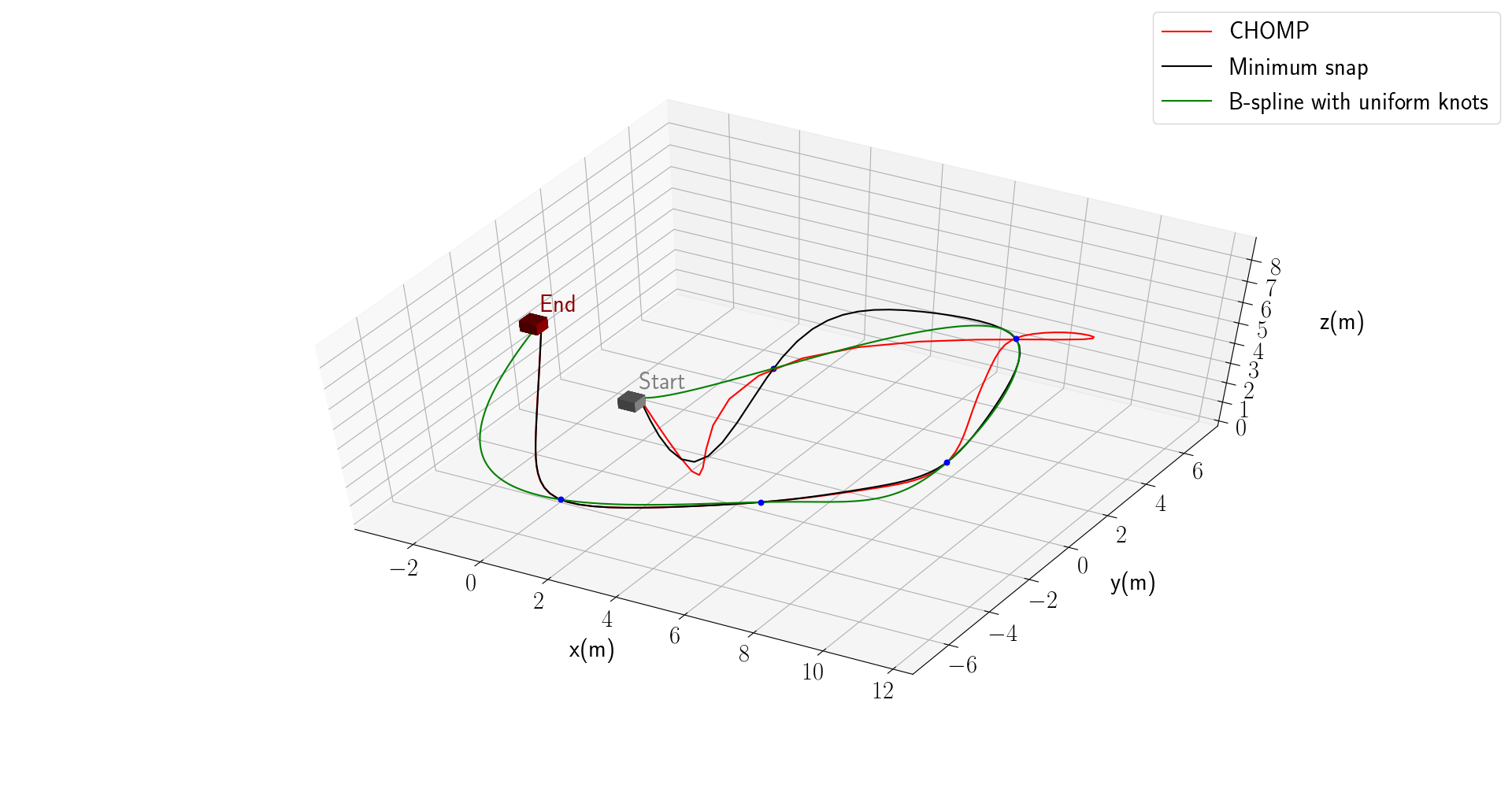}
\caption{Generated trajectories using three different approaches for a given sequence of control points and knot vector (\ref{dataset_trajectory})}
\label{f:trajectory_comparison_3d}
\end{center}
\end{figure}

% \usepackage{multirow}
\begin{table}[!ht]
\tiny
\caption{Velocity, acceleration, jerk, and snap profile for generating an optimal trajectory for a given set of knot vector and control points (Fig.~\ref{f:trajectory_comparison}) using three different techniques: Minimum-snap~\cite{mellinger2011minimum}, Poly-traj~\cite{richter2016polynomial}, and CHOMP~\cite{ratliff2009chomp}}\label{table:parameter_comparison}
\begin{center}
 
\begin{tabular}{|l|l|l|l|l|l|l|l|l|}
\hline

\multirow{2}{*}{Type} & \multicolumn{4}{c|}{Velocity} & \multicolumn{4}{c|}{Acceleration} \\ \cline{2-9} 
                      & mean   & std   & min   & max  & mean    & std    & min    & max   \\ \hline
                    
Poly-traj, d: 8,  mc: 2&0.0058& 1.0154& -1.4545& 3.9179& 0.0056& 0.9051& -2.835& 3.6449\\ \hline
Poly-traj, d: 8,  mc: 6&0.0& 0.0& 0.0& 0.0& 0.0& 0.0& 0.0& 0.0 \\ \hline
Poly-traj, d: 6,  mc: 4&0.006& 1.0708& -1.7716& 3.7864& 0.0043& 0.9307& -2.7987& 3.6032\\ \hline
Poly-traj, d: 8,  mc: 4&0.0059& 1.0299& -1.4728& 3.934& 0.0053& 0.9131& -2.9157& 3.5214 \\ \hline
Poly-traj, d: 10,  mc: 4&0.0058& 1.0057& -1.4428& 3.9213& 0.0052& 0.8918& -2.7541& 3.631 \\ \hline
\begin{tabular}[c]{@{}l@{}}Minimum-snap, d: 8,\\ mc: 2\end{tabular}  &0.1258& 1.2154& -1.4345& 3.1259& 0.0676& 0.1259& -2.2874& 3.3278\\ \hline
\begin{tabular}[c]{@{}l@{}}Minimum-snap, d: 8,\\ mc: 6\end{tabular}&0.0045& 0.0094& -0.07& 0.019& 0.09& 0.0097& -0.0098& 0.0014 \\ \hline
\begin{tabular}[c]{@{}l@{}}Minimum-snap, d: 6,\\ mc: 4\end{tabular}&0.0689& 1.0009& -1.3416& 3.2388& 0.0012& 0.4584& -2.3189& 3.2185\\ \hline
\begin{tabular}[c]{@{}l@{}}Minimum-snap, d: 8,\\ mc: 4\end{tabular}&0.0015& 1.0412& -1.3215& 3.7543& 0.0075& 0.8763& -2.5487& 3.3215 \\ \hline
\begin{tabular}[c]{@{}l@{}}Minimum-snap, d: 10,\\ mc: 4\end{tabular}&0.0036& 1.0006& -1.3428& 3.7832& 0.0099& 0.4378& -2.4548& 3.4893\\ \hline
CHOMP, pd: 3&0.0068& 0.6421& -0.9522& 1.7255& 0.0045& 0.3876& -1.131& 1.476 \\ \hline
CHOMP, pd: 5&0.0065& 0.644& -0.9634& 1.7161& 0.0044& 0.3909& -1.1082& 1.4418 \\ \hline
CHOMP, pd: 7&0.0064& 0.6443& -0.966& 1.7105& 0.0043& 0.3916& -1.0951& 1.4205 \\ \hline
\multirow{2}{*}{Type} & \multicolumn{4}{c|}{Jerk}     & \multicolumn{4}{c|}{Snap}         \\ \cline{2-9} 
                      & mean   & std   & min   & max  & mean    & std    & min    & max   \\ \hline
Poly-traj, d: 8,  mc: 2&  0.007& 1.2544& -4.8056& 3.9318& -0.0151& 2.3178& -9.8029& 6.9483 \\ \hline
Poly-traj, d: 8,  mc: 6& 0.0& 0.0& 0.0& 0.0& 0.0& 0.0& 0.0& 0.0  \\ \hline
Poly-traj, d: 6,  mc: 4&0.0117& 1.568& -5.5746& 5.7423& -0.1288& 3.5271& -13.4562& 10.2578 \\ \hline
Poly-traj, d: 8,  mc: 4& -0.0021& 1.2562& -4.7192& 3.7562& 0.0131& 1.9593& -7.7131& 6.0049 \\ \hline    
Poly-traj, d: 10,  mc: 4& 0.0074& 1.3399& -5.5769& 4.409& -0.0504& 3.1073& -12.3429& 9.9933 \\ \hline             
\begin{tabular}[c]{@{}l@{}}Minimum-snap, d: 8,\\ mc: 2\end{tabular}&  0.0006& 1.1125& -4.3413& 3.5153& -0.0042& 2.1383& -9.0056& 6.3418 \\ \hline
\begin{tabular}[c]{@{}l@{}}Minimum-snap, d: 8,\\ mc: 6\end{tabular}& 0.0005& 0.0004& -0.0007& 0.0089& 0.0005& 0.004& -0.0008& 0.0009  \\ \hline
\begin{tabular}[c]{@{}l@{}}Minimum-snap, d: 6,\\ mc: 4\end{tabular}&0.01& 1.3456& -5.2167& 5.321& -0.0093& 3.214& -12.5124& 9.2134 \\ \hline
\begin{tabular}[c]{@{}l@{}}Minimum-snap, d: 8,\\ mc: 4\end{tabular}& -0.001& 1.1321& -3.7192& 3.3217& 0.0093& 1.2145& -5.6527& 4.7854 \\ \hline    
\begin{tabular}[c]{@{}l@{}}Minimum-snap, d: 10,\\ mc: 4\end{tabular}& 0.0009& 1.2145& -3.9987& 3.9983& -0.0067& 2.8731& -10.7653& 8.8416 \\ \hline  
CHOMP, pd: 3&0.0021& 0.3643& -1.2594& 1.1584& -0.0014& 0.4239& -1.8326& 1.5425 \\ \hline
CHOMP, pd: 5&0.0023& 0.3628& -1.2553& 1.1639& 0.0005& 0.4241& -1.8526& 1.6243 \\ \hline
CHOMP, pd: 7&0.0022& 0.3614& -1.2732& 1.1769& 0.0021& 0.4247& -1.7462& 1.5906 \\ \hline
\end{tabular}
\begin{tablenotes}
d: order of the polynomial, mc: maximum continuity or maximum continuity order in between consecutive segments, pd: number of proposed points or point density per defined time duration of the trajectory
\end{tablenotes}\label{tab:traj_execution}
\end{center}
\end{table}

\section{Free Space Extraction}\label{sec:free_space_extraction}
Obstacle region identification is of utmost essential for optimal trajectory planning in real-time. In a cluttered environment, the way the trajectory planning problem formulated is matters for fast reaction. Such trajectory planning approaches can be designed as~\gls*{QP} mainly due to less computation power required for such tasks. Hence, forming obstacles-free regions in the form of convex has more advantages in terms of reducing the computation power, simplicity and fast convergence. Chen~\cite{chen2015real} attempted to define free space as a series of cubes between the start and goal pose. Thenceforth, OctoMap~\cite{hornung2013octomap} was used for constructing the map surrounding the quadrotor, where regions with no obstacles are considered free spaces. After obtaining the free space information, obstacle constraints are enforced into~(\ref{eq:only_jerk}) to generate optimal trajectory.

Let $C = [c^m_1, c^m_2,...]$ be a set of consecutive grids within the OctoMap and corresponding free space regions be $C_{free} = [c^f_1, c^f_2,...]$. Both $c^m_i$ and $c^f_i$ were defined as cubes, each of which is described by \begin{equation}\label{eq:window_est}
\begin{aligned}
    c^m_i = [\underbrace{c^m_{i_{x_0}}, c^m_{i_{y_0}}, c^m_{i_{z_0}}}_{l^i_m}, \underbrace{c^m_{i_{x_1}}, c^m_{i_{y_1}}, c^m_{i_{z_1}}}_{u^i_m}], \quad
    c^f_i = [\underbrace{c^f_{i_{x_0}}, c^f_{i_{y_0}}, c^f_{i_{z_0}}}_{l^i_f}, \underbrace{c^f_{i_{x_1}}, c^f_{i_{y_1}}, c^f_{i_{z_1}}}_{u^i_f}], 
\end{aligned}
\end{equation} Once $C_{free}$ was obtained, free space regions can be considered as a set of inequality constraints that can be added into the piece-wise polynomials trajectory generation as $l^i_f \leq \Gamma_{T}(t_i) \leq u^i_f$, where $i=1,...,m_d-1$ and $\Gamma_{T}$ was defined~(\ref{eq:minimum_snap_total_time}). In such a trajectory, additional boundary constraints should be introduced if the extrema of $dth$ order polynomial violates the boundary constraints corresponding to each axis, i.e., x, y and z in each segment~\cite[ eq.10]{chen2015real}. Similar to the preceding approach, Gao and Shen~\cite{gao2016online} proposed a sequence of spheres to represent free space from the initial position to the final position. To construct the environment, a map was not built; instead, they bypassed map building by constructing a KD-tree~\cite{bentley1975multidimensional} based placeholder to store raw point cloud for the LiDAR. Afterwards, a relative map to the current pose of the \gls*{MAV} was retrieved using nearest neighbour search; \gls*{RRG}~\cite{kala2013rapidly} combined with A* was used to find a flight corridor or intermediate waypoints. Such intermediate waypoints were connected by overlapping spheres.

% \begin{equation}
%     \left\{\begin{matrix}
% \Gamma_i^{k+1}(t_{k}) \leq c_i^{l_1}-\delta_p & if \; \Gamma_i^k(t_{k})>l_m^i\\ 
% \Gamma_i^{k+1}(t_{k}) \geq c_i^{l_1}+\delta_p & if \; \Gamma_i^k(t_{k})<u_m^i
% \end{matrix}\right.
% \end{equation} where $\Gamma_i^{.}$ is the trajectory which corresponds to ith segment. 

\gls*{IRIS}~\cite{deits2015computing} is one of the first successful ideas in which obstacle-free spaces are extracted using a convex optimization technique. In this proposed approach, initially, it is required to provide a seeking point and an area with a boundary box where an obstacle-free region is to be searched. Seeking point is defined as a unit ball: $\varepsilon (C, \gls*{p}_0) = \{ \gls*{p} = C\tilde{\gls*{p}} + \gls*{p}_0 \;|\; \left \| \tilde{\gls*{p}} \right \|_2 \leq 1 \}$, where $\gls*{p}_0$ is the center point. The linear constraints, which separate the boundary box into obstacle-free and obstacle-rich regions, are defined as a set of hyper-planes: $P = \{\gls*{p} \; |\; A\gls*{p} \leq b\}$. Subsequently, finding the optimal representation of $\varepsilon (C, \gls*{p}_0)$ and $P$ with respect to given obstacles, $\imath_j, j=1,...,N$ is solved as an iterative process~(\ref{plance_popint}). 

\begin{equation}\label{plance_popint}
\begin{aligned}
 \min_{C, \gls*{p}_0, A, \mathbf{b}} \quad & -log(det C) \\
\textrm{s.t.} \quad & A_j^T\gls*{p}_k \geq \mathbf{b}_j \quad \forall \gls*{p}_k \in \imath_j, \quad j=1,...,N \\
  \quad & \underset{\left \| \tilde{\gls*{p}} \right \|}{sup} \; A_i^T (C\tilde{\gls*{p}}+ \gls*{p}_0) \leq \mathbf{b}_i \quad \forall i = 1,...,N ,\\
\end{aligned}
\end{equation} where $\gls*{hrep}_i$ and $\mathbf{b}_i$ correspond to ith row of A and $\mathbf{b}$. The first constraint, i.e., $A_j^T\gls*{p}_k \geq \mathbf{b}_j$, is imposed to move the obstacle into one side of the plane, $A_j^T\gls*{p} = \mathbf{b}_j$, whereas the second constraint, i.e., $\underset{\left \| \tilde{\gls*{p}} \right \|}{sup} \; A_i^T (C\tilde{\gls*{p}}+ \gls*{p}_0) \leq \mathbf{b}_i$, ensures the ellipsoid lies on the other side of the plane. The researchers proposed to solve the (\ref{plance_popint}) as a two-step process: searching, first, for proper constraints (i.e., $A_i$ and $\mathbf{b}_i$) and then the maximum volume that satisfies ellipsoid, ensuring preceding constraints. In other words, they attempted to find hyperplanes that separate obstacle regions and free regions. Conceptually,  hyperplane separation is done by finding planes that intersect with obstacle boundaries. Afterwards, the ellipsoid is uniformly expanded until it intersects with obstacle boundaries. Let $\alpha$ be the scaling factor which defines the expansion. Let $ \varepsilon_{\alpha} = \{ C\tilde{\gls*{p}} + \gls*{p}_0 \;|\; \left \| \tilde{\gls*{p}} \right \|_2 \leq \alpha \}$ for $\alpha \geq 1$ be the expanded ellipsoid. Hence, the optimal $\alpha^*$ can be determined by 
\begin{equation}
\begin{aligned}
   \alpha^* = \underset{\alpha}{arg \; min} \\
   \textrm{s.t.} \quad & \varepsilon_{\alpha} \cap \imath_j \neq \varnothing  
\end{aligned}
\end{equation}
After finding $\alpha^*$, it is possible to define the optimal inscribed ellipsoid ($\varepsilon^*$), which is the obstacle-free region~\cite[sec.3.3]{deits2015computing}. 

Sikang et al.~\cite{liu2017planning} proposed a new, quite different from the aforesaid~\gls*{IRIS}, approach for extracting obstacle-free regions as a convex set~\gls*{SFC}. \gls*{SFC} searches a set of overlapping polyhedra from the start pose to the goal pose. To get intermediate obstacle-free positions, the researchers  utilized a graph search technique, namely \gls*{JPS}~\cite{harabor2011online}. The main reason for selecting~\gls*{JPS} over sampling-based algorithms (e.g., RRT* and \gls*{PRM}) or search-based techniques such as A* or Dijkstra is due to the nature of~\gls*{JPS}; it uses uniform-cost grid map with uniform voxels. In general, sampling-based techniques are not deterministic though probabilistically complete. Thus, there is no guarantee about the duration of searching time. On the other hand, the computational time for search-based methods is pretty high if the environment is cluttered. However, \gls*{JPS} has a lower searching time compared to A*~\cite{liu2017planning}. Let $p_c = {\gls*{p}_0, \gls*{p}_1, ..., \gls*{p}_n}$ be the intermediate waypoints from start to goal pose and $l_i = <\gls*{p}_i, \gls*{p}_{i+1}>$ be the $i^{th}$ line segment, where $i=1,...,n-1$. Each line segment constitutes convex polyhedra, namely, $E_i$. Along with that, \gls*{SFC} can be expressed as $SFC(P) = \{E_i \; | \; i=0,...,n-1\}$. \gls*{SFC} has two steps: finding $E_i$ that fits the $l_i$ and seeking a set of linear inequalities that are tangent to $E_i$. Let $E_i$ be $\varepsilon_i (C_i, \gls*{p}_i^0) = \{ \gls*{p} = C_i\tilde{\gls*{p}} + \gls*{p}_i^0 \;|\; \left \| \tilde{\gls*{p}} \right \|_2 \leq 1 \}$. In $\mathbb{R}^3$, $C_i$ can be decomposed as $R^TSR$, where R gives the axis of rotation between considered line segment in between $\gls*{p}_i$ and $\gls*{p}_{i+1}$). Semi-axis of $E_i$ is given by $S = diag(a,b,c)$ as a diagonal matrix. $\gls*{p}_i^0$ is the center of $l_i$. The objective of \gls*{SFC} is to find each pair $E_i$ and $\gls*{p}_0^i$, given the $l_i$ and obstacles set ($Obs_i$), which touches the $E_i$. 

\begin{figure}[!ht]
\begin{center}
\includegraphics[width=\textwidth]{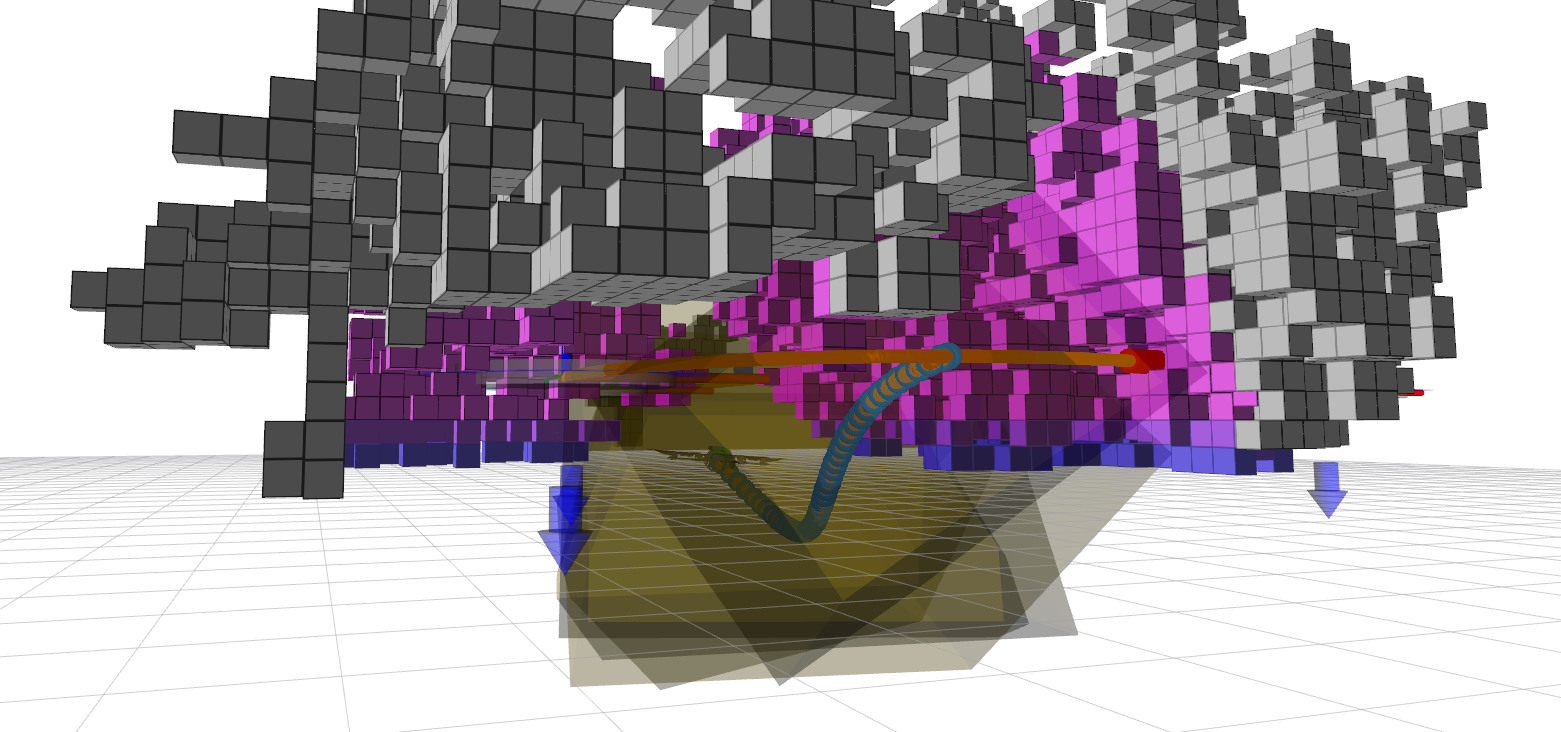}
\caption{Free space extraction using~\gls*{SFC}. Once intermediate initial waypoints are defined,~\gls*{SFC} calculates free space along the path, which is constructed from the initial waypoints}
\label{f:free_space_extraction}
\end{center}
\end{figure}

Initially, ellipsoids are spheres whose center poses are located as the midpoints of $l_i, \; i=1,...,n-1$. Afterwards, semi-axes except for the axis along $\gls*{p}_{i+1}-\gls*{p}_{i}$, are shrunk until the corresponding ellipsoid contains no obstacles. Let $\varepsilon _i^*(C_i, \gls*{p}_i^0)$ be the ith ellipsoid after applying the shrinking process. $\gls*{p}_j$ denotes the closest point that touches the $\varepsilon_i^*(C_i, \gls*{p}_i^0)$, where $j=1,...,m$ and m is the number of obstacles. Hence, corresponding half-space $H_j = \{ \gls*{p}_j \; | \; a_j^T\gls*{p}_j < \mathbf{b}_j \}$ is defined as a plane that is tangential to $\varepsilon^*_i(C, \gls*{p}_0)$, where $a_j$ and $\mathbf{b}_j$ are determined by:
\begin{equation}
    \mathbf{a}_j = \frac{d\varepsilon_i}{dp}_{\gls*{p}=\gls*{p}_j} = 2C_i^{-1}C_i^{-T}(\gls*{p}_j-\gls*{p}_i^0), \quad \mathbf{b}_j = \mathbf{a}_j^T\gls*{p}_j.
\end{equation} Hence, the intersections of these m half spaces create a convex polyhedron $C= \cup_{j=0}^m H_j = \{ \gls*{p} \;|\; A^T\gls*{p} < \mathbf{b} \}$. The same approach is applied to each line segment, $l_i$ in which we can generate each $C_i$. All in all, $SFC(P) = \{C\;|\; i=0,...,n-1\}$ can be constructed. A more descriptive formulation is in~\cite[Algorithm 1]{liu2017planning}.

\section{Continuous Trajectory Refinement}\label{sec:continuous_trajectory}
The objective function consists of several sub-objective functions: for improving the smoothness, for avoiding obstacles and so forth. In this section, a precise explanation is given on how to construct sub-objective functions for each of the various occasions. First, we examine the simplest case where only dynamic feasibility and obstacle avoidance constraints are taken into consideration. Let $J$ be the objective function or performance index
\begin{equation}
    J(\Gamma) =  \gls*{params}_{smooth}J_{smooth}(\Gamma) + \gls*{params}_{obs}J_{obs}(\Gamma).
\end{equation}There are various formulations of how $J_{obs}$ and $J_{smooth}$ are determined. In general, $J_{smooth}$ can be expressed as:
\begin{equation}
    \begin{aligned}
    J_{smooth}(\Gamma) = \frac{1}{2}\int_{0}^{1}\left \| \frac{d\Gamma(t)}{dt} \right \|^2dt.
    \end{aligned}
\end{equation} Eliminating unnecessary higher-order motion is the main objective of the $J_{smooth}$. On the other hand, $J_{obs}$ encourages to generate or modify collision-free trajectory by trying to push control points away from the obstacle zone if the trajectory is already in collisions or penalizing parts of the trajectory that is close to the obstacles. Let $B \subset \mathbb{R}^d$ be the exterior boundary of the \gls*{MAV} and $c$ is the cost function of penalizing close-in obstacles with respect to B. Along with that, $J_{obs}$ can be formulated as follows:
\begin{equation}
    \begin{aligned}
    J_{obs}(\Gamma) = \int_{0}^{1} \int_{u \in B}^{} c(f_c(\Gamma(t), \gls*{p}))\left \| \frac{d f_c(\Gamma(t), \gls*{p})}{dt} \right \|^2d\gls*{p} dt,
    \end{aligned}
\end{equation} where the function $f_c(\Gamma(t),\gls*{p})$, which was proposed by Ratliff at al.~\cite{ratliff2009chomp}, can be defined as follows:
\begin{equation}\label{obs_map}
    f_c(\Gamma(t),\gls*{p}) = \left\{\begin{matrix}
 -dis(\Gamma(t),\gls*{p})+\frac{1}{2}\delta_{dis} & if \: dis(\Gamma(t),\gls*{p}) < 0\\ 
 \frac{1}{2\delta_{dis}}(dis(\Gamma(t),\gls*{p})-\delta_{dis})^2 &  if \: 0 < dis(\Gamma(t),\gls*{p}) \leq \delta_{dis} \\ 
 0 & otherwise
\end{matrix}\right.,
\end{equation} where $\delta_{dis}$ denotes the distance from the boundary (B) of the quadrotor to a given obstacle position. Before taking gradient at i, $J(\Gamma)$ is linearized around i, $J(\Gamma) \approx J(\Gamma_i) + (\Gamma-\Gamma_i)^T \bigtriangledown J(\Gamma_i)$. Defining c and d is detailed in~\cite[eqs.(22-28)]{ratliff2009chomp}.

In~\cite{zhou2019robust}, the cost of the trajectory was estimated based on the following formulation:
\begin{equation}
     J(\Gamma) =  \gls*{params}_{obs}J_{obs}(\Gamma) + \gls*{params}_{smooth}J_{smooth}(\Gamma) + \gls*{params}_{soft}J_{soft}(\Gamma), \quad J_{soft}(\Gamma) = J_{v}(\Gamma) + J_{a}(\Gamma),
\end{equation} where $J_{soft}(\Gamma)$ is determined by soft limits on acceleration and velocity. $J_{smooth}(\Gamma)$ is defined by considering only geometric information without minimizing snap and/or jerk~\cite{mellinger2011minimum}. Such minimization is required because of the following stages of trajectory optimization. In such trajectory optimization, time reallocation has less impact on the objective function. Hence, $J_{smooth}(\Gamma)$ is defined as follows:
\begin{equation}
    J_{smooth}(\Gamma) = \Sigma_{i=d-1}^{n+1-d}  \left \| \underbrace{\gls*{p}_{i+1} - \gls*{p}_i }_{f_{i+1,i}}+ \underbrace{\gls*{p}_{i-1}-\gls*{p}_i}_{f_{i-1,i}} \right \|^2,
\end{equation} where a number of control points, denoted n, and $\mathbf{f}_{i+1,i}$ and $\mathbf{f}_{i-1,i}$ can be interpreted as connecting joint force of two springs between control points pairs: $(\gls*{p}_{i+1}, \gls*{p}_i)$ and $(\gls*{p}_{i-1}, \gls*{p}_i)$, for example, control points lie on a straight line if the sum of all terms equals zero. As aside, similar approaches were proposed in~\cite{zhu2015convex, quinlan1993elastic}.  

The value of $J_{obs}(\Gamma)$ is determined by calculating the distance to the closest object pose from each control point in which the distance to the obstacle, i.e., $f_c(\gls*{p}_i)$, is given by 
\begin{equation}
    f_c(\gls*{p}_i) = \left\{\begin{matrix}
(dis(\gls*{p}_i)-\delta)^2 & dis(\gls*{p}_i) \leq \delta_{dis}\\ 
0 & dis(\gls*{p}_i) > \delta_{dis}
\end{matrix}\right.,
\end{equation} where $\delta_{dis}$ is the free distance between~\gls*{MAV}'s center and the pose of the closest obstacle. Hence, $J_{obs}(\Gamma) = \Sigma_{i=d}^{n}  f_c(\gls*{p}_i)$ can be estimated based on a given trajectory in the form of control points. Soft constraints are defined by not exceeding both acceleration and velocity within those max limits. 
\begin{equation}
\begin{aligned}
 J_{\gls*{v}}(\Gamma) = \sum_\mu \sum _{i=d-1}^{n-d}f_v(\gls*{v}_{i,\mu}), \quad J_{a}(\Gamma) = \sum_\mu \sum _{i=d-2}^{k_d-d}f_a(\mathbf{a}_{i,\mu}) \\
  f(\gls*{v}) = \left\{\begin{matrix}
 (\gls*{v}_\mu^2 - \gls*{v}_{max}^2)^2 & \gls*{v}_\mu ^2 > \gls*{v}_{max}^2\\ 
 0 & \gls*{v}_\mu ^2 \leq \gls*{v}_{max}^2 
\end{matrix}\right. ,\quad f(\mathbf{a}) = \left\{\begin{matrix}
 (\mathbf{a}_\mu^2 - \mathbf{a}_{max}^2)^2 & \mathbf{a}_\mu ^2 > \mathbf{a}_{max}^2\\ 
 0 & \mathbf{a}_\mu ^2 \leq \mathbf{a}_{max}^2 
\end{matrix}\right.,
\end{aligned}
\end{equation}To calculate acceleration and velocity at each control point and when both acceleration and velocity exceed their maximum limits, convex hull property~(\ref{eq:convex_hull}) of b-spline is utilized  to penalize those control points. Based on the previous method,~\cite{usenko2017real} proposed an endpoint cost $J_{endpoint}(\Gamma)$, into the objective function as an additional term. The key intuition behind adding $J_{endpoint}(\Gamma)$ is to reduce the error between local trajectory and global trajectory since $J_{endpoint}(\Gamma)$ penalizes error of both velocity and position with respect to the desired global trajectory. $J_{endpoint}(\Gamma)$ is determined as follows:
\begin{equation}
    J_{endpoint}(\Gamma) =   J_{end}(\Gamma) =  \gls*{params}_{end}^p (\gls*{p}(t_{end})-\gls*{p}_{end})^2 + \gls*{params}_{end}^v(\dot{\gls*{p}}(t_{end})-\dot{\gls*{p}}_{end})^2,
\end{equation} where $\gls*{params}_{end}^p$ and $\gls*{params}_{end}^v$ are regularization parameters, whereas $\gls*{p}_{end}$ and $\dot{\gls*{p}}_{end}$ are desired end position and velocity of the trajectory. 

% Once waypoints are provided, the global trajectory is generated by using the selected technique, i.e., Minimum-snap or B-spline. The global trajectory is refined locally as a residing horizon passion. Let $\Gamma_0, \Gamma_i$ and $\Gamma_{i+1}$ be the initial, preceding and current trajectory within the residing horizon. 

\section{Receding Horizon Trajectory Planning}\label{sec:residing_horizon}

On most occasions, paths which are obtained by planning techniques are sub-optimal. Hence, the initial trajectory that is generated based on the initial path is to be refined, ensuring dynamic feasibility for controlling the \gls*{MAV}. Various approaches can be applied for trajectory refinement. However, enabling recursive feasibility, incorporating terminal constraints and convergence to the desired state are the utmost importance considerations to be contemplated throughout the process. \gls*{LQR} and \gls*{MPC} are the two most popular approaches that are being used for receding horizon planning. \gls*{LQR} is applied for linear systems, whereas \gls*{ILQR} and \gls*{DDP} are applied for non-linear system. Both in~\gls*{LQR} or \gls*{ILQR}, \gls*{OCP} is defined as an open-loop control problem. On the other hand, \gls*{MPC} is designed as a close-loop \gls*{OCP}. In other words, \gls*{OCP} is seeking actions knowing the behaviour of the surrounding environment.

\subsection{LQR based trajectory generation}\label{sec:lqr}

\gls*{DDP}~\cite{jacobson1970differential, theodorou2013information} is one of the first techniques proposed for solving optimal control problems. Let $\gls*{state}_{k+1} = \gls*{fdes}(\gls*{state}_k, \gls*{input}_k)$ be the discrete-time system dynamics; the total cost of the trajectory can be formulated for a given control policy, i.e., $\pi_{k+i}$, for all $i = \{0,1,...,N-1\}$.
% J_k(x_k,\pi_N) =
\begin{equation}\label{eq:cost_lqr_img}
     \sum_{i=0}^{N-1}c(\gls*{state}_{k+i}, \gls*{input}_{k+i}) + c_{goal}(\gls*{state}_{k+N}).
\end{equation}The optimal control input, i.e., $\gls*{input}_{k+i} = \pi_{k+i}(\gls*{state}_{k+i})$, for a given time index, i.e., i+k, can be obtained by minimizing the~(\ref{eq:cost_lqr_img}). Thus, cost (cost-to-go) which was proposed in~\cite{lewis1995optimal} is fully determined by
\begin{equation}\label{eq:costtogo}
    V_{k+i}(\gls*{state}_{k+i}) = \min_{\gls*{input}_{k+i}} \quad (c(\gls*{state}_{k+i}, \gls*{input}_{k+i}) + V_{k+1}(\gls*{fdes}(\gls*{state}_{k+i}, \gls*{input}_{k+i})).
\end{equation} The same procedure can be applied recursively in a backward direction for seeking the optimal $\pi_{k+i}(\gls*{state}_{k+i}) = \argmin_{\gls*{input}_{k+i}}(c(\gls*{state}_{k+i}, \gls*{input}_{k+i}) + V_{k+i}(\gls*{fdes}(\gls*{state}_{k+i}, \gls*{input}_{k+i})))$. \gls*{DDP} yields almost the same behaviour: first estimate optimal control and then apply a forward pass to determine the updated nominal trajectory. Consequently, \gls*{LQR} is a simplified version of \gls*{DDP}. \gls*{LQR} is one of the fundamental ways to obtain a closed-form solution for a given optimal control problem under which system dynamics is assumed to be linear. Let us assume the system dynamics is defined as in~(\ref{eq:euler_discretization}). The intuition of~\gls*{LQR} is to estimate the optimal control sequence for maneuvering the quadrotor from an initial position to the desired pose. Let N be the receding horizon whose optimal trajectory is to be determined. The total cost, i.e.,  $J_k(\gls*{state}_k,\pi_N)$, consists of three parts: initial, intermediate and final costs, where $\pi_N = \{\pi_k,\pi_{k+1},...,\pi_{k+i},...,\pi_{N-1}\}$
\begin{equation}\label{eq:lqr_1}
    J_k(\gls*{state}_k,\pi_N) = c_{start}(\gls*{state}_k) + \sum^{N-1}_{i=0} c(\gls*{state}_{k+i}, \gls*{input}_{k+i}) dt + c_{end}(\gls*{state}_{k+N}),
\end{equation} where $\frac{\partial^2 C_{start}(x_k)}{\partial x \partial x} \leq 0, \quad \frac{\partial^2 C_{goal}(x_{k+N})}{\partial x \partial x} \leq 0, \quad \frac{\partial^2 C}{\partial\begin{bmatrix}
x\\ 
u
\end{bmatrix}  \partial \begin{bmatrix}
x\\ 
u
\end{bmatrix} } \leq 0 $, and $\frac{\partial^2 C}{\partial u  \partial u} \leq 0$ are positive semidefinite Hessians to guarantee the minimizing of the total cost. The total cost can be formulated in various ways. In \gls*{LQR}, the total cost is defined as Quadratic costs as follows:
\begin{equation}\label{eq:lqr_2}
\begin{aligned}
    c_{start}(\gls*{state}_k) = \frac{1}{2}\gls*{state}_k^TQ_{start}\gls*{state}_k + \gls*{state}_k^Tq_{start}, \\ \quad c_{goal}(\gls*{state}_{k+N}) = \frac{1}{2}\gls*{state}_{k+N}^TQ_{goal}\gls*{state}_{k+N} + \gls*{state}_{k+N}^Tq_{goal}, \quad \\
    c(\gls*{state}_{k+i},\gls*{input}_{k+i}) = \frac{1}{2}\gls*{state}_{k+i}^TQ\gls*{state}_{k+i} + \frac{1}{2}\gls*{input}_{k+i}^TR\gls*{input}_{k+i} + \gls*{input}_{k+i}^TP\gls*{state}_{k+i} + \gls*{state}_{k+i}^Tp \\+ \gls*{input}_{k+i}^Tr + \gls*{params} = \frac{1}{2}\begin{bmatrix}
\gls*{state}_{k+i}\\\gls*{input}_{k+i}
\end{bmatrix}^T\underbrace{\begin{bmatrix}
Q & P^T\\ 
P & R
\end{bmatrix}}_{J_k}\begin{bmatrix}
\gls*{state}_{k+1}\\\gls*{input}_{k+1}
\end{bmatrix}_{k+i} + \begin{bmatrix}
\gls*{state}_{k+1}\\\gls*{input}_{k+1} 
\end{bmatrix} \underbrace{\begin{bmatrix}
p\\r
\end{bmatrix}}_{j_{k}} + \gls*{params},
\end{aligned}
\end{equation} where $i \in \{0,1,..., N-1$\}, $Q_{start} \in \mathbb{R}^{n_x \times n_x},Q_{goal} \in \mathbb{R}^{n_x \times n_x}, Q \in \mathbb{R}^{n_x \times n_x}, R \in \mathbb{R}^{n_u\times n_u}, P \in \mathbb{R}^{n_u \times n_x},  q_{start} \in \mathbb{R}^{n_x}, q_{goal} \in \mathbb{R}^{n_x}, p \in \mathbb{R}^{n_x}$, $r\in \mathbb{R}^{n_u}$, and $\gls*{params} \in \mathbb{R}$ are predefined in which $Q_{start}, Q_{goal}, Q$, and $R$ are positive definite, whereas $ J_k \geq 0 $ and $j_k \geq 0$ assumed to be positive semi-definite. \gls*{LQR} problem(~\ref{eq:lqr_1}) and~(\ref{eq:lqr_2}) provides an optimal $\pi_N$ in close form solution as expressed in~\ref{eq:costtogo}; the cost-to-go function, i.e.,~\ref{eq:costtogo}, can be reformulated as an explicit quadratic formulation as follows:
\begin{equation}\label{eq:lqr_ddt}
    V_{k+i}(\gls*{state}_{k+i}) = \frac{1}{2}\begin{bmatrix}
\gls*{state}_{k+i}\\\gls*{input}_{k+i}
\end{bmatrix}^TJ_{k+i}\begin{bmatrix}
\gls*{state}_{k+i}\\\gls*{input}_{k+i}
\end{bmatrix} +\begin{bmatrix}
\gls*{state}_{k+i}\\\gls*{input}_{k+i}
\end{bmatrix}^Tj_{k+i} + \gls*{params}.
\end{equation}

Estimation of both $J_{k+i}$ and $j_{k+i}$ can be obtained in a recursive way starting from the goal position $\gls*{state}_{x+N}$ to the initial position $\gls*{state}_k$, using Riccati differential equation for all $i= \{0,...,N-1\}$.
\begin{equation}\label{eq:riccadi}
    \begin{aligned}
    J_k = Q+A_k^TJ_{k+1}A_k - \\ (P+B_k^TJ_{k+1}A_k)^T\cdot(R+B_k^TJ_{k+1}B_k)^{-1}\cdot (P+B_k^TJ_{k+1}A_k) \\
    j_k = p +A_k^Tj_{k+1}+A_k^TJ_{k+1}c_k \\ - (P+B_k^TJ_{k+1}A_k)^T\cdot(R+B_k^TJ_{k+1}B_k)_k^{-1}\cdot(r+B_k^Tj_{k+1}+B_k^TJ_{k+1}c_k).
    \end{aligned}
\end{equation}In general, system dynamics is described by:
\begin{equation}
     \gls*{state}_{k+1} = \gls*{fdes}(\gls*{state}_k, \gls*{input}_k) = A_k\gls*{state}_k + B_k\gls*{input}_k,
\end{equation}. If the system dynamics is non-linear, $A_k$ and $B_k$ are recalculated by linearizing the $\gls*{fcons}$ at each time index. Since linearization has to be carried out in each iteration, it is called the~\gls*{ILQR}~\cite{li2004iterative}.
\begin{equation}
\begin{aligned}
    A_k = \frac{\partial \gls*{fcons}}{\partial \gls*{state}}(\gls*{state}_k, \gls*{input}_k),  \quad B_k = \frac{\partial \gls*{fcons}}{\partial \gls*{input}}(\gls*{state}_k, \gls*{input}_k).
\end{aligned}
\end{equation} Boundary or goal position conditions are given by $ S_{k+N} = Q_{goal}, \quad j_{k+N} = q_{goal}$. The feedback control policy in \gls*{LQR} is fully determined as follows:
\begin{equation}
\begin{aligned}
     \pi_k(\gls*{state}_k) = -(R+B_k^TJ_{k+1}B_k)^{-1} \cdot (P+B_k^TJ_{k+1}A_t)\gls*{state}_k \\ - (R+B_k^TJ_{k+1}B_k)^{-1}\cdot (r+B_k^Tj_{k+1}+B_k^TJ_{k+1}B_k).
\end{aligned}
\end{equation}

As given in~(\ref{eq:riccadi}), system stability depends on system dynamics. When quadrotor dynamics is non-linear, the stability of \gls*{ILQR} is not guaranteed. Jur and Berg~\cite{van2014iterated} attempted to address the stability issue by proposing a novel method called~\gls*{LQR} smoothing; this method can be applied for linear or non-linear systems to acquire the minimum-cost trajectory. The main difference in~\gls*{LQR} smoothing compared to~\gls*{LQR} is that LQR minimizes the cost of not only backward direction, i.e., cost-to-go, but also applies forward direction, i.e., cost-to-come~\cite{van2014iterated, van2016extended, sun2015stochastic}. However, the output of \gls*{LQR}, \gls*{ILQR} or~\gls*{LQR} smoothing does not address the system noise. Both linear or nonlinear state estimator may eliminate the system noise. \gls*{LQG}~\cite{todorov2008general, van2012lqg} is one of the ways to solve this problem.~\gls*{LQG} consists of a state estimator ,i.e., \gls*{KF}, and state feedback, i.e.,~\gls*{ILQR} or \gls*{LQR}.

\subsection{MPC based trajectory generation}
 
\begin{figure}[!ht]
\begin{center}
\includegraphics[width=12cm]{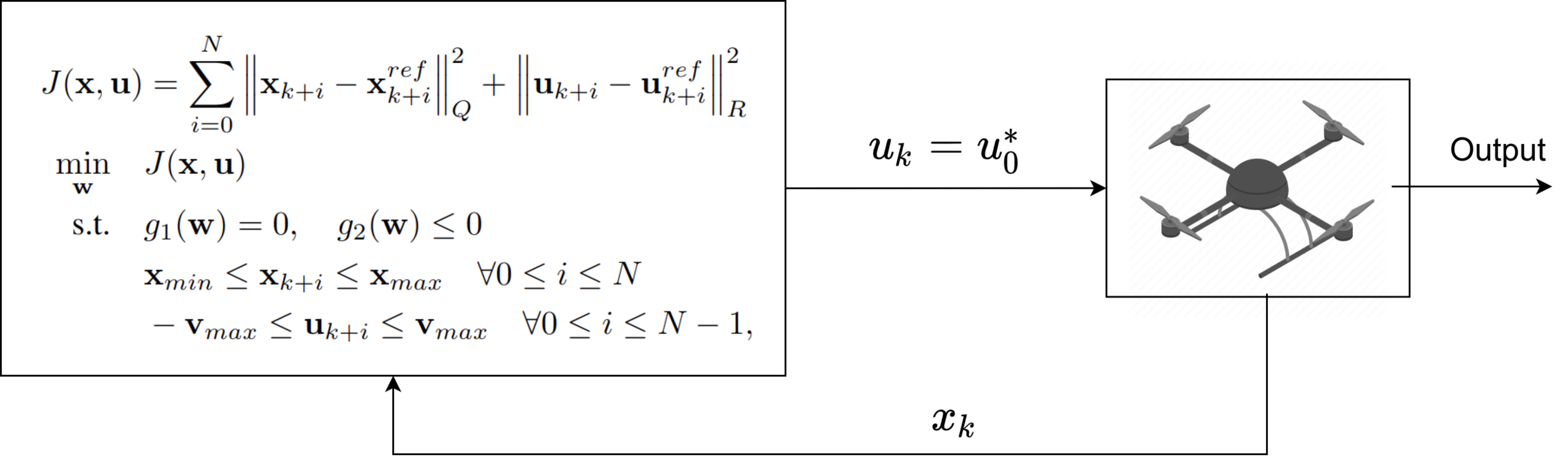}
\caption{The basic idea of MPC-based receding horizon planning}
\label{f:mpc_overview}
\end{center}
\end{figure}

As detailed in section~\ref{sec:lqr},  unaccountability of addressing sudden disturbances is the main limitation of OCP techniques (e.g.,~\gls*{LQR},~\gls*{DDP}); this is due to its nature. \gls*{LQR} calculates fixed receding control policy and applies to the system; there is no intervention during the control policy execution. \gls*{MPC} is one of the ways to address the preceding problem, which is characteristic of both~\gls*{LQR} and \gls*{DDP}. The difference between \gls*{MPC} and \gls*{LQR} is that only the first portion of the control policy is applied to system in \gls*{MPC} through the calculate of full control policy rather than employing full control policy as in \gls*{LQR}. Let us assume the system dynamics as given in~(\ref{eq:euler_discretization_dis}). In general, \gls*{MPC} can be formed as follows:

\begin{equation}\label{eq:mpc_general}
\begin{aligned}
\min_{w} \quad &  J_{end}(\gls*{state}_{k+N}, \gls*{state}^{ref}_{k+N}) +  J_k(\gls*{state}, \gls*{input}, \gls*{state}^{ref}, \gls*{input}^{ref})\\
\textrm{s.t.} \quad & \gls*{state}_{k+1} = \gls*{fdes}(\gls*{state}_{k}, \gls*{input}_{k})\\
  & \gls*{state}^{min} \leq \gls*{state}_{k+i} \leq \gls*{state}^{max} \quad \forall 0 \leq i \leq N \\
  & \gls*{input}^{min} \leq \gls*{input}_{k+i} \leq \gls*{input}^{max} \quad \forall 0 \leq i < N-1 \\
  & \gls*{g}_1(w) = 0 \\
  &  \gls*{g}_2(w) \leq 0,
\end{aligned}
\end{equation} where $w={\gls*{input}_k,..., \gls*{input}_{k+N-1}}$ is the optimal control sequence to be estimated in each iteration. Variable $ J_{end}(\gls*{state}_{k+N})$ plays a significant role in terms of the stability of the system locally and globally. Presenting local stability is relatively easy, e.g., Lyapunov's analysis compared to global stability. In addition to terminal cost, terminal constraints for states should be enforced, which is quite computationally challenging for real-time applications. Moreover, enforcing terminal constraints is even more difficult for non-linear dynamics. Thus, in most of the practical applications, terminal constraints are not enforced into the optimization procedure. Furthermore, classical \gls*{MPC} lacks recursive feasibility. Several varieties of \gls*{MPC} have been proposed to address processing issues to a certain extent. For a linear system, the performance index, i.e., $J_k(\gls*{state}, \gls*{input}, \gls*{z}^{ref}, \gls*{input}^{ref})$, can be defined as follows:
\begin{equation}
\begin{aligned}
     J_k(\gls*{state}, \gls*{input}, \gls*{z}^{ref}, \gls*{input}^{ref}) \\ = \sum^{N-1}_{i=0} ((\gls*{state}_{k+i}-\gls*{state}_{k+i}^{ref})^TQ_x(\gls*{state}_{k+i}-\gls*{state}_{k+i}^{ref}) + (\gls*{input}_{k+i}-\gls*{input}_{k+i}^{ref})^TR_u(\gls*{input}_{k+i}-\gls*{input}_{k+i}^{ref})) \\ + (\gls*{state}_{k+N}-\gls*{state}_{k+N}^{ref})^TP(\gls*{state}_{k+N}-\gls*{state}_{k+N}^{ref}),
\end{aligned}
\end{equation} where $Q_x$, which is a positive semi-definite matrix, consists of the state error penalty coefficients, whereas $R_u$ should be positive definite and P is state error on the terminal cost. In principle, stability and feasibility are not assured implicitly. Consequently, stability and feasibility tend to improve for the longer receding horizon, which is quite challenging due to computational demands. 

Quadrotor dynamics are usually expressed in a non-linear fashion. Therefore \gls*{LQR} or linear \gls*{MPC} can not be applied without linear approximation. Hence, \gls*{NLP}-based approach has to be applied. Direct multiple shooting and direct collocation are the main two techniques that are used to transform \gls*{OCP} into \gls*{NLP}. In both direct multiple shooting and direct collocation, the state is minimized in addition to controlling inputs. Direct multiple shooting differs from direct collocation due to the way of the problem formulation. In multiple shooting, the problem is quantized into N subintervals, i.e., receding horizon length. In direct collocation, however, those subintervals are further described by a set of polynomials such as B-spline or Lagrangian; this will increase the problem sparsity. On the contrary, the number of optimization parameters to be optimized has dramatically increased in direct collocation compared to multiple shooting. This, collocation is better when it is accuracy-wise, but direct multiple shooting is better when it is performance-wise. In~\cite{mpc_mc}, the trajectory tracking problem is formulated based on direct collocation and multiple shooting. Further, the researchers  have proved that multiple shooting has a lower computational footprint compared to direct collocation.

\subsection{Disturbance Estimation}

In the context of optimal trajectory planning, simultaneously computing optimal control policy, which is required to respond to unknown, sudden disturbances, and handling kinematics (i.e., obstacle avoidance) as well as dynamics (i.e., satisfying velocity and acceleration constraints) yields a challenging problem, especially for quadrotors. While geometry-based path planning techniques~\cite{likhachev2004ara, karaman2011sampling} ensure the asymptotical optimality of a path, they however do not consider quadrotor dynamics. But, it is essential that the generation of an optimal control policy ensures dynamic feasibility. So, in~\cite{perez2012lqr, kulathunga2020path}, \gls*{LQR} was incorporated into path planning, by which both dynamic feasibility and local optimality were guaranteed. However, local optimality does not necessarily yield global optimality~\cite{pacelli2018integration}. In~\cite{liu2017planning, gao2018online}, a set of motion primitives was used to find feasible trajectories ensuring both global and local optimality. When dealing with unknown disturbances,~\gls*{MPC} is a more robust technique than~\gls*{LQR}. In ~\cite{liu2017planning}, \gls*{MPC}-based trajectory planning approach was proposed, ensuring both the local and global optimality. However, none of the aforesaid approaches formally guarantees stability and safety. Lyapunov's analysis can be applied to confirm the local stability. Moreover, the terminal constraints set~\cite{mason1985robot} can be incorporated. However, those measures are time-consuming, which directly affects the real-time performance~\cite{liu2017search}. A set of \gls*{CBF} was proposed for improving real-time performance without affecting the system stability in~\cite{ames2014rapidly,wu2015safety, ames2016control}. Recently, reference governors-based techniques were proposed in~\cite{kolmanovsky2014reference,garone2015explicit}, enforcing safety constraints. It is natural that designing a path planer is followed by the actual controller to maneuver~\gls*{MAV}. In such approaches, a reference governor can be used to handle the stability and constraint satisfaction separately to ensure system stability~\cite{arslan2017smooth}. 

The above approaches are employed to estimate optimal control policy for safe navigation while imposing stability either using Lyapunov functions or reference governors. On the other hand, Li et al.~\cite{li2020fast} proposed to obtain an optimal control policy using a \gls*{SDDM}. They have modelled the system dynamics as a linear, time-invariant as follows:
\begin{equation}
    \dot{\gls*{state}} = A\gls*{state} + B\gls*{input},
\end{equation} where $\gls*{input}$ indicates the control input. System state, i.e., $\gls*{state}:=(\gls*{p}(t),\mathbf{y}(t))$, consists of two parts: $\gls*{p}$ and $\mathbf{y}$, where $\gls*{p}(t)$ denotes the quadrotor position at a given time t and $\mathbf{y}(t)$ describes the higher-order terms, e.g., velocity, acceleration, etc. In the latter work, the quadratic norm was utilized to represent the error between robot position and close-in obstacles positions. The quadratic norm is defined as $\left \| \gls*{p} \right \|_R := \sqrt{\gls*{p}^TR\gls*{p}}$, where R is a symmetric positive matrix. R[$\gls*{psi}_z$] is fully determined by the ~\gls*{MAV} heading direction $\gls*{psi}_z$ at a given time instance as follows:

\begin{equation}
    R[\gls*{psi}_z] = \left\{\begin{matrix}
c_2I + (c_1-c_2)\frac{\gls*{psi}_z\gls*{psi}_z^T}{\left \| \gls*{psi}_z \right \|^2}, & \; if \; \gls*{psi}_z \neq 0\\ 
 c_1I,& otherwise  
\end{matrix}\right.,
\end{equation} where both $c_1$ and $c_2$ are predefined scales such that $c_2>c_1>0$; this process is called the~\gls*{SDDM}, trajectory will be bounded incorporating~\gls*{SDDM} information. Since quadrotor dynamics is linear, a reference governor~\cite{garone2015explicit} is introduced to maintain safety and stability. Other than LQR and MPC, there exist several receding horizon-based techniques for optimal trajectory planning as given in Table~\ref{recceding_horizon}.  

\begin{table}[]
\caption{Comparison of properties of receding horizon trajectory planning techniques}
\begin{tabular}{|l|c|c|c|c|}
\hline
\multicolumn{1}{|c|}{\multirow{2}{*}{Algorithm}}                                                               & \multicolumn{2}{c|}{Motion Model} & \multicolumn{2}{c|}{Gradient Estimator} \\ \cline{2-5} 
\multicolumn{1}{|c|}{}                                                                                         & Linear          & Nonlinear          & Hamiltonian      & \begin{tabular}[c]{@{}l@{}}Gradient \end{tabular}      \\ \hline
  \begin{tabular}[c]{@{}l@{}} Differential\\  Dynamic Programming \\ (DDP)~\cite{2005.00985}\end{tabular}                                                                        &     \xmark           &    \checkmark                 &        \xmark          &           \checkmark            \\ \hline
    \begin{tabular}[c]{@{}l@{}}Linear Quadratic \\ Regulator (LQR)~\cite{liu2016pid}\end{tabular}                                                                             &       \checkmark           &             \xmark       &         \xmark         &        \xmark               \\ \hline
    \begin{tabular}[c]{@{}l@{}}Iterative LQR \\ (iLQR)~\cite{cowling2006optimal} \end{tabular}                                                                                         &         \xmark         &     \checkmark                &       \xmark            &         \checkmark              \\ \hline
     \begin{tabular}[c]{@{}l@{}}Linear Model Predictive \\ Control (MPC)~\cite{BANGURA201411773}\end{tabular}                                                                      &    \checkmark               &         \xmark            &           \xmark        &       \xmark                \\ \hline
\begin{tabular}[c]{@{}l@{}} Nonlinear Model \\ Predictive Control \\ (NMPC)~\cite{mpc_mc}\end{tabular}             &      \checkmark           &      \checkmark              &      \checkmark            &    \xmark                  \\ \hline
\begin{tabular}[c]{@{}l@{}}Constrained Nonlinear \\ Model Predictive \\ Control CGMRES \\ (NMPC-CGMRES)~\cite{ohtsuka1997real}\end{tabular}  &      \xmark           &        \checkmark             &       \checkmark            &   \xmark                   \\ \hline
\begin{tabular}[c]{@{}l@{}} Corridor-based Model \\Predictive  Contouring \\ Control(CMPCC)~\cite{ji2020cmpcc}\end{tabular} &      \checkmark               &     \xmark               &          \xmark           &  \xmark                       \\ \hline
\begin{tabular}[c]{@{}l@{}}Constrained Nonlinear \\ Model Predictive \\ Control Newton \\ (NMPC-Newton)~\cite{deng2019parallel}\end{tabular} &     \xmark               &       \checkmark              &          \xmark           &  \xmark                       \\ \hline
\begin{tabular}[c]{@{}l@{}}Model Preidictive Path \\ Integral Control \\ (MPPI)~\cite{mohamed2020model}\end{tabular} &     \checkmark                &       \checkmark              &          \xmark          &  \xmark                       \\ \hline
\begin{tabular}[c]{@{}l@{}}Cross Entropy Method \\ (CEM)~\cite{olivares2012see}\end{tabular} &     \checkmark               &       \checkmark              &          \xmark           &  \xmark                       \\ \hline
\end{tabular}
\label{recceding_horizon}
\end{table}

\section{Solving Trajectory Planning Problem}\label{sec:solvers_opt}

As explained in the preceding sections, several constraints (e.g., soft and hard) are imposed to ensure dynamic feasibility, smooth navigation, handling disturbances, etc. Hence, optimal trajectory planning is posed as a constraint optimization problem in most situations. Constraint-based optimization problems are solved in two different ways: adding hard constraints or introducing soft constraints. In general, a constraint-based optimization problem can be formulated as a quadruple, i.e., $P_{constraint} = (c,  \gls*{g}1,  \gls*{g}2, J)$, where c stands for performance index or cost function, whereas equality and inequality constraints are given by g1 and g2, respectively. The objective function is given by $J$. In hard constraint-based formulation, the optimal solution, i.e., $\gls*{w}$, for $P_{constraint}$ is calculated, ensuring all the constraints. In soft constraints formulation, the objective function does not need to satisfy all the constraints, but satisfying those constraints will improve the final $\gls*{w}$. D. Mellinger and V.Kumar~\cite{mellinger2011minimum} took the lead in proposing a successful approach for trajectory generation as a hard constraint-based optimization approach, i.e., Minimum-snap. Subsequently, in~\cite{richter2016polynomial}, the researchers extended the Minimum-snap trajectory generation as an unconstrained or soft constraint-based optimization problem.

When generating trajectories, ensuring a collision-free path is essential. Hence, representing free space in a structured way and imposing obstacle constraints for trajectory generation is a must for safety. Free space can be represented in different ways such as cubes (~\cite{chen2015real, gao2018online}), spheres (\cite{gao2016online, gao2019flying}) and polyhedrons (\cite{liu2017planning}). The intuition of these approaches is to apply path planning through the free space to obtain the intermediate waypoints. Once intermediate waypoints are extracted, the trajectory generation procedure is utilised for retrieving a smooth, feasible, collision-free trajectory. On the other hand, in~\cite{ding2018trajectory, ding2019efficient, gao2018online}, kinodynamic path planning followed by B-spline-based trajectory generation is considered. Most of the works that were proposed for soft constraint-based trajectory generation formulated optimal trajectory planning as nonlinear optimization problems in which smoothness and safety were introduced as soft constraints. Most of the time gradient-descent based~\cite{zucker2013chomp} or gradient-free approaches~\cite{kalakrishnan2011stomp, oleynikova2016continuous} were applied for minimizing the cost of smoothness and safety. 

The constraints optimization problem can be designed in either~\gls*{QP} or~\gls*{NLP} form. In \gls*{QP}, the procedure is to minimize or maximize the objective subject to a set of linear constraints in most situations. On the other hand, non-quadratic programming is used to handle the non-linear constraints each of which has a unique nature to solve the problem.  In general,~\gls*{QP} objective can be described as:
 
\begin{equation}
\begin{aligned}
 \min_{\gls*{state}} \quad & \frac{1}{2}\gls*{state}^TQ\gls*{state}+c^T\gls*{state}  \\
\textrm{s.t.} \quad & A\gls*{state} \succeq b ,
\end{aligned}
\end{equation} where $A\gls*{state} \succeq b $ stands for the set of linear inequalities. Q is a positive symmetric matrix. There are various ways to solve~\gls*{QP}, including interior point, active set and gradient projection. In some situations, multiple variables that are to be optimized are integer values; those are solved as~\gls*{MIQP}. For example, FASTER~\cite{tordesillas2020faster} used \gls*{MIQP} for safe trajectory planning with aggressive controls~\cite{landry2016aggressive}.  

Most of the recent optimal trajectory planning techniques~\cite{zhou2019robust, usenko2017real, gao2017gradient, oleynikova2016continuous} were formulated as~\gls*{GTO} in which optimization problem was designed as a non-linear form. The gradient descent is performed with respect to each parametrization index of $\Gamma$ to minimize the different, i.e.,  $\Gamma_{i+1}-\Gamma_i$. Hence, $\Gamma_{i+1}$ can be determined by solving the following optimization problem as given in~\cite{quinlan1994real, zucker2013chomp}.
\begin{equation}
    \Gamma_{i+1} = \argmin_{\Gamma}  J(\Gamma_i)  + (J(\Gamma)- J(\Gamma_i))^T\bigtriangledown J(\Gamma_i) + \frac{\eta}{2} \left \|  \Gamma-\Gamma_i \right \|^2_M,
\end{equation}where M is a weighting matrix and $\eta$ is a regularization parameter. \gls*{GTO} is rather popular due to its ability to deform ineffability trajectory segments, low memory requirement and high throughput. Despite having the listed advantages, \gls*{GTO} can not avoid the problem of a local minimum. STOMP~\cite{kalakrishnan2011stomp} is one of the early techniques proposed to address the local minimum problem. STOMP is based on the gradient-free technique. However, STOMP is unable to obtain real-time performance. Besides STOMP, the local minimum problem has been addressed by various recent works. Yet, this remains an open problem to be solved. Zhou~\cite{zhou2020robust} proposed a method, i.e., \gls*{PGO}, for overcoming local minima problem by generating topologically distinct paths and doing parallel optimization. Furthermore, various solvers can be utilized for solving optimization problems, including BOBYQA~\cite{powell2009bobyqa}, L-BFGS~\cite{liu1989limited, kulathunga2022optimization}, ACADO~\cite{houska2011acado}, SLSQP~\cite{kraft1988software}, Proximal Operator Graph  Solver (POGS)~\cite{parikh2014block, fougner2018parameter}, \gls*{SQP} and MMA~\cite{svanberg2002class}. Shravan et al.~\cite{krishnan2019towards} proposed a trajectory optimization technique in a distributed setup in which the researchers evaluated their formulation with several solvers. According to their observations, BOBYQA is faster compared to BFGS and SLSQP, while MMA yielded a similar performance to that of BOBYQA. In~\cite{liu2019optimization}, L-BFGS was proposed for finding the shortest path in real-time; in this research effort however L-BFGS does not guarantee optimality, only feasibility is enforced. \gls*{MPCC}~\cite{foehn2017fast} yet another proposed method for fast trajectory optimization in real-time. Moreover, Mathieu and Nicolas~\cite{geisert2016trajectory} proposed a~\gls*{SQP}-based trajectory generation approach for carrying augmented loads. The intuition behind selecting \gls*{SQP} over other solvers is due to its superlinear convergency and ability to handle non-linear constraints within milliseconds.

\section{Conclusion}\label{sec:conclusion}
All in all, we have thoroughly reviewed the trajectory planning problem in the paradigm of plan-based control for~\gls*{MAVs}. Such trajectory planning problem was broken down into a set of subproblems: free-space segmentation, motion model selection, initial waypoints identification, initial trajectory generation, continuous trajectory refinement, and receding horizon trajectory planning. Afterwards, for each subproblem, we examined how previous research has addressed those by presenting and evaluating various approaches to the considered subproblem. Finally, several selected recent approaches were listed (Table 3) according to the listed subproblems we have identified. With that, we concluded that the trajectory planning problem can be designed by addressing those subproblems carefully for \gls*{MAVs}.

\begin{adjustbox}{angle=90} \label{t:selected_methods}
\centering
\tiny
\begin{tabular}{|l|l|l|l|l|l|l|} 
\hline
Approach & \begin{tabular}[c]{@{}l@{}}Dynamics Model\\(Exact\textbar{}Empirical\\Differential \\flatness (DF))\end{tabular} & \begin{tabular}[c]{@{}l@{}}Intermediate Waypoint \\ Selection \end{tabular}  & \begin{tabular}[c]{@{}l@{}}Initial Trajectory \\ Generation \end{tabular}                & \begin{tabular}[c]{@{}l@{}}Continuous trajectory \\ refinement and solver \end{tabular} & \begin{tabular}[c]{@{}l@{}}Free space \\ extraction \end{tabular}      & \begin{tabular}[c]{@{}l@{}}Receding\\horizon \\ planning or\\controlling \end{tabular}  \\ 
\hline
       \cite{zhou2020robust}  & DF  & \begin{tabular}[c]{@{}l@{}}Sampling-based \\topological search \end{tabular} & PGO based B-splines                                                                      & GTO   & ESDF                                                                   & -                                                                                       \\ 
\hline
       \cite{ding2018trajectory}  & DF                                                                                                               & Kinodynamic-based search                                                     & B-splines                                                                                & EO using QCQP                                          & ESDF                                                                   & -                                                                                       \\ 
\hline
       \cite{liu2017search}  & DF                                                                                                               & Kinodynamic-based search                                                     & Linear Quadratic Minimum Time~                                                           & unconstrained QP                                                                        & ~\cite{shen2012stochastic}                                                      & RHC                                                                                     \\ 
\hline
        \cite{oleynikova2016continuous}  & DF                                                                                                               & Informed-RRT*                                                                & Continuous time polynomial                                                               & BFGS                                                                                    & ESDF                                                                   & -                                                                                       \\ 
\hline
       \cite{gao2020teach}  & DF                                                                                                               & -                                                                            & \begin{tabular}[c]{@{}l@{}}Piecewise Bézier-based curve\\with minimum-jerk \end{tabular} & Elastic band optimization                                                               & Convex Cluster                                                         & -                                                                                       \\ 
\hline
        \cite{zhou2019robust} & DF                                                                                                               & A*~ kinodynamic search                                                       & B-splines                                                                                & NLopt~\cite{johnson2014nlopt}                                                                  & ESDF                                                                   & GTC~~                                                                                   \\ 
\hline
        \cite{ding2019efficient} & DF    & \begin{tabular}[c]{@{}l@{}}B-spline kinodynamic \\search \end{tabular}       & EO                                                                      & \textasciitilde{}QCQP                                                                   & TSDF                                                                   & -                                                                                       \\ 
\hline
        \cite{zucker2013chomp}  & DF                                                                                                               & -                                                                            & CHOMP                                                                   & Functional gradient~\cite{quinlan1994real}                                                    & ESDF                                                                   & CHOMP                                                                                   \\ 
\hline
        \cite{zhou2020ego} & DF  & A*                                                                           & uniform B-spline                                                                         & \begin{tabular}[c]{@{}l@{}}BFGS, L-BFGS,\\T-NEWTON~\cite{steiha1983truncatednewton} \end{tabular}       & ESDF                                                                   & -                                                                                       \\ 
\hline
       \cite{gao2018online}  & DF    & fast marching-based~search    & Bernstein polynomial     & Mosek~\cite{andersen2000mosek}    & TSDF                                                                   & -                                                                                       \\ 
\hline
        \cite{gao2016online} & DF  & RRG combined with A*                                                         & piecewise polynomials                                                                    & QCQP                                                                                    & KD-tree                                                                & GTC                                                                                     \\ 
\hline
       \cite{van2014iterated}  & Exact                                                                                                            & line search                                                                  & Iterated LQR Smoothing                                                                   & Iterated LQR Smoothing                                                                  & -                                                                      & -                                                                                       \\ 
\hline
       \cite{lin2018autonomous}   & Exact                                                                                                            & A*                                                                           & \begin{tabular}[c]{@{}l@{}}Visual-InertialNavigation\\System (VINS) \end{tabular}        & Gradient-based                                                                          & TSDF                                                                   & GTC                                                                                     \\ 
\hline
        \cite{hu2019real} & DF                                                                                                               & RRT*                                                                         & Uniform-Bspline                                                                          & MMA and BFGS                                                                            & \begin{tabular}[c]{@{}l@{}}OctoMap and \\Circular Buffer \end{tabular} & GTC                                                                                     \\ 
\hline
       \cite{liu2017planning}   & DF                                                                                                               & JPS                                                                          & Minimum-span                                                                             & Constrained QP                                                                          & SFC                                                                    & RHC                                                                                     \\ 
\hline
        \cite{li2020fast}  & Empirical                                                                                                        & piecewise-linear path                                                        & SDDM                                                                                     & SDDM                                                                                    & Constrained QP                                                         & -                                                                                       \\ 
\hline
       \cite{tordesillas2020faster}  & DF                                                                                                               & JPS                                                                          & Cubic Bézier curve                                                                       & MIQP using Gurobi~\cite{gurobi2018gurobi}                                                      & SFC                                                   & -                                                                                       \\ 
\hline
     \cite{ji2020cmpcc}    & Empirical                                                                                                        & -                                                                            & CMPCC                                                                                    & OSQP\cite{stellato2020osqp}                                                                 & SFC                                                                    & RHC                                                                                     \\ 
\hline
       \cite{mehrez2017optimization}  & Empirical                                                                                                        & -                                                                            & NMPC                                                                                     & ACADO~\cite{houska2011acado}                                                                  & -                                                                      & MHE                                                                                     \\ 
\hline
       \cite{kamel2017model}  & Empirical                                                                                                        & -                                                                            & NMPC                                                                                     & SQP                                                                                     & -                                                                      & RHC                                                                                     \\ 
\hline
         \cite{chen2015real} & DF                                                                                                               & A*                                                                           & Multi-segment polynomial                                                                 & OOQP~\cite{gertz2003object}                                                             & OctoMap                                                                & GTC                                                                                     \\ 
\hline
      \cite{richter2016polynomial}   & Exact                                                                                                            & RRT*                                                                         & Minimum-Span                                                                             & Unconstrained QP                                                                        & OctoMap                                                                & GTC                                                                                     \\ 
\hline
       \cite{tordesillas2020mader}  & DF                                                                                                               & MINVO basis~\cite{tordesillas2020minvo}                                                & Uniform B-spline                                                                         & Augmented Lagrangian~\cite{conn1991globally}                                                                     & Outer polyhedral~                                                      & -                                                                                       \\ 
\hline
        \cite{deits2015efficient}  & DF                                                                                                               & Piecewise linear path                                                        & Piecewise polynomial                                                                     & MIQP using Mosek                                                                        & IRIS                                                                   & -                                                                                       \\ 
\hline
        \cite{tang2019real}  & DF                                                                                                               & \begin{tabular}[c]{@{}l@{}}Non-uniform kinodynamic \\search \end{tabular}    & Uniform B-spline                                                                         & Constrained QP                                                                          & ESDF                                                                   & RHC                                                                                     \\ 
\hline
       \cite{mpc_mc}  & Empirical                                                                                                        & Uniform B-spline                                                             & NMPC                                                                                     &  CasADi\cite{andersson2019casadi} with Ipopt\cite{biegler2009large}                                                                       & ESDF                                                                   & PID                                                                                     \\
\hline
\end{tabular}
\end{adjustbox}

% \section*{References}

\bibliography{mybibfile}

\begin{thebibliography}{100}
\expandafter\ifx\csname url\endcsname\relax
  \def\url#1{\texttt{#1}}\fi
\expandafter\ifx\csname urlprefix\endcsname\relax\def\urlprefix{URL }\fi
\expandafter\ifx\csname href\endcsname\relax
  \def\href#1#2{#2} \def\path#1{#1}\fi

\bibitem{px4}
Pixhawk 4, \url{https://dev.px4.io/v1.9.0} (2020).

\bibitem{dji}
Dji, \url{https://www.dji.com/} (2020).

\bibitem{mellinger2011minimum}
D.~Mellinger, V.~Kumar, Minimum snap trajectory generation and control for
  quadrotors, in: 2011 IEEE International Conference on Robotics and
  Automation, IEEE, 2011, pp. 2520--2525.

\bibitem{ramasamy2014dynamically}
S.~Ramasamy, G.~Wu, K.~Sreenath, Dynamically feasible motion planning through
  partial differential flatness., in: Robotics: Science and Systems, 2014.

\bibitem{wanasinghe2015relative}
T.~R. Wanasinghe, G.~K. Mann, R.~G. Gosine, Relative localization approach for
  combined aerial and ground robotic system, Journal of Intelligent \& Robotic
  Systems 77~(1) (2015) 113--133.

\bibitem{van2016extended}
J.~van~den Berg, Extended lqr: Locally-optimal feedback control for systems
  with non-linear dynamics and non-quadratic cost, in: Robotics Research,
  Springer, 2016, pp. 39--56.

\bibitem{deits2015computing}
R.~Deits, R.~Tedrake, Computing large convex regions of obstacle-free space
  through semidefinite programming, in: Algorithmic foundations of robotics XI,
  Springer, 2015, pp. 109--124.

\bibitem{ling2017building}
Y.~Ling, S.~Shen, Building maps for autonomous navigation using sparse visual
  slam features, in: 2017 IEEE/RSJ International Conference on Intelligent
  Robots and Systems (IROS), IEEE, 2017, pp. 1374--1381.

\bibitem{savin2017algorithm}
S.~Savin, An algorithm for generating convex obstacle-free regions based on
  stereographic projection, in: 2017 International Siberian Conference on
  Control and Communications (SIBCON), IEEE, 2017, pp. 1--6.

\bibitem{zhong2020generating}
X.~Zhong, Y.~Wu, D.~Wang, Q.~Wang, C.~Xu, F.~Gao, Generating large convex
  polytopes directly on point clouds, arXiv preprint arXiv:2010.08744.

\bibitem{kulathunga2022optimization}
G.~Kulathunga, H.~Hamed, D.~Devitt, A.~Klimchik, Optimization-based trajectory
  tracking approach for multi-rotor aerial vehicles in unknown environments,
  IEEE Robotics and Automation Letters 7~(2) (2022) 4598--4605.

\bibitem{bentley1975multidimensional}
J.~L. Bentley, Multidimensional binary search trees used for associative
  searching, Communications of the ACM 18~(9) (1975) 509--517.

\bibitem{chen2015real}
J.~Chen, K.~Su, S.~Shen, Real-time safe trajectory generation for quadrotor
  flight in cluttered environments, in: 2015 IEEE International Conference on
  Robotics and Biomimetics (ROBIO), IEEE, 2015, pp. 1678--1685.

\bibitem{gao2018online}
F.~Gao, W.~Wu, Y.~Lin, S.~Shen, Online safe trajectory generation for
  quadrotors using fast marching method and bernstein basis polynomial, in:
  2018 IEEE International Conference on Robotics and Automation (ICRA), IEEE,
  2018, pp. 344--351.

\bibitem{gao2016online}
F.~Gao, S.~Shen, Online quadrotor trajectory generation and autonomous
  navigation on point clouds, in: 2016 IEEE International Symposium on Safety,
  Security, and Rescue Robotics (SSRR), IEEE, 2016, pp. 139--146.

\bibitem{gao2019flying}
F.~Gao, W.~Wu, W.~Gao, S.~Shen, Flying on point clouds: Online trajectory
  generation and autonomous navigation for quadrotors in cluttered
  environments, Journal of Field Robotics 36~(4) (2019) 710--733.

\bibitem{liu2017planning}
S.~Liu, M.~Watterson, K.~Mohta, K.~Sun, S.~Bhattacharya, C.~J. Taylor,
  V.~Kumar, Planning dynamically feasible trajectories for quadrotors using
  safe flight corridors in 3-d complex environments, IEEE Robotics and
  Automation Letters 2~(3) (2017) 1688--1695.

\bibitem{stentz1997optimal}
A.~Stentz, Optimal and efficient path planning for partially known
  environments, in: Intelligent unmanned ground vehicles, Springer, 1997, pp.
  203--220.

\bibitem{noreen2016comparison}
I.~Noreen, A.~Khan, Z.~Habib, A comparison of rrt, rrt* and rrt*-smart path
  planning algorithms, International Journal of Computer Science and Network
  Security (IJCSNS) 16~(10) (2016) 20.

\bibitem{gao2017gradient}
F.~Gao, Y.~Lin, S.~Shen, Gradient-based online safe trajectory generation for
  quadrotor flight in complex environments, in: 2017 IEEE/RSJ International
  Conference on Intelligent Robots and Systems (IROS), IEEE, 2017, pp.
  3681--3688.

\bibitem{zhou2019robust}
B.~Zhou, F.~Gao, L.~Wang, C.~Liu, S.~Shen, Robust and efficient quadrotor
  trajectory generation for fast autonomous flight, IEEE Robotics and
  Automation Letters 4~(4) (2019) 3529--3536.

\bibitem{webb2013kinodynamic}
D.~J. Webb, J.~Van Den~Berg, Kinodynamic rrt*: Asymptotically optimal motion
  planning for robots with linear dynamics, in: 2013 IEEE International
  Conference on Robotics and Automation, IEEE, 2013, pp. 5054--5061.

\bibitem{allen2016real}
R.~Allen, M.~Pavone, A real-time framework for kinodynamic planning with
  application to quadrotor obstacle avoidance, in: AIAA Guidance, Navigation,
  and Control Conference, 2016, p. 1374.

\bibitem{ding2019efficient}
W.~Ding, W.~Gao, K.~Wang, S.~Shen, An efficient b-spline-based kinodynamic
  replanning framework for quadrotors, IEEE Transactions on Robotics 35~(6)
  (2019) 1287--1306.

\bibitem{rousseau2019minimum}
G.~Rousseau, C.~S. Maniu, S.~Tebbani, M.~Babel, N.~Martin, Minimum-time
  b-spline trajectories with corridor constraints. application to
  cinematographic quadrotor flight plans, Control Engineering Practice 89
  (2019) 190--203.

\bibitem{usenko2017real}
V.~Usenko, L.~von Stumberg, A.~Pangercic, D.~Cremers, Real-time trajectory
  replanning for mavs using uniform b-splines and a 3d circular buffer, in:
  2017 IEEE/RSJ International Conference on Intelligent Robots and Systems
  (IROS), IEEE, 2017, pp. 215--222.

\bibitem{ratliff2009chomp}
N.~Ratliff, M.~Zucker, J.~A. Bagnell, S.~Srinivasa, Chomp: Gradient
  optimization techniques for efficient motion planning, in: 2009 IEEE
  International Conference on Robotics and Automation, IEEE, 2009, pp.
  489--494.

\bibitem{li2004iterative}
W.~Li, E.~Todorov, Iterative linear quadratic regulator design for nonlinear
  biological movement systems., in: ICINCO (1), 2004, pp. 222--229.

\bibitem{van2014iterated}
J.~van~den Berg, Iterated lqr smoothing for locally-optimal feedback control of
  systems with non-linear dynamics and non-quadratic cost, in: 2014 American
  Control Conference, IEEE, 2014, pp. 1912--1918.

\bibitem{todorov2008general}
E.~Todorov, General duality between optimal control and estimation, in: 2008
  47th IEEE Conference on Decision and Control, IEEE, 2008, pp. 4286--4292.

\bibitem{nageli2017real}
T.~N{\"a}geli, J.~Alonso-Mora, A.~Domahidi, D.~Rus, O.~Hilliges, Real-time
  motion planning for aerial videography with dynamic obstacle avoidance and
  viewpoint optimization, IEEE Robotics and Automation Letters 2~(3) (2017)
  1696--1703.

\bibitem{ji2020cmpcc}
J.~Ji, X.~Zhou, C.~Xu, F.~Gao, Cmpcc: Corridor-based model predictive
  contouring control for aggressive drone flight, arXiv preprint
  arXiv:2007.03271.

\bibitem{ames2016control}
A.~D. Ames, X.~Xu, J.~W. Grizzle, P.~Tabuada, Control barrier function based
  quadratic programs for safety critical systems, IEEE Transactions on
  Automatic Control 62~(8) (2016) 3861--3876.

\bibitem{romero2022time}
A.~Romero, R.~Penicka, D.~Scaramuzza, Time-optimal online replanning for agile
  quadrotor flight, arXiv preprint arXiv:2203.09839.

\bibitem{wang2022geometrically}
Z.~Wang, X.~Zhou, C.~Xu, F.~Gao, Geometrically constrained trajectory
  optimization for multicopters, IEEE Transactions on Robotics.

\bibitem{upadhyay2022generation}
S.~Upadhyay, T.~Richardson, A.~Richards, Generation of dynamically feasible
  window traversing quadrotor trajectories using logistic curve, Journal of
  Intelligent \& Robotic Systems 105~(1) (2022) 1--17.

\bibitem{torrente2021data}
G.~Torrente, E.~Kaufmann, P.~Foehn, D.~Scaramuzza, Data-driven mpc for
  quadrotors, IEEE Robotics and Automation Letters.

\bibitem{tang2021real}
L.~Tang, H.~Wang, Z.~Liu, Y.~Wang, A real-time quadrotor trajectory planning
  framework based on b-spline and nonuniform kinodynamic search, Journal of
  Field Robotics 38~(3) (2021) 452--475.

\bibitem{heidari2021trajectory}
H.~Heidari, M.~Saska, Trajectory planning of quadrotor systems for various
  objective functions, Robotica 39~(1) (2021) 137--152.

\bibitem{gao2020teach}
F.~Gao, L.~Wang, B.~Zhou, X.~Zhou, J.~Pan, S.~Shen, Teach-repeat-replan: A
  complete and robust system for aggressive flight in complex environments,
  IEEE Transactions on Robotics.

\bibitem{lee2010geometric}
T.~Lee, M.~Leok, N.~H. McClamroch, Geometric tracking control of a quadrotor
  uav on se (3), in: 49th IEEE conference on decision and control (CDC), IEEE,
  2010, pp. 5420--5425.

\bibitem{zinage20203d}
V.~Zinage, S.~H. Arul, D.~Manocha, 3d-ogse: Online smooth trajectory generation
  for quadrotors using generalized shape expansion in unknown 3d environments,
  arXiv preprint arXiv:2005.13229.

\bibitem{xi2020trajectory}
L.~Xi, Z.~Peng, L.~Jiao, Trajectory generation for quadrotor while tracking a
  moving target in cluttered environment, in: 2020 39th Chinese Control
  Conference (CCC), IEEE, 2020, pp. 6792--6797.

\bibitem{han2019fiesta}
L.~Han, F.~Gao, B.~Zhou, S.~Shen, Fiesta: Fast incremental euclidean distance
  fields for online motion planning of aerial robots, arXiv preprint
  arXiv:1903.02144.

\bibitem{murali2019perception}
V.~Murali, I.~Spasojevic, W.~Guerra, S.~Karaman, Perception-aware trajectory
  generation for aggressive quadrotor flight using differential flatness, in:
  2019 American Control Conference (ACC), IEEE, 2019, pp. 3936--3943.

\bibitem{abadi2019optimal}
A.~Abadi, A.~El~Amraoui, H.~Mekki, N.~Ramdani, Optimal trajectory generation
  and robust flatness--based tracking control of quadrotors, Optimal Control
  Applications and Methods 40~(4) (2019) 728--749.

\bibitem{ding2018trajectory}
W.~Ding, W.~Gao, K.~Wang, S.~Shen, Trajectory replanning for quadrotors using
  kinodynamic search and elastic optimization, in: 2018 IEEE International
  Conference on Robotics and Automation (ICRA), IEEE, 2018, pp. 7595--7602.

\bibitem{blochliger2018topomap}
F.~Blochliger, M.~Fehr, M.~Dymczyk, T.~Schneider, R.~Siegwart, Topomap:
  Topological mapping and navigation based on visual slam maps, in: 2018 IEEE
  International Conference on Robotics and Automation (ICRA), IEEE, 2018, pp.
  1--9.

\bibitem{lin2018autonomous}
Y.~Lin, F.~Gao, T.~Qin, W.~Gao, T.~Liu, W.~Wu, Z.~Yang, S.~Shen, Autonomous
  aerial navigation using monocular visual-inertial fusion, Journal of Field
  Robotics 35~(1) (2018) 23--51.

\bibitem{rosmann2017integrated}
C.~R{\"o}smann, F.~Hoffmann, T.~Bertram, Integrated online trajectory planning
  and optimization in distinctive topologies, Robotics and Autonomous Systems
  88 (2017) 142--153.

\bibitem{richter2016polynomial}
C.~Richter, A.~Bry, N.~Roy, Polynomial trajectory planning for aggressive
  quadrotor flight in dense indoor environments, in: Robotics Research,
  Springer, 2016, pp. 649--666.

\bibitem{landry2016aggressive}
B.~Landry, R.~Deits, P.~R. Florence, R.~Tedrake, Aggressive quadrotor flight
  through cluttered environments using mixed integer programming, in: 2016 IEEE
  international conference on robotics and automation (ICRA), IEEE, 2016, pp.
  1469--1475.

\bibitem{oleynikova2016continuous}
H.~Oleynikova, M.~Burri, Z.~Taylor, J.~Nieto, R.~Siegwart, E.~Galceran,
  Continuous-time trajectory optimization for online uav replanning, in: 2016
  IEEE/RSJ International Conference on Intelligent Robots and Systems (IROS),
  IEEE, 2016, pp. 5332--5339.

\bibitem{chen2016online}
J.~Chen, T.~Liu, S.~Shen, Online generation of collision-free trajectories for
  quadrotor flight in unknown cluttered environments, in: 2016 IEEE
  International Conference on Robotics and Automation (ICRA), IEEE, 2016, pp.
  1476--1483.

\bibitem{deits2015efficient}
R.~Deits, R.~Tedrake, Efficient mixed-integer planning for uavs in cluttered
  environments, in: 2015 IEEE international conference on robotics and
  automation (ICRA), IEEE, 2015, pp. 42--49.

\bibitem{mueller2015computationally}
M.~W. Mueller, M.~Hehn, R.~D'Andrea, A computationally efficient motion
  primitive for quadrocopter trajectory generation, IEEE Transactions on
  Robotics 31~(6) (2015) 1294--1310.

\bibitem{krusi2015lighting}
P.~Kr{\"u}si, B.~B{\"u}cheler, F.~Pomerleau, U.~Schwesinger, R.~Siegwart,
  P.~Furgale, Lighting-invariant adaptive route following using iterative
  closest point matching, Journal of Field Robotics 32~(4) (2015) 534--564.

\bibitem{wright2015coordinate}
S.~J. Wright, Coordinate descent algorithms, Mathematical Programming 151~(1)
  (2015) 3--34.

\bibitem{schulman2014motion}
J.~Schulman, Y.~Duan, J.~Ho, A.~Lee, I.~Awwal, H.~Bradlow, J.~Pan, S.~Patil,
  K.~Goldberg, P.~Abbeel, Motion planning with sequential convex optimization
  and convex collision checking, The International Journal of Robotics Research
  33~(9) (2014) 1251--1270.

\bibitem{pham2014general}
Q.-C. Pham, A general, fast, and robust implementation of the time-optimal path
  parameterization algorithm, IEEE Transactions on Robotics 30~(6) (2014)
  1533--1540.

\bibitem{pivtoraiko2013incremental}
M.~Pivtoraiko, D.~Mellinger, V.~Kumar, Incremental micro-uav motion replanning
  for exploring unknown environments, in: 2013 IEEE International Conference on
  Robotics and Automation, IEEE, 2013, pp. 2452--2458.

\bibitem{macallister2013path}
B.~MacAllister, J.~Butzke, A.~Kushleyev, H.~Pandey, M.~Likhachev, Path planning
  for non-circular micro aerial vehicles in constrained environments, in: 2013
  IEEE International Conference on Robotics and Automation, IEEE, 2013, pp.
  3933--3940.

\bibitem{zucker2013chomp}
M.~Zucker, N.~Ratliff, A.~D. Dragan, M.~Pivtoraiko, M.~Klingensmith, C.~M.
  Dellin, J.~A. Bagnell, S.~S. Srinivasa, Chomp: Covariant hamiltonian
  optimization for motion planning, The International Journal of Robotics
  Research 32~(9-10) (2013) 1164--1193.

\bibitem{mellinger2012mixed}
D.~Mellinger, A.~Kushleyev, V.~Kumar, Mixed-integer quadratic program
  trajectory generation for heterogeneous quadrotor teams, in: 2012 IEEE
  international conference on robotics and automation, IEEE, 2012, pp.
  477--483.

\bibitem{bhattacharya2012topological}
S.~Bhattacharya, M.~Likhachev, V.~Kumar, Topological constraints in
  search-based robot path planning, Autonomous Robots 33~(3) (2012) 273--290.

\bibitem{kalakrishnan2011stomp}
M.~Kalakrishnan, S.~Chitta, E.~Theodorou, P.~Pastor, S.~Schaal, Stomp:
  Stochastic trajectory optimization for motion planning, in: 2011 IEEE
  international conference on robotics and automation, IEEE, 2011, pp.
  4569--4574.

\bibitem{harabor2011online}
D.~D. Harabor, A.~Grastien, et~al., Online graph pruning for pathfinding on
  grid maps., in: AAAI, 2011, pp. 1114--1119.

\bibitem{lovi2010incremental}
D.~Lovi, N.~Birkbeck, D.~Cobzas, M.~Jagersand, Incremental free-space carving
  for real-time 3d reconstruction, in: Fifth international symposium on 3D data
  processing visualization and transmission (3DPVT), 2010.

\bibitem{trawny2010interrobot}
N.~Trawny, X.~S. Zhou, K.~Zhou, S.~I. Roumeliotis, Interrobot transformations
  in 3-d, IEEE Transactions on Robotics 26~(2) (2010) 226--243.

\bibitem{mehrez2017optimization}
M.~W. Mehrez, G.~K. Mann, R.~G. Gosine, An optimization based approach for
  relative localization and relative tracking control in multi-robot systems,
  Journal of Intelligent \& Robotic Systems 85~(2) (2017) 385--408.

\bibitem{mpc_mc}
G.~Kulathunga, D.~Devitt, A.~Klimchik, Trajectory tracking for quadrotors: An
  optimization-based planning followed by controlling approach, Journal of
  Field Robotics 39~(7) (2022) 1003--1013.

\bibitem{van1998real}
M.~J. Van~Nieuwstadt, R.~M. Murray, Real-time trajectory generation for
  differentially flat systems, International Journal of Robust and Nonlinear
  Control: IFAC-Affiliated Journal 8~(11) (1998) 995--1020.

\bibitem{sferrazza2016numerical}
C.~Sferrazza, D.~Pardo, J.~Buchli, Numerical search for local (partial)
  differential flatness, in: 2016 IEEE/RSJ International Conference on
  Intelligent Robots and Systems (IROS), IEEE, 2016, pp. 3640--3646.

\bibitem{krishnan2019towards}
S.~Krishnan, G.~A. Rajagopalan, S.~Kandhasamy, M.~Shanmugavel, Towards scalable
  continuous-time trajectory optimization for multi-robot navigation, arXiv
  preprint arXiv:1910.13463.

\bibitem{dolgov2010path}
D.~Dolgov, S.~Thrun, M.~Montemerlo, J.~Diebel, Path planning for autonomous
  vehicles in unknown semi-structured environments, The International Journal
  of Robotics Research 29~(5) (2010) 485--501.

\bibitem{florence2020integrated}
P.~Florence, J.~Carter, R.~Tedrake, Integrated perception and control at high
  speed: Evaluating collision avoidance maneuvers without maps, in: Algorithmic
  Foundations of Robotics XII, Springer, 2020, pp. 304--319.

\bibitem{lopez2017aggressive}
B.~T. Lopez, J.~P. How, Aggressive 3-d collision avoidance for high-speed
  navigation., in: ICRA, 2017, pp. 5759--5765.

\bibitem{gordon1974b}
W.~J. Gordon, R.~F. Riesenfeld, B-spline curves and surfaces, in: Computer
  aided geometric design, Elsevier, 1974, pp. 95--126.

\bibitem{sethian1999level}
J.~A. Sethian, Level set methods and fast marching methods: evolving interfaces
  in computational geometry, fluid mechanics, computer vision, and materials
  science, Vol.~3, Cambridge university press, 1999.

\bibitem{sava20013}
P.~Sava, S.~Fomel, 3-d traveltime computation using huygens wavefront tracing,
  Geophysics 66~(3) (2001) 883--889.

\bibitem{lavalle2006planning}
S.~M. LaValle, Planning algorithms, Cambridge university press, 2006.

\bibitem{bergman2019optimization}
K.~Bergman, O.~Ljungqvist, T.~Glad, D.~Axehill, An optimization-based receding
  horizon trajectory planning algorithm, arXiv preprint arXiv:1912.05259.

\bibitem{head1985broyden}
J.~D. Head, M.~C. Zerner, A broyden—fletcher—goldfarb—shanno optimization
  procedure for molecular geometries, Chemical physics letters 122~(3) (1985)
  264--270.

\bibitem{de1971subroutine}
C.~de~Boor, Subroutine package for calculating with b-splines, Los Alamos
  Scient. Lab. Report LA-4728-MS.

\bibitem{qin2000general}
K.~Qin, General matrix representations for b-splines, The Visual Computer
  16~(3-4) (2000) 177--186.

\bibitem{hu2019real}
J.~Hu, Z.~Ma, Y.~Niu, W.~Tian, W.~Yao, Real-time trajectory replanning for
  quadrotor using octomap and uniform b-splines, in: International Conference
  on Intelligent Robotics and Applications, Springer, 2019, pp. 727--741.

\bibitem{flores2008real}
M.~E. Flores~Contreras, Real-time trajectory generation for constrained
  nonlinear dynamical systems using non-uniform rational b-spline basis
  functions, Ph.D. thesis, California Institute of Technology (2008).

\bibitem{preiss2017downwash}
J.~A. Preiss, W.~H{\"o}nig, N.~Ayanian, G.~S. Sukhatme, Downwash-aware
  trajectory planning for large quadrotor teams, in: 2017 IEEE/RSJ
  International Conference on Intelligent Robots and Systems (IROS), IEEE,
  2017, pp. 250--257.

\bibitem{hornung2013octomap}
A.~Hornung, K.~M. Wurm, M.~Bennewitz, C.~Stachniss, W.~Burgard, Octomap: An
  efficient probabilistic 3d mapping framework based on octrees, Autonomous
  robots 34~(3) (2013) 189--206.

\bibitem{kala2013rapidly}
R.~Kala, Rapidly exploring random graphs: motion planning of multiple mobile
  robots, Advanced Robotics 27~(14) (2013) 1113--1122.

\bibitem{zhu2015convex}
Z.~Zhu, E.~Schmerling, M.~Pavone, A convex optimization approach to smooth
  trajectories for motion planning with car-like robots, in: 2015 54th IEEE
  Conference on Decision and Control (CDC), IEEE, 2015, pp. 835--842.

\bibitem{quinlan1993elastic}
S.~Quinlan, O.~Khatib, Elastic bands: Connecting path planning and control, in:
  [1993] Proceedings IEEE International Conference on Robotics and Automation,
  IEEE, 1993, pp. 802--807.

\bibitem{jacobson1970differential}
D.~Jacobson, D.~Mayne, Differential dynamic programming elsevier new york
  (1970).

\bibitem{theodorou2013information}
E.~Theodorou, D.~Krishnamurthy, E.~Todorov, From information theoretic
  dualities to path integral and kullback-leibler control: Continuous and
  discrete time formulations, in: The Sixteenth Yale Workshop on Adaptive and
  Learning Systems, 2013.

\bibitem{lewis1995optimal}
F.~L. Lewis, V.~L. Syrmos, Optimal control, john-wiley\&sons, New York.

\bibitem{sun2015stochastic}
W.~Sun, J.~Van Den~Berg, R.~Alterovitz, Stochastic extended lqr:
  Optimization-based motion planning under uncertainty, in: Algorithmic
  Foundations of Robotics XI, Springer, 2015, pp. 609--626.

\bibitem{van2012lqg}
J.~Van Den~Berg, D.~Wilkie, S.~J. Guy, M.~Niethammer, D.~Manocha,
  Lqg-obstacles: Feedback control with collision avoidance for mobile robots
  with motion and sensing uncertainty, in: 2012 IEEE International Conference
  on Robotics and Automation, IEEE, 2012, pp. 346--353.

\bibitem{likhachev2004ara}
M.~Likhachev, G.~J. Gordon, S.~Thrun, Ara*: Anytime a* with provable bounds on
  sub-optimality, in: Advances in neural information processing systems, 2004,
  pp. 767--774.

\bibitem{karaman2011sampling}
S.~Karaman, E.~Frazzoli, Sampling-based algorithms for optimal motion planning,
  The international journal of robotics research 30~(7) (2011) 846--894.

\bibitem{perez2012lqr}
A.~Perez, R.~Platt, G.~Konidaris, L.~Kaelbling, T.~Lozano-Perez, Lqr-rrt*:
  Optimal sampling-based motion planning with automatically derived extension
  heuristics, in: 2012 IEEE International Conference on Robotics and
  Automation, IEEE, 2012, pp. 2537--2542.

\bibitem{kulathunga2020path}
G.~Kulathunga, D.~Devitt, R.~Fedorenko, A.~Klimchik, Path planning followed by
  kinodynamic smoothing for multirotor aerial vehicles (mavs), Russian Journal
  of Nonlinear Dynamics 17~(4) (2021) 491--505.

\bibitem{pacelli2018integration}
V.~Pacelli, O.~Arslan, D.~E. Koditschek, Integration of local geometry and
  metric information in sampling-based motion planning, in: 2018 IEEE
  International Conference on Robotics and Automation (ICRA), IEEE, 2018, pp.
  3061--3068.

\bibitem{mason1985robot}
M.~T. Mason, J.~K. Salisbury~Jr, Robot hands and the mechanics of manipulation,
  The MIT Press, Cambridge, MA, 1985.

\bibitem{liu2017search}
S.~Liu, N.~Atanasov, K.~Mohta, V.~Kumar, Search-based motion planning for
  quadrotors using linear quadratic minimum time control, in: 2017 IEEE/RSJ
  international conference on intelligent robots and systems (IROS), IEEE,
  2017, pp. 2872--2879.

\bibitem{ames2014rapidly}
A.~D. Ames, K.~Galloway, K.~Sreenath, J.~W. Grizzle, Rapidly exponentially
  stabilizing control lyapunov functions and hybrid zero dynamics, IEEE
  Transactions on Automatic Control 59~(4) (2014) 876--891.

\bibitem{wu2015safety}
G.~Wu, K.~Sreenath, Safety-critical and constrained geometric control synthesis
  using control lyapunov and control barrier functions for systems evolving on
  manifolds, in: 2015 American Control Conference (ACC), IEEE, 2015, pp.
  2038--2044.

\bibitem{kolmanovsky2014reference}
I.~Kolmanovsky, E.~Garone, S.~Di~Cairano, Reference and command governors: A
  tutorial on their theory and automotive applications, in: 2014 American
  Control Conference, IEEE, 2014, pp. 226--241.

\bibitem{garone2015explicit}
E.~Garone, M.~M. Nicotra, Explicit reference governor for constrained nonlinear
  systems, IEEE Transactions on Automatic Control 61~(5) (2015) 1379--1384.

\bibitem{arslan2017smooth}
O.~Arslan, D.~E. Koditschek, Smooth extensions of feedback motion planners via
  reference governors, in: 2017 IEEE International Conference on Robotics and
  Automation (ICRA), IEEE, 2017, pp. 4414--4421.

\bibitem{li2020fast}
Z.~Li, O.~Arslan, N.~Atanasov, Fast and safe path-following control using a
  state-dependent directional metric, arXiv preprint arXiv:2002.02038.

\bibitem{2005.00985}
Y.~Aoyama, G.~Boutselis, A.~Patel, E.~A. Theodorou, Constrained differential
  dynamic programming revisited (2020).
\newblock \href {http://arxiv.org/abs/arXiv:2005.00985}
  {\path{arXiv:arXiv:2005.00985}}.

\bibitem{liu2016pid}
C.~Liu, J.~Pan, Y.~Chang, Pid and lqr trajectory tracking control for an
  unmanned quadrotor helicopter: Experimental studies, in: 2016 35th Chinese
  Control Conference (CCC), IEEE, 2016, pp. 10845--10850.

\bibitem{cowling2006optimal}
I.~D. Cowling, J.~F. Whidborne, A.~K. Cooke, Optimal trajectory planning and
  lqr control for a quadrotor uav, in: International Conference on Control,
  2006.

\bibitem{BANGURA201411773}
M.~Bangura, R.~Mahony,
  \href{http://www.sciencedirect.com/science/article/pii/S1474667016434890}{Real-time
  model predictive control for quadrotors}, IFAC Proceedings Volumes 47~(3)
  (2014) 11773 -- 11780, 19th IFAC World Congress.
\newblock \href
  {http://dx.doi.org/https://doi.org/10.3182/20140824-6-ZA-1003.00203}
  {\path{doi:https://doi.org/10.3182/20140824-6-ZA-1003.00203}}.
\newline\urlprefix\url{http://www.sciencedirect.com/science/article/pii/S1474667016434890}

\bibitem{ohtsuka1997real}
T.~Ohtsuka, H.~A. Fujii, Real-time optimization algorithm for nonlinear
  receding-horizon control, Automatica 33~(6) (1997) 1147--1154.

\bibitem{deng2019parallel}
H.~Deng, T.~Ohtsuka, A parallel newton-type method for nonlinear model
  predictive control, Automatica 109 (2019) 108560.

\bibitem{mohamed2020model}
I.~S. Mohamed, G.~Allibert, P.~Martinet, Model predictive path integral control
  framework for partially observable navigation: A quadrotor case study, arXiv
  preprint arXiv:2004.08641.

\bibitem{olivares2012see}
M.~A. Olivares-Mendez, P.~Campoy, I.~Mellado-Bataller, L.~Mejias, See-and-avoid
  quadcopter using fuzzy control optimized by cross-entropy, in: 2012 IEEE
  International Conference on Fuzzy Systems, Ieee, 2012, pp. 1--7.

\bibitem{tordesillas2020faster}
J.~Tordesillas, B.~T. Lopez, M.~Everett, J.~P. How, Faster: Fast and safe
  trajectory planner for flights in unknown environments, arXiv preprint
  arXiv:2001.04420.

\bibitem{quinlan1994real}
S.~Quinlan, Real-time modification of collision-free paths, no. 1537, Stanford
  University Stanford, 1994.

\bibitem{zhou2020robust}
B.~Zhou, F.~Gao, J.~Pan, S.~Shen, Robust real-time uav replanning using guided
  gradient-based optimization and topological paths, in: 2020 IEEE
  International Conference on Robotics and Automation (ICRA), IEEE, 2020, pp.
  1208--1214.

\bibitem{powell2009bobyqa}
M.~J. Powell, The bobyqa algorithm for bound constrained optimization without
  derivatives, Cambridge NA Report NA2009/06, University of Cambridge,
  Cambridge (2009) 26--46.

\bibitem{liu1989limited}
D.~C. Liu, J.~Nocedal, On the limited memory bfgs method for large scale
  optimization, Mathematical programming 45~(1-3) (1989) 503--528.

\bibitem{houska2011acado}
B.~Houska, H.~J. Ferreau, M.~Diehl, Acado toolkit—an open-source framework
  for automatic control and dynamic optimization, Optimal Control Applications
  and Methods 32~(3) (2011) 298--312.

\bibitem{kraft1988software}
D.~Kraft, A software package for sequential quadratic programming.
  forschungsbericht-deutsche forschungs-und versuchsanstalt fur luft-und
  raumfahrt, DFVLR, K{\"o}ln.

\bibitem{parikh2014block}
N.~Parikh, S.~Boyd, Block splitting for distributed optimization, Mathematical
  Programming Computation 6~(1) (2014) 77--102.

\bibitem{fougner2018parameter}
C.~Fougner, S.~Boyd, Parameter selection and preconditioning for a graph form
  solver, in: Emerging Applications of Control and Systems Theory, Springer,
  2018, pp. 41--61.

\bibitem{svanberg2002class}
K.~Svanberg, A class of globally convergent optimization methods based on
  conservative convex separable approximations, SIAM journal on optimization
  12~(2) (2002) 555--573.

\bibitem{liu2019optimization}
X.~Liu, R.~D. Wiersma, Optimization based trajectory planning for real-time
  6dof robotic patient motion compensation systems, PloS one 14~(1) (2019)
  e0210385.

\bibitem{foehn2017fast}
P.~Foehn, D.~Falanga, N.~Kuppuswamy, R.~Tedrake, D.~Scaramuzza, Fast trajectory
  optimization for agile quadrotor maneuvers with a cable-suspended payload.

\bibitem{geisert2016trajectory}
M.~Geisert, N.~Mansard, Trajectory generation for quadrotor based systems using
  numerical optimal control, in: 2016 IEEE international conference on robotics
  and automation (ICRA), IEEE, 2016, pp. 2958--2964.

\bibitem{shen2012stochastic}
S.~Shen, N.~Michael, V.~Kumar, Stochastic differential equation-based
  exploration algorithm for autonomous indoor 3d exploration with a
  micro-aerial vehicle, The International Journal of Robotics Research 31~(12)
  (2012) 1431--1444.

\bibitem{johnson2014nlopt}
S.~G. Johnson, The nlopt nonlinear-optimization package, 2014, URL
  http://ab-initio. mit. edu/nlopt.

\bibitem{zhou2020ego}
X.~Zhou, Z.~Wang, H.~Ye, C.~Xu, F.~Gao, Ego-planner: An esdf-free
  gradient-based local planner for quadrotors, IEEE Robotics and Automation
  Letters.

\bibitem{steiha1983truncatednewton}
R.~D.~T. STEIHA, Truncatednewton algorithmsforlarge-scale optimization, Math.
  Programming 26 (1983) 190--212.

\bibitem{andersen2000mosek}
E.~D. Andersen, K.~D. Andersen, The mosek interior point optimizer for linear
  programming: an implementation of the homogeneous algorithm, in: High
  performance optimization, Springer, 2000, pp. 197--232.

\bibitem{gurobi2018gurobi}
I.~Gurobi~Optimization, Gurobi optimizer reference manual (2018).

\bibitem{stellato2020osqp}
B.~Stellato, G.~Banjac, P.~Goulart, A.~Bemporad, S.~Boyd, Osqp: An operator
  splitting solver for quadratic programs, Mathematical Programming Computation
  (2020) 1--36.

\bibitem{kamel2017model}
M.~Kamel, T.~Stastny, K.~Alexis, R.~Siegwart, Model predictive control for
  trajectory tracking of unmanned aerial vehicles using robot operating system,
  in: Robot operating system (ROS), Springer, 2017, pp. 3--39.

\bibitem{gertz2003object}
E.~M. Gertz, S.~J. Wright, Object-oriented software for quadratic programming,
  ACM Transactions on Mathematical Software (TOMS) 29~(1) (2003) 58--81.

\bibitem{tordesillas2020mader}
J.~Tordesillas, J.~P. How, Mader: Trajectory planner in multi-agent and dynamic
  environments, arXiv preprint arXiv:2010.11061.

\bibitem{tordesillas2020minvo}
J.~Tordesillas, J.~P. How, Minvo basis: Finding simplexes with minimum volume
  enclosing polynomial curves, arXiv preprint arXiv:2010.10726.

\bibitem{conn1991globally}
A.~R. Conn, N.~I. Gould, P.~Toint, A globally convergent augmented lagrangian
  algorithm for optimization with general constraints and simple bounds, SIAM
  Journal on Numerical Analysis 28~(2) (1991) 545--572.

\bibitem{tang2019real}
L.~Tang, H.~Wang, P.~Li, Y.~Wang, Real-time trajectory generation for
  quadrotors using b-spline based non-uniform kinodynamic search, in: 2019 IEEE
  International Conference on Robotics and Biomimetics (ROBIO), IEEE, 2019, pp.
  1133--1138.

\bibitem{andersson2019casadi}
J.~A. Andersson, J.~Gillis, G.~Horn, J.~B. Rawlings, M.~Diehl, Casadi: a
  software framework for nonlinear optimization and optimal control,
  Mathematical Programming Computation 11~(1) (2019) 1--36.

\bibitem{biegler2009large}
L.~T. Biegler, V.~M. Zavala, Large-scale nonlinear programming using ipopt: An
  integrating framework for enterprise-wide dynamic optimization, Computers \&
  Chemical Engineering 33~(3) (2009) 575--582.

\end{thebibliography}

\end{document}